\documentclass[runningheads]{llncs}

\usepackage[final,year=2026,ID=4717]{eccv}
\usepackage{url}
\usepackage{enumitem}
\usepackage{amsmath}
\usepackage{graphicx}   % 导入插图宏包
\usepackage{subcaption} % 导入子图宏包
\usepackage{graphicx}
\graphicspath{{figure/}}
\usepackage{multirow}
\usepackage{booktabs}                % 实线横线
\usepackage[table]{xcolor}           % 表格背景色
\usepackage{multirow}                % 表头合并
\usepackage{caption}                 % 表题格式
\definecolor{grayrow}{gray}{0.85}
\usepackage{amssymb}       % 用于 \checkmark 符号
\usepackage{pifont}        % 用于 \ding{55} (叉号)
\usepackage[table]{xcolor} % 用于 \rowcolor (需要 xcolor 宏包并加载 table 选项)
\usepackage{nicematrix}
\usepackage{adjustbox}
\usepackage{makecell}
\usepackage{hhline} % <--- 在导言区添加这个包
\usepackage{tabularx}
\usepackage{ragged2e}
\usepackage{colortbl} % 用于 \rowcolor
\usepackage{mathtools}
\usepackage{xcolor}
\usepackage{threeparttable}
\usepackage{wrapfig}
\usepackage{eccv}

\usepackage{eccvabbrv}

\usepackage{graphicx}
\usepackage{booktabs}

\usepackage[accsupp]{axessibility}  % Improves PDF readability for those with disabilities.

\usepackage{orcidlink}

\begin{document}

% ---------------------------------------------------------------
% TODO REVIEW: Replace with your title
\title{ResilPhase: Plug-and-Play Phase Mapping and Noise-Resilient Macro-Trajectory Extrapolation for Diffusion Acceleration} 

% TODO REVIEW: If the paper title is too long for the running head, you can set
% an abbreviated paper title here. If not, comment out.
\titlerunning{Abbreviated paper title}

% TODO FINAL: Replace with your author list. 
% Include the authors' OCRID for the camera-ready version, if at all possible.
\author{Qicheng Zhao\orcidlink{0009-0001-4389-1742} \and
Yu Li \and
Qi Sun\orcidlink{0000-0001-5153-6698} \and
Zheyu Yan\thanks{Corresponding author.}}

% TODO FINAL: Replace with an abbreviated list of authors.
\authorrunning{Q.~Zhao et al.}
% First names are abbreviated in the running head.
% If there are more than two authors, 'et al.' is used.

% TODO FINAL: Replace with your institution list.
% \institute{Princeton University, Princeton NJ 08544, USA \and
% Springer Heidelberg, Tiergartenstr.~17, 69121 Heidelberg, Germany
% \email{lncs@springer.com}\\
\institute{Zhejiang University, Hangzhou, China\\
\email{zyan2@zju.edu.cn}}
% \url{http://www.springer.com/gp/computer-science/lncs} \and
% ABC Institute, Rupert-Karls-University Heidelberg, Heidelberg, Germany\\
% \email{\{abc,lncs\}@uni-heidelberg.de}}

\maketitle

\begin{abstract}
The adoption of powerful diffusion models is hindered by their significant inference latency. Recent ``cache-then-forecast'' schemes alleviate this issue by accelerating DiTs using derivative-based polynomials, but they suffer from severe quality degradation at high acceleration ratios. Our analysis reveals its root cause: the discrete extrapolation performed on representations that are misaligned with the continuous diffusion trajectory and are numerically unstable. Thus, accelerated DiTs suffer from accumulated spatial errors, noisy derivative amplification, and high-order instability. We therefore reformulate accelerated inference as stable macro-trajectory extrapolation in ordinary differential equation (ODE) space. Instead of predicting intermediate features, we align forecasting with the model's Global Drift (GD), i.e., the end-to-end state evolution, thereby eliminating feature inconsistency and memory overhead. However, even this smooth macro-trajectory remains vulnerable to the derivative fallacy: its higher-order temporal derivatives are intrinsically noisy. Thus, we introduce a derivative-free barycentric Lagrange extrapolator to effectively bypass derivative instability and approximation error. We further propose a bounded Phase Mapping that regularizes the extrapolation domain, suppressing oscillatory error growth. These elements collectively constitute ResilPhase, a noise-resilient acceleration framework. Experiments on FLUX.1-dev and HunyuanVideo demonstrate state-of-the-art fidelity under aggressive acceleration ratios. Code is publicly available at \url{https://github.com/zqc214/ResilPhase}.
    % \keywords{Diffusion Transformers \and Efficient Inference \and Trajectory Extrapolation}
\end{abstract}

\section{Introduction}
\label{sec:intro}
Diffusion Transformers (DiTs)~\cite{ho2020denoising,peebles2023scalable} have become the gold standard for high-fidelity visual generation~\cite{blattmann2023stable,rombach2022high,ma2024latte,zheng2024open,songdenoising}, because of their scalable architecture. However, their performance is achieved through an iterative denoising process, requiring tens to hundreds of sequential forward steps that can not be parallelized. This has become a significant latency bottleneck that obstructs real-time and large-scale deployment of DiTs. FLUX.1-dev takes 23.69 seconds to generate an image on an A100 graphics card.

To address this latency bottleneck, recent training-free acceleration efforts have evolved from passive ``cache-then-reuse'' paradigms \cite{ma2024deepcache,wang2024pfdiff,li2024faster} to ``cache-then-forecast'' approaches \cite{liu2025reusing,liu2025speca,zheng2025forecast} that employ polynomials to predict feature trajectories. However, this prevailing paradigm suffers from severe quality degradation at high acceleration ratios. We argue this failure stems from a single root cause: discrete extrapolation is performed on intermediate representations misaligned with the continuous diffusion trajectory and numerically unstable. This root cause manifests in three intertwined limitations.

First, from a spatial perspective, most methods rely on a computationally intensive layer-wise paradigm. Fitting polynomials to highly erratic micro-features within Transformer blocks cascades errors across the network depth, incurring massive memory overhead. While directly forecasting absolute outputs (e.g., FreqCa~\cite{liu2025freqca}) bypasses this, as highlighted by $\Delta$-DiT~\cite{chen2024delta-dit}, it discards highly correlated input priors, restricting prediction fidelity bounds. Conversely, methods reusing residual displacements (e.g., $\Delta$-DiT~\cite{chen2024delta-dit}) remain trapped in localized micro-updates, missing macroscopic continuous dynamics. Thus, existing paradigms lack an end-to-end target that simultaneously preserves input priors and aligns strictly with the ODE vector field.

\begin{wrapfigure}{r}{0.6\textwidth}
\vspace{-23pt}
\centering
\includegraphics[width=0.6\textwidth]{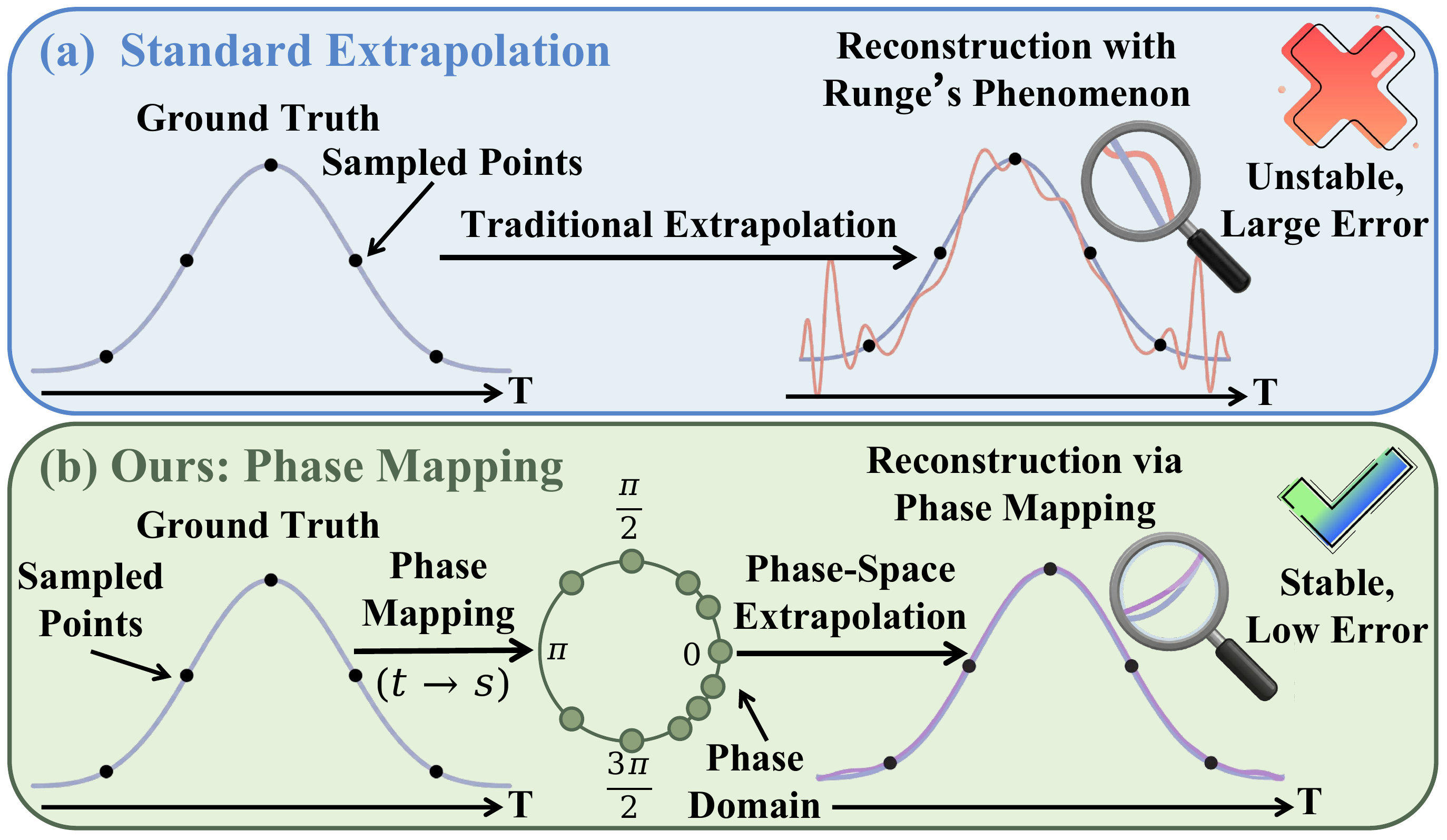} 
\vspace{-20pt}  
% \caption{\textbf{Suppressing numerical instability via Phase Mapping.} (a) Standard trajectory extrapolation on uniform timesteps suffers from Runge's phenomenon, causing chaotic edge oscillations. (b) Our method maps discrete steps onto a bounded phase space, strictly minimizing the error bound for a stable, low-error fit.}
\caption{\textbf{Suppressing numerical instability via Phase Mapping.} (a) Standard trajectory extrapolation on uniform timesteps suffers from Runge's phenomenon, causing chaotic edge oscillations. (b) Our method maps discrete steps to a bounded phase space, minimizing the error bound for a stable fit.}
\label{fig:pm-motivation}
\vspace{-25pt} 
\end{wrapfigure}

\begin{figure}[t]
\centering  % ✅ 关键：让整个图居中
\begin{subfigure}[b]{\linewidth}
  \centering
  \includegraphics[width=\linewidth]{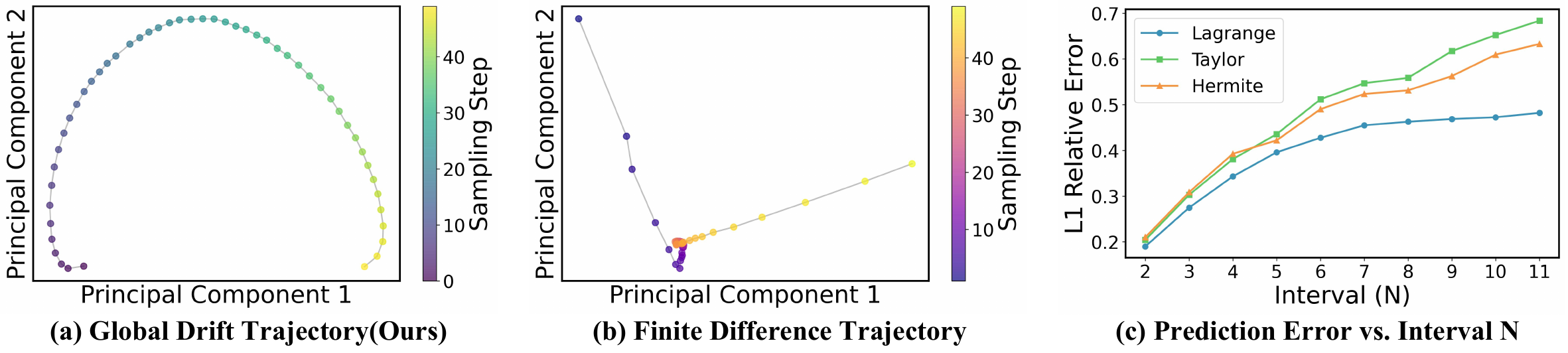}
  
\end{subfigure}

\vspace{-1em} % 手动调间距（可选）

\caption{\textbf{Effectiveness of the derivative-free prediction strategy.} Unlike (a) our smooth global macro-trajectory, (b) the finite-difference derivative trajectory used by prior methods is intrinsically chaotic. Applying polynomial extrapolation to such unstable dynamics causes severe prediction deviations. As shown in (c), evaluating pure mathematical predictors reveals that our derivative-free Lagrange extrapolator maintains significantly lower $L_1$ relative error across varying extrapolation intervals ($N$) than derivative-based solvers (Taylor, Hermite).}

\label{fig:pca}
\vspace{-2em}
\end{figure}

Second, from a temporal perspective, current forecasting methods (e.g., TaylorSeer~\cite{liu2025reusing} and HiCache~\cite{feng2025hicache}) rely exclusively on derivative-based approximations. They critically overlook the mathematical nature of the dynamic trajectory: while the macro-trajectory itself is smooth, its higher-order derivatives over time are intrinsically chaotic (as shown in Fig.~\ref{fig:pca}(b)). Estimating these derivatives via finite differences exponentially amplifies the noise. 

Third, from a numerical approximation standpoint, performing polynomial extrapolation over uniform, discrete time steps inherently triggers Runge's phenomenon. This numerical instability causes the extrapolation error bound to grow uncontrollably at the interval edges, inevitably leading to catastrophic quality degradation at aggressive acceleration ratios.

% \begin{wrapfigure}{r}{0.6\textwidth}
% \vspace{-23pt}
% \centering
% \includegraphics[width=0.6\textwidth]{figure/pm-motivation.pdf} 
% \vspace{-20pt}  
% % \caption{\textbf{Suppressing numerical instability via Phase Mapping.} (a) Standard trajectory extrapolation on uniform timesteps suffers from Runge's phenomenon, causing chaotic edge oscillations. (b) Our method maps discrete steps onto a bounded phase space, strictly minimizing the error bound for a stable, low-error fit.}
% \caption{\textbf{Suppressing numerical instability via Phase Mapping.} (a) Standard trajectory extrapolation on uniform timesteps suffers from Runge's phenomenon, causing chaotic edge oscillations. (b) Our method maps discrete steps to a bounded phase space, minimizing the error bound for a stable fit.}
% \label{fig:pm-motivation}
% \vspace{-20pt} 
% \end{wrapfigure}

To overcome these intertwined bottlenecks, we propose \textbf{ResilPhase}, a noise-resilient acceleration framework explicitly addressing the spatial, temporal, and numerical limitations of existing paradigms. First, to resolve spatial cascading errors inherent in layer-wise forecasting, ResilPhase shifts the prediction objective to the network's macroscopic dynamic evolution. We propose ODE-Aligned Macro-Trajectory Targeting, formulating the prediction target as the Global Drift (GD): the end-to-end state displacement between the model's final output and initial input. Unlike layer-wise approaches fitting highly oscillatory micro-signals within Transformer blocks, forecasting the GD aligns the extrapolator with the continuous probability flow ODE. This paradigm shift eliminates the memory overhead of caching block features and strictly severs error accumulation across network layers.

While predicting the GD yields a smooth macro-trajectory ideal for forecasting (Fig.~\ref{fig:pca}(a)), it paradoxically amplifies the weakness of derivative-based methods. Like layer-wise micro-features, the GD's higher-order temporal derivatives remain intrinsically chaotic. Applying derivative-based solvers like Taylor or Hermite to such noisy signals causes divergent predictions. Furthermore, finite-difference estimation of these derivatives across discrete steps introduces compounding approximation errors. To avoid this, we propose a derivative-free extrapolator inspired by Lagrange interpolation. Avoiding the standard $O(N^2)$ formulation, we develop a barycentric prediction scheme caching and reusing weights. This reduces complexity to $O(N)$ while eliminating derivative noise. As a purely mathematical predictor, our approach achieves significantly lower and more stable error across extrapolation intervals than prior solvers (Fig.~\ref{fig:pca}(c)).

Even with a robust derivative-free formulation, polynomial extrapolation over uniform discrete time steps remains susceptible to Runge's phenomenon. As shown in Fig.~\ref{fig:pm-motivation}(a), this numerical instability causes severe edge oscillations; ResilPhase is the first work to identify this as a critical bottleneck in diffusion acceleration. To address this, we introduce a plug-and-play Phase Mapping mechanism (Fig.~\ref{fig:pm-motivation}(b)). As the first theoretically grounded technique to remap the discrete temporal extrapolation domain, we bridge classical numerical analysis and diffusion acceleration by utilizing Chebyshev nodes to project linear time steps ($t$) into a bounded phase domain ($s$), achieving optimal stability for class-conditional generation. Furthermore, we propose a data-driven Balanced Mapping tailored to complex text-to-image and video tasks. By performing extrapolation within this bounded continuous space, our mappings effectively transform a divergent numerical issue into a highly stable prediction, strictly minimizing the mathematical upper bound of the extrapolation error.

To summarize, our primary contributions are as follows:
\begin{itemize}
    \item We propose ODE-Aligned Macro-Trajectory Targeting. By forecasting the model's Global Drift (GD) rather than layer-wise features, we strictly sever spatial cascading errors, align with the continuous diffusion ODE, and eliminate heavy memory overhead.
    \item We introduce a noise-resilient, derivative-free Barycentric Lagrange extrapolator. This $O(N)$ framework perfectly synergizes with the GD by completely bypassing the inherent noise and chaotic high-order derivatives of finite-difference approximations.
    \item We design a plug-and-play Phase Mapping mechanism. By non-linearly projecting discrete time steps into a bounded phase space, it regularizes the extrapolation domain to suppress oscillatory error growth, strictly minimizing the polynomial extrapolation error bound.
    \item Extensive experiments show ResilPhase achieves $\sim$5$\times$ speedups on FLUX.1-dev and HunyuanVideo while maintaining highly competitive fidelity. Furthermore, Phase Mapping acts as a plug-and-play stabilizer, mitigating extrapolation errors in existing derivative-based accelerators.
\end{itemize}

\section{Related Work}
\label{sec:related}
\subsection{Diffusion Transformers and Traditional Acceleration}
Diffusion Transformers (DiTs) achieve high-fidelity generation through a sequential noise-reversal process, which fundamentally imposes a severe inference latency bottleneck. To mitigate this, early acceleration efforts primarily focused on model compression, such as network pruning~\cite{fang2023structuralpruningdiffusionmodels,zhu2024dip,wu2025anas} and quantization~\cite{kim2025ditto,li2023q,shang2023post,guo2024hardware}, or step reduction via efficient solvers~\cite{lu2022dpm,lu2025dpm,Yin_2026_CVPR,yin2026depmatch} and knowledge distillation~\cite{salimansprogressive,song2023consistency,li2023snapfusion,zhou2025deploying}. However, these approaches typically demand computationally expensive retraining to recover generation quality and often rely on complex algorithmic designs. This reliance fundamentally limits their plug-and-play generality, driving recent research toward more flexible, training-free acceleration paradigms.

\subsection{Acceleration Based on Feature Caching}
Feature caching, the main training-free diffusion acceleration strategy, evolved from passively reusing temporal features (DeepCache~\cite{ma2024deepcache}, FORA~\cite{selvaraju2024fora}, $\Delta$-DiT~\cite{chen2024delta-dit}, TeaCache~\cite{liu2025timestep}, ToCa~\cite{zou2024accelerating}) to actively predicting them (TaylorSeer~\cite{liu2025reusing}, FoCa~\cite{zheng2025forecast}, HiCache~\cite{feng2025hicache}, SpeCa~\cite{liu2025speca}, ClusCa~\cite{zheng2025compute}, FreqCa~\cite{liu2025freqca}). While recent methods like DiCache~\cite{bu2025dicache} dynamically adjust caching intervals, these predictive approaches share critical limitations. 
Spatially, layer-wise forecasting incurs memory overhead and amplifies micro-feature errors across network depth. Temporally, these methods implicitly assume smooth high-order derivatives, relying on finite-difference approximations. However, approaches using Taylor series~\cite{liu2025reusing, liu2025speca}, Hermite polynomials~\cite{feng2025hicache, liu2025freqca}, or ODE-based forecasting~\cite{zheng2025forecast} encounter a mathematical bottleneck: while the macro-trajectory is smooth, its temporal derivatives remain intrinsically chaotic. Consequently, applying derivative-based forecasting to these noisy signals makes predictions susceptible to numerical instabilities like Runge’s phenomenon, restricting maximum acceleration and robustness.

\begin{figure*}[t]
\centering
\includegraphics[trim= 0 670 1850 0, clip, width=\linewidth]{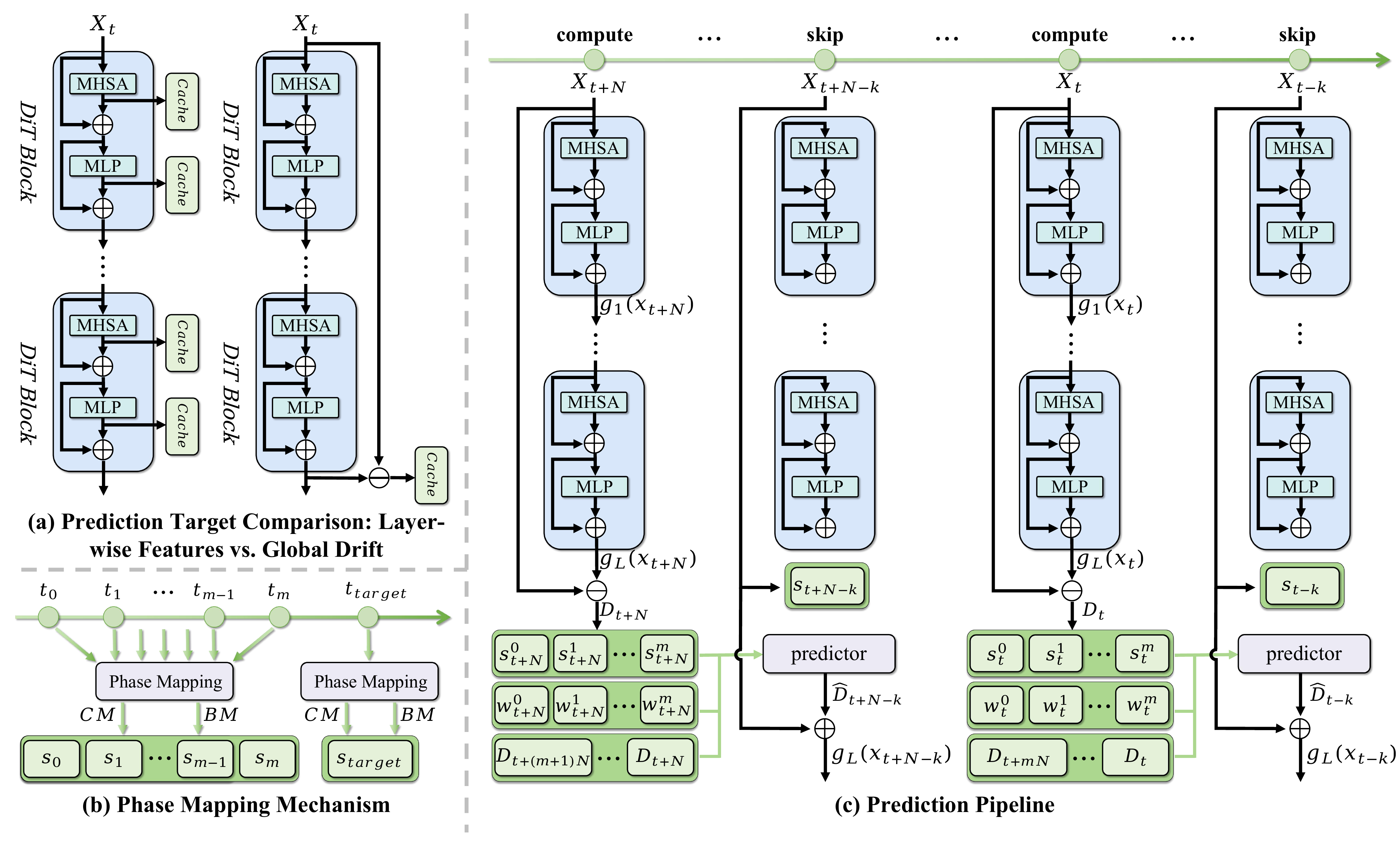}
\vspace{-8mm}
\caption{The ResilPhase acceleration framework. (a) We shift from layer-wise features to the ODE-aligned Global Drift (GD), the end-to-end state displacement. (b) Phase Mapping non-linearly projects discrete time steps into a bounded phase space via Chebyshev (CM) or Balanced (BM) to suppress numerical instability. (c) The pipeline runs full computations every $N$ steps, while our derivative-free Barycentric Lagrange extrapolator of order $m$ estimates the GD for intermediate steps.}
\vspace{-6mm}
\label{fig:framework}
\end{figure*}

\section{Methodology}
\label{headings}

% \subsection{Preliminaries and Motivations}
\subsection{Preliminaries and the Derivative-Induced Instability}
\label{sec:3.1}
% \textbf{Diffusion Transformers (DiT)} A DiT model functions as a learnable velocity field within a probability flow ODE. Architecturally, it consists of a stack of $L$ Transformer blocks. We denote the transformation of the $l$-th block as $g_l$, where $l \in \{1,\dots,L\}$. For the input latent state $\mathbf{x}_t$ at timestep $t$ and condition $c$, the end-to-end mapping is a composition of these blocks: $G(\mathbf{x}_t, c) = g_L(g_{L-1}(\dots g_1(\mathbf{x}_t, c) \dots))$.
\textbf{Diffusion Transformers (DiT)} A DiT model functions as a learnable velocity field in a probability flow ODE. It consists of a stack of $L$ Transformer blocks. We denote the transformation of the $l$-th block as $g_l$, where $l \in \{1,\dots,L\}$. For the input latent state $\mathbf{x}_t$ at timestep $t$ and condition $c$, the end-to-end mapping is a composition of these blocks: $G(\mathbf{x}_t, c) = g_L(g_{L-1}(\dots g_1(\mathbf{x}_t, c)\dots))$. Thus, the input $(\mathbf{x}_t, c)$ is transformed into the final output velocity (or noise prediction) through this complete process $G$.

\textbf{Predictive Caching for Acceleration.} To reduce inference latency, predictive caching uses a sparse schedule. Given an acceleration rate $N$, full forward passes occur only at anchor timesteps (e.g., $t, t - N, \dots$). For intermediate steps, computations are skipped and features are estimated via a lightweight predictor.

Conventionally, this prediction is layer-wise. Let $F(x_t^l)$ denote the feature output of layer $l$ at timestep $t$. Existing methods (e.g., TaylorSeer~\cite{liu2025reusing}, SpeCa~\cite{liu2025speca}) assume smooth feature trajectories and employ polynomial expansions (like Taylor series) for forecasting:
\vspace{-0.2cm}
\begin{equation}
\label{eq:taylor_expansion}
\vspace{-0.2cm}
F_{\text{pred},m}(x_{t-k}^l) = F(x_t^l) + \sum_{i=1}^{m} \frac{\Delta^i F(x_t^l)}{i!} (-k)^i,
\end{equation} 
where $\Delta^i F$ is the $i$-th order finite difference derivative approximation.

\textbf{The Derivative-Induced Instability.} Applying Equation~\eqref{eq:taylor_expansion} to discrete diffusion dynamics is inherently unstable. While feature values are smooth, their finite-difference derivatives ($\Delta^i F$) are dominated by chaos (Fig.~\ref{fig:pca}). Because finite differences act as high-pass filters that exponentially amplify noise, derivative-based solvers (including Hermite splines~\cite{feng2025hicache}) inherently destabilize trajectory reconstruction. This necessitates our derivative-free paradigm.

\subsection{The ResilPhase Framework}

To bypass derivative instability, we adopt a derivative-free approach. For $m + 1$ historical data points $\{(t_0, F_0), \dots, (t_m, F_m)\}$, the unique degree-$m$ interpolation polynomial $P(t)$ is:
\begin{equation}
\label{eq:3}
    P(t) = \sum_{j=0}^{m} F_j L_j(t).
\end{equation}
ResilPhase (Fig.~\ref{fig:framework}) makes two choices. First, the target $F_j$ is the Global Drift, replacing erratic layer-wise features with a smooth macro-trajectory. Second, we stabilize the basis $L_j$ via Barycentric Interpolation and Phase Mapping. This enables accurate extrapolation every $N$ steps without gradient noise.

% To strictly bypass the derivative instability identified above, we ground our framework in a derivative-free approach. Given $m+1$ historical data points $\{(t_0, F_0), \dots, (t_m, F_m)\}$, there exists a unique interpolation polynomial $P(t)$ of degree at most $m$:

% \begin{equation}
% \label{eq:3}
% P(t) = \sum_{j=0}^m F_j L_j(t).
% \end{equation}

% Our ResilPhase framework (illustrated in Fig.~\ref{fig:framework}) instantiates this general form with two mathematically coupled choices. First, we define the target value $F_j$ as the Global Drift , replacing erratic layer-wise features with the smooth macro-trajectory. Second, to stabilize the basis polynomials $L_j$, we employ Barycentric Interpolation combined with a Phase Mapping mechanism. This design enables accurate extrapolation of intermediate states every $N$ steps without incurring gradient noise.

\subsection{ODE-Aligned Macro-Trajectory Targeting}
Previous caching-based acceleration methods rely on layer-wise prediction. However, from a dynamic systems perspective, forcing polynomials to fit highly non-linear, erratic micro-features within individual Transformer blocks causes prediction errors to cascade across the network depth.

\textbf{The Spatial Cascading Error of Layer-wise Forecasting.} Let $x_t^{l-1}$ be the latent input to the $l$-th Transformer block at timestep $t$. In a DiT block, the exact forward computation updates the hidden state via residual connections across the self-attention ($f_{\text{attn},l}$) and MLP ($f_{\text{mlp},l}$) modules. For mathematical simplicity, we group the sum of these residual updates at layer $l$ into a single function $f_l$, meaning the exact transformation is $x_t^l = x_t^{l-1} + f_l(x_t^{l-1})$. During a skipped step, layer-wise methods must approximate these internal updates via a polynomial predictor $\mathcal{P}_l$, yielding the estimated feature:
\begin{equation}
    \hat{x}_t^l = \hat{x}_t^{l-1} + \mathcal{P}_l(\hat{x}_t^{l-1}).
\end{equation}

% Let $e_l = \| \mathcal{P}_l(\hat{x}_t^{l-1}) - f_l(\hat{x}_t^{l-1}) \|$ denote the local prediction error introduced by the polynomial at layer $l$, and let $E_l = \| \hat{x}_t^l - x_t^l \|$ represent the total accumulated state error up to layer $l$.Assuming the transformation $f_l$ is Lipschitz continuous with a constant $L_f$, we can bound the accumulated error using the triangle inequality:
Let $e_l = \| \mathcal{P}_l(\hat{x}_t^{l-1}) - f_l(\hat{x}_t^{l-1}) \|$ and $E_l = \| \hat{x}_t^l - x_t^l \|$ denote the local prediction error and the total accumulated state error up to layer $l$, respectively. Assuming the transformation $f_l$ is Lipschitz continuous with constant $L_f$, we bound the accumulated error via the triangle inequality:
\begin{equation}
\begin{aligned}
    E_l &= ||(\hat{x}_t^{l-1} + \mathcal{P}_l(\hat{x}_t^{l-1})) - (x_t^{l-1} + f_l(x_t^{l-1}))|| \\
    &\leq ||\hat{x}_t^{l-1} - x_t^{l-1}|| + ||\mathcal{P}_l(\hat{x}_t^{l-1}) - f_l(\hat{x}_t^{l-1})|| + ||f_l(\hat{x}_t^{l-1}) - f_l(x_t^{l-1})|| \\
    &\leq E_{l-1} + e_l + L_f E_{l-1} = (1 + L_f)E_{l-1} + e_l.
\end{aligned}
\end{equation}

Solving this recurrence relation from the first layer $l = 1$ to the final layer $L$ (noting that the input is exact, thus $E_0 = 0$), the total spatial cascading error is bounded by:
\begin{equation}
    E_L \leq \sum_{l=1}^L (1 + L_f)^{L-l} e_l.
\end{equation}

This rigorous mathematical bound reveals a fundamental flaw in existing paradigms: spatial prediction errors amplify exponentially across the network depth $L$. Fitting high-frequency micro-features at each intermediate layer merely exacerbates the local error $e_l$, driving the total divergence $E_L$ uncontrollably high at aggressive acceleration ratios.

\textbf{Formulating the Global Drift.} To fundamentally sever this spatial error accumulation, we must elevate the prediction objective from intermediate micro-features to the macroscopic evolution of the ODE. We define the Global Drift (GD), denoted as $D(x_t)$, as the end-to-end state displacement between the model's final output $G(x_t)$ and its initial input $x_t$:
\begin{equation}
    D(x_t) = G(x_t) - x_t.
\end{equation}

Instead of predicting $L$ distinct layer updates, a single macro-predictor $\mathcal{P}_{\text{macro}}$ forecasts this trajectory during skipped steps, yielding $\hat{D}(x_t)$. The final output is reconstructed via addition: $\hat{G}(x_t) = x_t + \hat{D}(x_t)$.
% Rather than predicting $L$ distinct layer updates, our framework uses a single macro-predictor $\mathcal{P}_{\text{macro}}$ to forecast this trajectory during skipped steps, yielding $\hat{D}(x_t)$. The final output is then reconstructed via simple addition:
% \begin{equation}
%     \hat{G}(x_t) = x_t + \hat{D}(x_t)
% \end{equation}

Mathematically, the approximation error of our macro-trajectory framework is strictly determined by a single term:
\begin{equation}
    E_{\text{macro}} = ||\hat{G}(x_t) - G(x_t)|| = ||\hat{D}(x_t) - D(x_t)|| = e_{\text{macro}}.
\end{equation}

By treating the entire DiT stack as a unified probability flow step, we completely bypass the cascading recurrence relation. The error bound is decoupled from the network depth $L$, successfully reducing the spatial error accumulation from $O((1 + L_f)^L)$ to $O(1)$.

% While formulating the GD successfully eliminates spatial error amplification, forecasting this macro-trajectory over time using conventional derivative-based solvers remains vulnerable to the derivative chaos highlighted in Section~\ref{sec:3.1}. This necessitates a robust, derivative-free extrapolation framework, which we introduce next.
While the GD formulation successfully eliminates spatial error amplification, forecasting this macro-trajectory with derivative-based solvers remains vulnerable to the chaotic noise highlighted in Section~\ref{sec:3.1}. This necessitates a robust, derivative-free extrapolation framework, which we introduce next.

\subsection{Derivative-Free Barycentric Interpolation}

To circumvent the catastrophic noise amplification inherent in derivative-based solvers as highlighted in the Derivative Paradox, we strictly ground our framework in a derivative-free approach. We define our target feature value $F_j$ as the fully computed Global Drift $D(x_t)$ at historical timestep $t_j$. A straightforward yet effective formulation of the interpolation polynomial $P(t)$ relies on the classical Lagrange basis polynomials $L_j$, where
\begin{equation}
    L_j(t) = \prod_{k=0,k \neq j}^m \frac{t - t_k}{t_j - t_k}.
\end{equation}

However, the standard Lagrange formula is computationally expensive with a computational complexity of $O(m^2)$. To address this issue, we adopt a more stable and efficient variant: the Barycentric Lagrange Interpolation Formula:
\begin{equation}
    P(t) = \frac{\sum_{j=0}^m \frac{w_j}{t-t_j} F_j}{\sum_{j=0}^m \frac{w_j}{t-t_j}},
\end{equation}
\begin{equation}
    \text{where} \quad w_j = \frac{1}{\prod_{k=0,k \neq j}^m (t_j - t_k)}.
\end{equation}

In this formulation, the barycentric weights $w_j$ depend only on the relative positions of the interpolation nodes $\{t_j\}$, not the target feature values $\{F_j\}$. This pivotal property allows the weights to be precomputed and cached during the full computation steps, making them readily available for all subsequent predictions. Thus, the complexity can be reduced to $O(m)$.

\subsection{Phase Mapping for Prediction Stability}
While the barycentric extrapolator isolates gradient noise, applying polynomial extrapolation over uniform discrete timesteps triggers severe numerical instability. This classic issue, Runge's phenomenon, causes uncontrollable edge oscillations that degrade fidelity at high acceleration ratios. To suppress this error growth, we introduce a plug-and-play Phase Mapping mechanism (Fig.~\ref{fig:framework}(b)).
% While the derivative-free barycentric extrapolator effectively isolates gradient noise, applying high-degree polynomial extrapolation directly over uniformly spaced, discrete diffusion timesteps inevitably suffers from severe numerical instability. This classic issue, known as Runge's phenomenon, causes uncontrollable oscillations near the edges of the extrapolation interval, degrading fidelity at high acceleration ratios. To strictly suppress this oscillatory error growth, we introduce a plug-and-play Phase Mapping mechanism, conceptually illustrated in Fig.~\ref{fig:framework}(b).

\textbf{The Source of Numerical Instability.} To understand this instability mathematically, we examine the error term for Lagrange interpolation. For a function $F(t)$ that is $m + 1$ times differentiable, there exists a real number $\xi \in [t_{\min}, t_{\max}]$ such that the error $E(t) = F(t) - P(t)$ of its degree-$m$ polynomial interpolation $P(t)$ is:
\begin{equation}
    E(t) = \frac{F^{(m+1)}(\xi)}{(m+1)!} \prod_{j=0}^{m} (t - t_j),
\end{equation}
where $t_j$s are time steps used for interpolation. There are two independent components of this function: $D_{m+1} = \frac{F^{(m+1)}(\xi)}{(m+1)!}$, and $e(t) = \prod_{j=0}^{m} (t - t_j)$. 

Because the $D_{m+1}$ term is an inherent property of the DiT model and cannot be modified, the only way to reduce the total error $E(t)$ is by minimizing the node-dependent term $e(t)$. We achieve this by remapping the uniformly spaced time steps $t_j$ to a new, non-uniform distribution of nodes $s_j$ (i.e., a mapping of $t \to s$). Based on the properties of DiTs, we propose two such mapping techniques suitable for different tasks: Chebyshev Mapping and Balanced Mapping.

\subsubsection{Chebyshev Mapping}
\label{sec:chebyshev_implementation_and_theory}

% \subsection{Chebyshev Mapping}

For a given set of $m+1$ recent, fully computed timesteps $\{t_0, \dots, t_m\}$, we generate the corresponding Chebyshev nodes $\{s_k\}$ as follows:
\begin{equation}
    s_k = \cos\left(\frac{(2k+1)\pi}{2(m+1)}\right), \quad \text{for } k=0, \dots, m.
\end{equation}
% However, this formula only prescribes the discrete set of nodes $\{s_k\}$ for our known timesteps and is not a continuous function applicable to an arbitrary $t_{\text{target}}$. To extrapolate the phase coordinate $s_{\text{target}}$ for a target time step where $t_{\text{target}} < t_m$, we employ a stable linear extrapolation based on the two most recent points:
However, this formula only prescribes the discrete nodes $\{s_k\}$ for known timesteps and is not a continuous function applicable to an arbitrary $t_{\text{target}}$. To extrapolate the phase coordinate $s_{\text{target}}$ for a target step $t_{\text{target}} < t_m$, we employ stable linear extrapolation using the two most recent points:
\begin{equation}
    s_{\text{target}} = s_m + \frac{s_m - s_{m-1}}{t_m - t_{m-1}} (t_{\text{target}} - t_m).
\end{equation}

\subsubsection{Balanced Mapping}

While the fixed structure of Chebyshev nodes effectively bounds the error space, its predetermined nature lacks the flexibility to dynamically adjust to varying temporal distributions encountered during highly complex text-conditioned tasks. To provide a more robust and adaptable phase transformation, we propose a novel, data-driven mechanism called Balanced Mapping.

Unlike Chebyshev mapping, this adaptive strategy process begins by analyzing the current set of $m+1$ fully computed timesteps $\{t_j\}_{j=0}^m$ to compute their mean $\mu_t$ and maximum absolute deviation $d_{\max} = \max_j |t_j - \mu_t|$.
% Unlike Chebyshev mapping, this adaptive strategy process begins by analyzing the current set of $m + 1$ fully computed timesteps $\{t_j\}_{j=0}^m$ to compute their mean $\mu_t$ and maximum absolute deviation $d_{\max}$:
% \begin{equation}
%     d_{\max} = \max_j |t_j - \mu_t|
% \end{equation}

The complete non-linear mapping is then given by:
\begin{equation}
    s = \tanh \left( \alpha \cdot \frac{t - \mu_t}{d_{\max}} \right),
\end{equation}

where $\alpha > 0$ is a configurable hyperparameter. This data-driven transformation dynamically confines mapped coordinates via a hyperbolic tangent function, adapting to the spread of recent timesteps. Although $\alpha$ enables fine-grained control, a default $\alpha=0.55$ consistently ensures robust performance across diverse settings. Ultimately, Balanced Mapping delivers a resilient, plug-and-play extrapolation domain robust to numerical outliers.

% where $\alpha > 0$ is a configurable hyperparameter. This data-driven transformation dynamically adapts to the localized spread of recent timesteps and employs a hyperbolic tangent function to strictly confine the mapped coordinates. As a result, Balanced Mapping provides a highly resilient extrapolation domain that is significantly less sensitive to numerical outliers.While $\alpha$ provides fine-grained control, empirical findings show that a default $\alpha = 0.5$ yields consistently robust results across diverse settings, preserving its plug-and-play utility.

\subsubsection{Error Analysis}
\label{sec:error}

We quantitatively compare the node-dependent error bound, $\max_t |e(t)|$, with and without phase mapping.

\paragraph{Without Phase Mapping.}
Without phase mapping, the maximum value of the interpolation error polynomial is:
\begin{align}
|e_{\text{ori}}(t_{\text{target}})| &= \prod_{j=0}^{m} |t_{\text{target}} - t_j| \nonumber \\
&\le |e_{\text{ori}}(t_m - N + 1)| = \prod_{j=0}^{m} |(m - j + 1)N - 1|.
\end{align}
% \vspace{-0.2cm}
% \begin{equation}
% \vspace{-0.2cm}
% \begin{aligned}
%     |e_{\text{ori}}(t_{\text{target}})| & = \prod_{j=0}^{m}\left|t_{\text{target}} - t_j\right| \\
%     & \leq |e_{\text{ori}}(t_m - N + 1)|\\
%     & = \prod_{j=0}^{m} \left|(m-j+1)N - 1\right|.
% \end{aligned}
% \end{equation}

\paragraph{Chebyshev Mapping.}
With Chebyshev mapping, the corresponding error polynomial is given by:
\vspace{-0.2cm}
\begin{equation}
\vspace{-0.2cm}
    e(s) = \prod_{j=0}^{m}(s - s_j).
\end{equation}
The point of maximum error still corresponds to the physical time $t_{\text{target}} = t_m - N + 1$, which is mapped to the coordinate $s_{\text{target}}$ in the phase space. Although the extrapolated target coordinate $s_{\text{target}}$ slightly exceeds the interval $(-1, 1)$, it maintains a mathematically bounded distance to the historical Chebyshev nodes. By algebraically mapping the physical time differences into the phase space, we derive the rigorous upper bound for the error polynomial's magnitude:
% The point of maximum error still corresponds to the physical time $t_{\text{target}} = t_m - N + 1$, which is mapped to the coordinate $s_{\text{target}}$ in the phase space. Since all Chebyshev nodes $\{s_j\}$ lie within the interval $(-1, 1)$, the distance between any two nodes, $\vert s_i - s_j \vert$, is strictly less than 2. This allows us to derive the upper bound for the error polynomial's magnitude:
\vspace{-0.2cm}
\begin{equation}
\vspace{-0.2cm}
    |e_{\text{cheby}}(s_{\text{target}})| \le \prod_{j=0}^{m} \left|2(m - j + 1) - \frac{2}{N}\right|.
\label{eq:cheby-error}
\end{equation}

\paragraph{Balanced Mapping.} 
Similar to Chebyshev Mapping, we have:
\vspace{-0.2cm}
\begin{equation}
\vspace{-0.2cm}
    |e_{\text{bal}}(s_{\text{target}})| \le \prod_{j=0}^{m} |T_{bal}|,
\label{eq:bal-error}
\end{equation}
where $T_{bal} = \frac{\sinh\left(\frac{2\alpha(N-1)}{mN}\right)}{\cosh(\alpha) \cdot \cosh\left(\alpha \frac{2-mN-2N}{mN}\right)} + 2(m-j)$.

\subsubsection{Comparative Analysis.}
We now compare the three error bounds. Let us define a function $f(N, j, m)$ as the difference between the terms inside the products. The difference between no mapping and Chebyshev Node Mapping is:
\begin{equation}
    f(N, j, m) = (m - j)(N - 2) + N + \frac{2}{N} - 3.
\end{equation}
In our acceleration task, constraints are $N \ge 2$ and $m - j > 0$. As $f(N, j, m)$ increases monotonically with $N \ge 2$, it follows that $f(N, j, m) > 0$. This demonstrates that Chebyshev Mapping significantly reduces the polynomial's error upper bound. A similar analysis shows Balanced Mapping also lowers the error bound versus the no-mapping baseline.

When comparing the two mapping schemes, extensive empirical evaluations reveal a distinct, task-dependent superiority. Specifically, Chebyshev Mapping proves to be highly effective for class-conditional image generation tasks. In contrast, for highly complex, text-conditioned generation tasks such as text-to-image and text-to-video, our novel Balanced Mapping consistently demonstrates a clear advantage in preserving semantic alignment and visual fidelity.

% \subsection{Global Residual Prediction}
% \label{sec:global residual}

% Previous caching-based acceleration methods typically rely on a cumbersome, resource-intensive layer-wise prediction paradigm. For a network with L Transformer blocks, this requires caching and predicting the output of each self-attention and MLP block. This approach is complex to implement and consumes large amounts of VRAM, limiting scalability.

% To address this, we shift our prediction target to the Global Residual, the net change between the model's output and input. This conceptual shift from the layer-wise paradigm is illustrated in Figure~\ref{fig:framework}(a).

% We define this global residual, which serves as our prediction target, as:
% \vspace{-0.2cm}
% \begin{equation}
% \vspace{-0.2cm}
%     R(x_t) = G(x_t) - x_t
% \end{equation}
% Thus, we can directly forecast this single, macro-level residual, denoted as $\hat{R}(x_t)$. The final predicted output is then obtained via a simple additive operation:
% \vspace{-0.2cm}
% \begin{equation}
% \vspace{-0.2cm}
%     \hat{G}(x_t) = x_t + \hat{R}(x_t)
% \end{equation}
% This approach simplifies the pipeline and substantially reduces VRAM usage.

\subsection{Methodology Summary}

In conclusion, ResilPhase integrates three core components to construct a noise-resilient acceleration framework: (1) targeting the ODE-aligned Global Drift to bypass spatial cascading errors; (2) a derivative-free Barycentric Lagrange Interpolator to eliminate temporal gradient noise; and (3) a bounded Phase Mapping mechanism to suppress numerical extrapolation instability.

The final predicted macro-trajectory displacement is formulated as:
\begin{equation}
    \hat{D}(x_{\text{pred}}) = P(s_{\text{pred}}) = \frac{\sum_{j=0}^m \frac{w_j}{s_{\text{pred}} - s_j} D_j}{\sum_{j=0}^m \frac{w_j}{s_{\text{pred}} - s_j}}.
\end{equation}

Ultimately, ResilPhase delivers remarkably stable, high-fidelity extrapolations across diverse generative tasks, maintaining exceptional generation quality even under extreme acceleration regimes.

\section{Experiments}
\subsection{Settings}
\textbf{Model Configurations.} We evaluate four mainstream models using a 50-step sampling schedule for fair comparison. For text-to-image, we use FLUX.1-dev~\cite{labs2025flux} with the Rectified Flow~\cite{liuflow} sampler and SDXL-base-1.0~\cite{podell2023sdxl}. For text-to-video, we use HunyuanVideo-Large~\cite{sun2024hunyuan}. For class-conditional generation, we employ DiT-XL/2 ~\cite{peebles2023scalable} with DDIM~\cite{song2020denoising}. Detailed settings and SDXL analyses are in the Supplementary Material.

\textbf{Evaluation.}
We evaluate across three tasks. For the text-to-image task, we use 200 DrawBench~\cite{saharia2022photorealistic} prompts, assessing quality and alignment with ImageReward~\cite{xu2023imagereward} and CLIP Score~\cite{hessel2021clipscore}. For the text-to-video task, we use 946 prompts, evaluating on 16 core dimensions. For both tasks, we also report PSNR, SSIM~\cite{wang2004image}, and LPIPS~\cite{zhang2018unreasonable} for fidelity against original results. For the class-conditional task, we generate images from 1,000 ImageNet~\cite{russakovsky2015imagenet} categories and evaluate using FID-50k~\cite{heusel2017gans}, sFID, and Inception Score (IS). Further details are in the supplementary material.

\begin{table*}[htbp]
\centering
\vspace{-5mm}
\caption{\textbf{Quantitative comparison of text-to-image generation for FLUX.1-dev. Methods are compared at similar latency acceleration ratios across three tiers of speedups to evaluate generation quality.
}}
\vspace{-3mm}
\setlength\tabcolsep{7.0pt} 
%\belowrulesep=0pt
%\aboverulesep=0pt
  \small
  \resizebox{0.98\textwidth}{!}{
  \begin{tabular}{l | c  c | c | c |c |c|c}
    \toprule
    {\bf Method} &\multicolumn{2}{c|}{\bf Acceleration} &{\bf ImageReward $\uparrow$} &\bf CLIP$\uparrow$ & \multirow{2}{*}{\bf PSNR$\uparrow$} & \multirow{2}{*}{\bf SSIM$\uparrow$} & \multirow{2}{*}{\bf LPIPS$\downarrow$}\\
    \cline{2-3}
    {\bf FLUX.1-dev} & {\bf Latency(s) $\downarrow$} & {\bf Speed $\uparrow$} & \bf DrawBench &\bf Score & & & \\
    \midrule
  
  $\textbf{[dev]: 50 steps}$ 
  %\citep{flux2024}         
                           &  {23.69}  & {1.00$\times$} &  {1.0804}  &{32.711}    & - &  -  & -\\

  \midrule
  
  $\textbf{[dev]: 11 steps}$ 
  %\citep{flux2024}         
                           &  {5.21}  & {4.55$\times$} &  {0.9541}  &{32.485}    & 28.397 &  {0.5939}  & {0.5001}\\ 
  
  $\textbf{FORA}$ $(\mathcal{N}=6)$~\cite{selvaraju2024fora} &  {5.20}   & {4.56$\times$} &{0.8468}    &{32.178}   & 28.230 & 0.5700 & 0.5405\\

  \textbf{TeaCache} $({l}=1.4)$~\cite{liu2025timestep} & 4.92 & $ 4.82\times$ &  0.7850  & {32.588} & 27.954 & 0.3837 &0.8349\\ 

  $\textbf{\texttt{ToCa}}$  $(\mathcal{N}=12,R=90\%)$~\cite{zou2024accelerating} &  {9.19}   & {2.58$\times$} & {0.7284}   &{31.475} & {28.444} & 0.5380 & 0.5768\\

  $\textbf{ClusCa} $ $(\mathcal{N}=12,O=2)$~\cite{zheng2025compute} &  {4.88}   & {4.85$\times$} &   0.4958 & 30.353 & 28.079 & 0.3724 & 0.7340\\

  $\textbf{SpeCa} $  $({\tau_0}=12,{\beta}=0.3)$~\cite{liu2025speca} & 4.96 & 4.78$\times$ & 0.9798 & {32.571} & {28.366} & {0.5567} & {0.5324}\\

  $\textbf{PFDiff} $ $(\mathcal{K}=3,H=3)$~\cite{wang2024pfdiff} & 5.82 & 4.07$\times$ & 1.0386 & \underline{32.816} & \underline{28.671} & \underline{0.6162} & \underline{0.4670} \\

  $\textbf{HiCache} $ $(\mathcal{N}=11,O=2)$~\cite{feng2025hicache} & 5.08 & 4.66$\times$ & 0.8040 & 31.604 & {28.268} & {0.5261} & {0.5955} \\ 

  $\textbf{FreqCa} $ $(\mathcal{N}=6,O=2)$~\cite{liu2025freqca} &  \underline{4.85}   & \underline{4.88$\times$} &   \underline{1.0130} & {32.114} & 28.120 & 0.4023 & 0.6877\\

  $\textbf{TaylorSeer} $ $(\mathcal{N}=11,O=2)$~\cite{liu2025reusing} &  {5.10}   & {4.65$\times$} &   0.6241 & 31.895 & 27.940 & 0.3014 & 0.8012\\
  
  \rowcolor{gray!20}
  
  $\textbf{ResilPhase} $ $(\mathcal{N}=6,O=1)$ &  \textbf{4.77}   & \textbf{4.97$\times$} &  \textbf{1.0258} & \textbf{32.847} & \textbf{29.536} & \textbf{0.6655} & \textbf{0.3834}\\

    \midrule
  % $\textbf{FORA}$ $(\mathcal{N}=4)$ {\textcolor{red}{†}} 
  % \citep{selvaraju2024fora}& \ding{52}  &  {5.43}   & {3.17$\times$} & {967.91}   & {3.84$\times$} & {0.8675}    &{18.560} & 29.975 &  0.7217& 0.3007\\

  $\textbf{[dev]: 13 steps}$ 
  %\citep{flux2024}         
                           &  {6.11}  & {3.88$\times$} &  {0.9821}  &{32.586}    & 28.498 &  {0.6136}  & {0.4682}\\ 
  
  $\textbf{FORA}$ $(\mathcal{N}=5)$~\cite{selvaraju2024fora} &  {5.69}   & {4.16$\times$} &{0.9235}    &{32.370}   & 28.288 & 0.5694 & 0.5156\\

  \textbf{TeaCache} $({l}=1.0)$~\cite{liu2025timestep} & 5.82 & $ 4.07\times$ &  0.8518  & \underline{32.750} & 27.960 & 0.3832 & 0.8253\\ 

  $\textbf{\texttt{ToCa}}$  $(\mathcal{N}=10,R=90\%)$~\cite{zou2024accelerating} &  {10.00}   & {2.37$\times$} & {0.8867}   &{32.038} & {28.601} & 0.5778 & 0.5176\\

  $\textbf{ClusCa} $ $(\mathcal{N}=8,O=2)$~\cite{zheng2025compute} &  {5.92}   & {4.00$\times$} &   {0.9913} & 32.410 & 28.430 & 0.5528 & 0.5310\\

  $\textbf{SpeCa} $  $({\tau_0}=8,{\beta}=0.3)$~\cite{liu2025speca} & 5.84 & 4.06$\times$ & 1.0459 & \underline{32.764} & {28.715} & {0.6176} & {0.4460}\\

  $\textbf{PFDiff} $ $(\mathcal{K}=3,H=2)$~\cite{wang2024pfdiff} & 5.83 & 4.06$\times$ & 1.0442 & 32.731 & \underline{28.923} & \underline{0.6481} & \underline{0.4223} \\

  $\textbf{HiCache} $ $(\mathcal{N}=7,O=2)$~\cite{feng2025hicache} & 5.96 & 3.97$\times$ & 1.0079 & 32.622 & {28.705} & {0.6147} & {0.4542} \\ 

  $\textbf{FreqCa} $ $(\mathcal{N}=5,O=2)$~\cite{liu2025freqca} &  \underline{5.76}   & \underline{4.11$\times$} &   \underline{1.0496} & {32.716} & 28.132 & 0.4101 & 0.6835\\

  $\textbf{TaylorSeer} $ $(\mathcal{N}=7,O=2)$~\cite{liu2025reusing} &  {5.98}   & {3.96$\times$} &   0.9406 & 32.657 & 28.049 & 0.3931 & 0.7119\\
  
  \rowcolor{gray!20}
  
  $\textbf{ResilPhase} $ $(\mathcal{N}=5,O=1)$ &  \textbf{5.68}   & \textbf{4.17$\times$} &  \textbf{1.0591} & \textbf{32.901} & \textbf{29.556} & \textbf{0.7024} & \textbf{0.3342}\\

  \midrule

  $\textbf{[dev]: 15 steps}$ 
  %\citep{flux2024}         
                           &  {7.00}  & {3.38$\times$} &  {1.0044}  &{32.618}    & 28.612 &  0.6345  & 0.4375\\ 

  {$\Delta$-DiT} ($\mathcal{N}=10$) &  {7.60}  & {3.12$\times$} & {0.1164}  &{30.369}    & 28.157 &  0.5587 & 0.6408\\
  
  $\textbf{FORA}$ $(\mathcal{N}=4)$~\cite{selvaraju2024fora} &  {7.50}   & {3.16$\times$} &{0.9791}    &{32.684}   & 28.342 & 0.6078 & 0.4725\\

  \textbf{TeaCache} $({l}=0.7)$~\cite{liu2025timestep} & 7.44 & $ 3.18\times$ &  0.9536  & 32.794 & 27.961 & 0.3813 &0.8176\\ 

  $\textbf{\texttt{ToCa}}$  $(\mathcal{N}=8,R=90\%)$~\cite{zou2024accelerating} &  {10.78}   & {2.20$\times$} & {0.9537}   & {32.490} & 28.786 & 0.5954 & 0.4764\\

  $\textbf{ClusCa} $ $(\mathcal{N}=6,O=2)$~\cite{zheng2025compute} &  {6.77}   & {3.50$\times$} &   1.0429 & {32.795} & 28.813 & 0.6205 & 0.4384\\

  $\textbf{SpeCa} $  $({\tau_0}=4,{\beta}=0.3)$~\cite{liu2025speca} & 6.78 & 3.49$\times$ & 1.0598 & \textbf{33.108} & {29.120} & {0.6662} & {0.3786}\\

  $\textbf{PFDiff} $ $(\mathcal{K}=2,H=1)$~\cite{wang2024pfdiff} & 7.66 & 3.09$\times$ & 1.0566 & 32.989 & \underline{29.557} & \underline{0.7030} & \underline{0.3517} \\

  $\textbf{HiCache} $ $(\mathcal{N}=5,O=2)$~\cite{feng2025hicache} & 7.34 & 3.23$\times$ & 1.0438 & 32.898 & {29.240} & {0.6830} & {0.3568} \\ 

  $\textbf{FreqCa} $ $(\mathcal{N}=4,O=2)$~\cite{liu2025freqca} &  \underline{6.74}   & \underline{3.51$\times$} &   1.0457 & \underline{33.041} & 28.146 & 0.4086 & 0.6850\\

  $\textbf{TaylorSeer} $ $(\mathcal{N}=5,O=2)$~\cite{liu2025reusing} &  {7.41}   & {3.20$\times$} &   \underline{1.0566} & 32.811 & {29.132} & {0.6701} & {0.3757}\\
  
  \rowcolor{gray!20}
  
  $\textbf{ResilPhase} $ $(\mathcal{N}=4,O=1)$ &  \textbf{6.62}   & \textbf{3.58$\times$} &  \textbf{1.0647} & {32.874} & \textbf{30.020} & \textbf{0.7216} & \textbf{0.2989}\\

    \bottomrule
  \end{tabular}}
  
  \label{table:FLUX-Metrics}
{\scriptsize
\begin{itemize}[leftmargin=10pt,topsep=0pt]
    % \item \textbf{Note:} ResilPhase used Balanced Mapping with the following hyperparameters: $\alpha=0.65$ for ($N=6, O=1$), $\alpha=0.45$ for ($N=5, O=1$), and $\alpha=0.6$ for ($N=4, O=1$).
    \item \textbf{Note:} ResilPhase results were obtained utilizing Balanced Mapping.
\end{itemize}
}
\vspace{-10mm}
\end{table*}

\subsection{Text-to-Image Generation}
\label{sec:Text-to-Image Generation}

\begin{wrapfigure}{r}{0.6\textwidth}
\vspace{-10mm} % [调整这里] 缩短图片顶部与上方正文的距离
\centering
\includegraphics[trim= 0 0 0 0, clip, width=\linewidth]{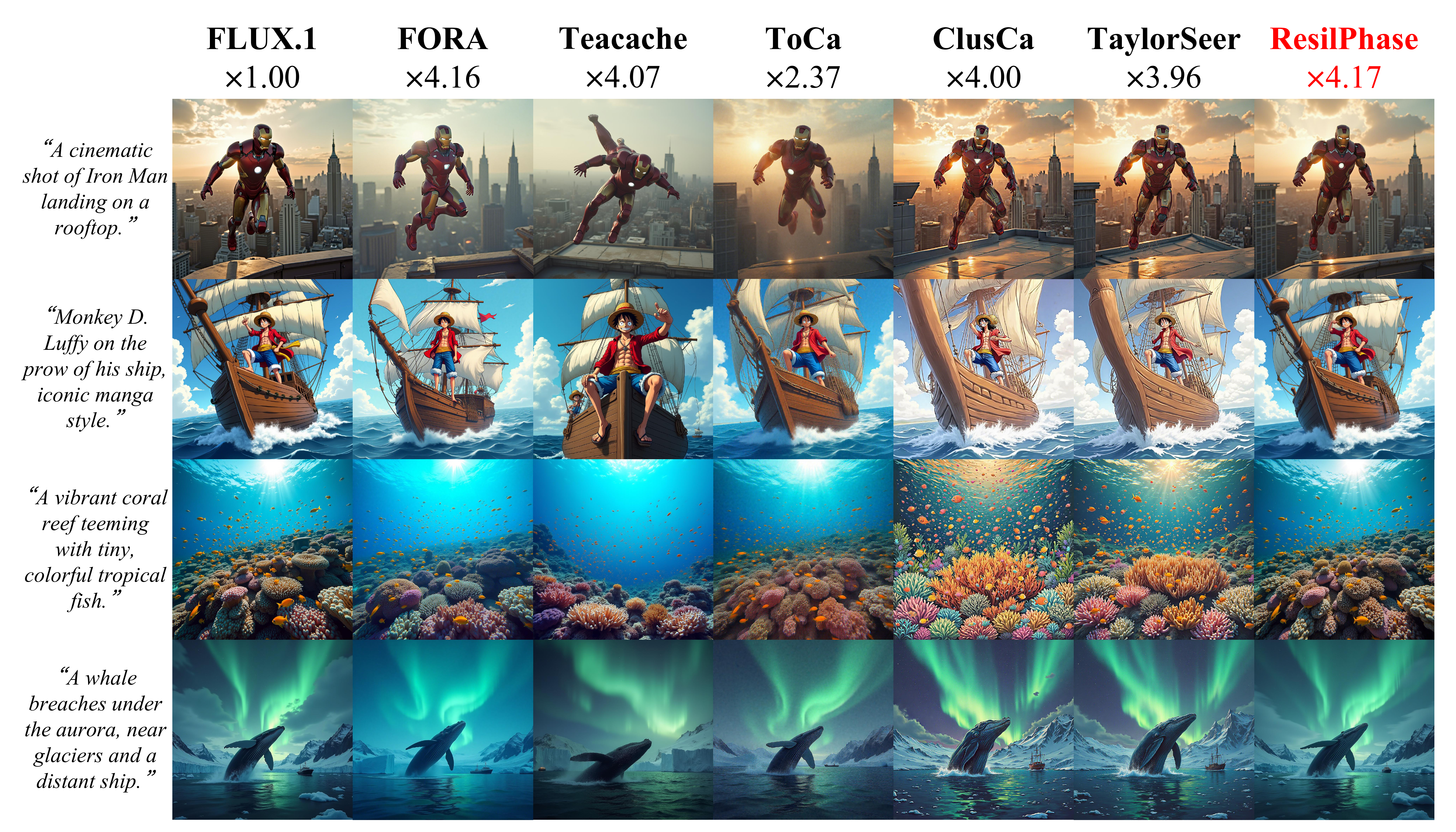}
\vspace{-6mm}  % [调整这里] 如果觉得图片和图注(Caption)之间太空，可以加这个
\caption{ Qualitative comparison on FLUX.1-dev. While baseline methods suffer from visual artifacts and semantic errors at high speedup ratios, ResilPhase preserves both high fidelity and prompt accuracy, showing its superiority.}
\label{fig:FLUX}
\vspace{-8mm} % [调整这里] 缩短图注底部与下方正文的距离
\end{wrapfigure}

As shown in Tab.~\ref{table:FLUX-Metrics}, ResilPhase consistently outperforms competitors in both speed and generation quality, with its advantage widening at higher acceleration ratios. Notably, at an aggressive 4.97x speedup on FLUX.1-dev, ResilPhase maintains an exceptional ImageReward of 1.0258, delivering a $>$ 64\% improvement over the leading forecasting baseline TaylorSeer (0.6241), alongside a 52\% lower LPIPS score. This quantitative robustness translates directly to superior visual fidelity. As visually confirmed in Fig.~\ref{fig:FLUX}, while baseline methods suffer from severe distortions and artifacts at extreme speedups, ResilPhase successfully preserves detailed structures and color accuracy, establishing a new state-of-the-art balance between inference acceleration and exceptional image quality.

\subsection{Text-to-Video Generation}
\begin{figure*}[t]
    \centering
    
    \includegraphics[trim=0 0 0 0, clip, width=\textwidth]{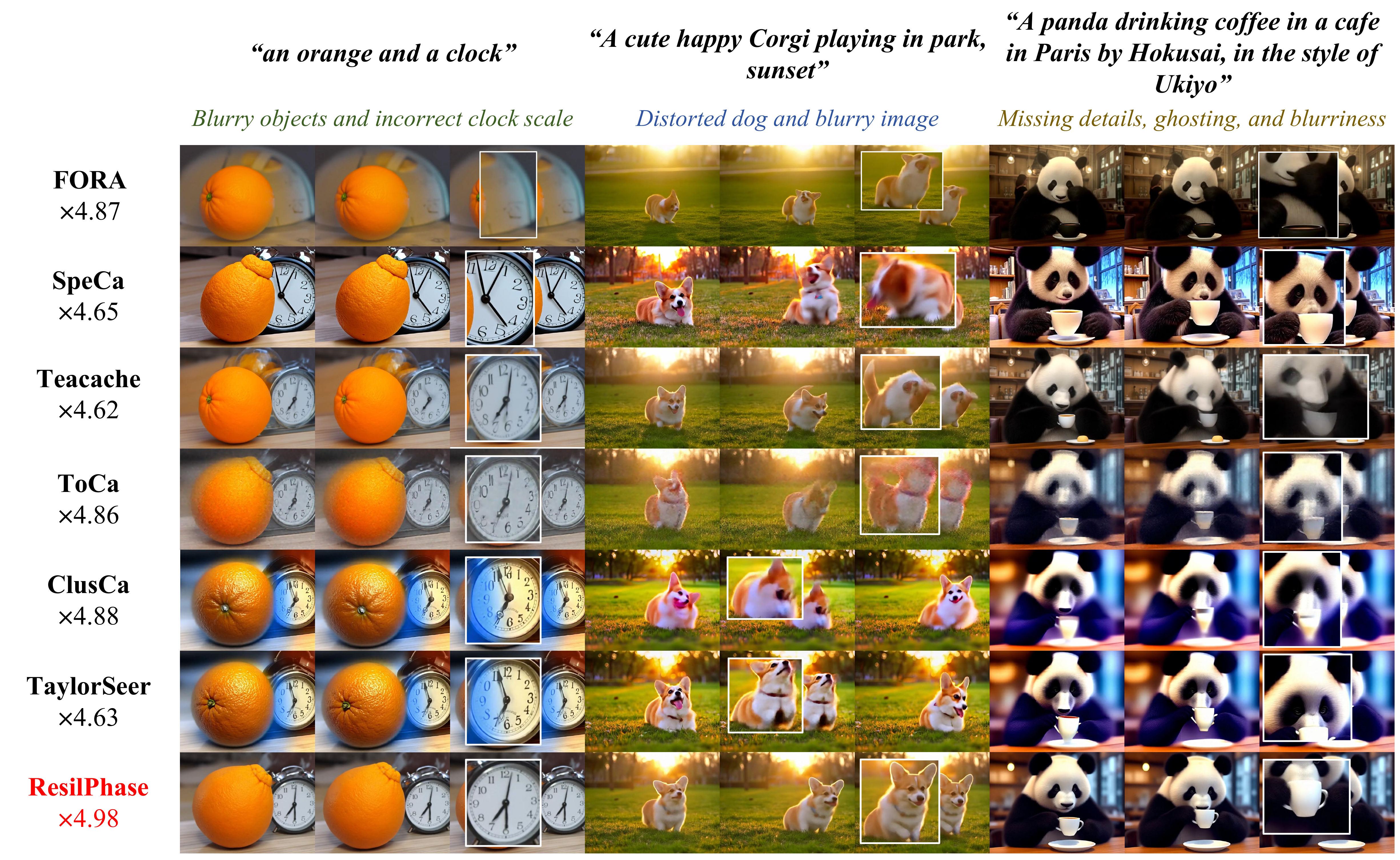}
    \vspace{-6mm}
    \caption{Qualitative comparison of text-to-video generation methods. While competing methods suffer from object distortion, missing details, and motion inconsistency, ResilPhase maintains superior temporal coherence and visual quality.}
    \label{fig:hyvideo}
    \vspace{-3mm}  % 期刊通常不鼓励手动负间距
\end{figure*}

\vspace{-20mm}
As shown in Tab.~\ref{table:HunyuanVideo-Metrics-Compact}, ResilPhase achieves the optimal speed-quality balance across all acceleration levels. At an aggressive $\sim$5$\times$ speedup, it outperforms the SOTA TeaCache with a higher VBench~\cite{huang2024vbench} score (79.78) and a 10\% lower LPIPS, an advantage that widens further at the $\sim$4.3$\times$ tier. These quantitative gains translate directly to superior visual fidelity (Fig.~\ref{fig:hyvideo}). While competing methods suffer from severe artifacts like object blur, incorrect scales, and distorted motions at high speeds, ResilPhase consistently delivers smooth, temporally coherent videos with crisp, semantically accurate details.

\noindent % 消除左侧首行缩进带来的留白
\begin{minipage}[t]{0.52\textwidth}
    \vspace{-0.2cm}
    \setlength\tabcolsep{3.5pt}
    \renewcommand{\cellalign}{l}
    \renewcommand{\theadalign}{l}
    \small
    \captionsetup{justification=raggedright, singlelinecheck=false, aboveskip=2pt, belowskip=0pt, hypcap=false}
    
    % 【关键终极技巧】用 \parbox 固定标题区域高度（1.2cm），确保下方表格绝对水平对齐
    \parbox[t][1.2cm][t]{\linewidth}{
        \captionof{table}{\textbf{Quantitative comparison for HunyuanVideo text-to-video generation, grouped by two tiers of similar latency acceleration ratio.}}
        \label{table:HunyuanVideo-Metrics-Compact}
    }
    
    \vspace{0.45cm} 
    \resizebox{\linewidth}{!}{%
    \begin{tabular}{l cccc c}
        \toprule
        \textbf{Method} & \textbf{Speed $\uparrow$} & \textbf{PSNR $\uparrow$} & \textbf{SSIM $\uparrow$} & \textbf{LPIPS $\downarrow$} & \textbf{VBench $\uparrow$} \\
        \midrule
        \textbf{50-steps} & 1.00$\times$ & - & - & - & 80.87 \\
        \midrule
        $\textbf{FORA}$ $(\mathcal{N}=6)$ & 4.87$\times$ & 15.871 & 0.6073 & 0.4584 & 78.70 \\
        \textbf{TeaCache} $({l}=0.4)$ & 4.62$\times$ & \underline{17.923} & \underline{0.6547} & \underline{0.3760} & \underline{79.77}\\
        $\textbf{\texttt{ToCa}}$  $(\mathcal{N}=10,R=90\%)$ & 4.86$\times$ & 17.581 & 0.5857 & 0.4501 & 76.37  \\
        $\textbf{ClusCa} $ $(\mathcal{N}=9,O=1)$ & \underline{4.88$\times$} & 14.690 & 0.5353 & 0.5139 & 76.98 \\
        $\textbf{SpeCa} $  $({\tau_0}=1.5,{\beta}=0.2)$ & {4.65$\times$} & 16.461 & 0.5883 & 0.4219 & 79.59  \\
        $\textbf{TaylorSeer} $ $(\mathcal{N}=7,O=1)$ & 4.63$\times$ & 15.520 & 0.5641 & 0.4581 & 79.07 \\
        \rowcolor{gray!20}
        $\textbf{ResilPhase} $ $(\mathcal{N}=6,O=1)$ & \textbf{4.98}$\times$ & \textbf{18.920} & \textbf{0.6709} & \textbf{0.3341} & \textbf{79.78} \\
        \midrule
        $\textbf{FORA}$ $(\mathcal{N}=4)$ & 3.57$\times$ & 16.582 & 0.6244 & 0.4010 & 80.10\\
        \textbf{TeaCache} $({l}=0.3)$ & 4.01$\times$ & \underline{19.050} & \underline{0.6846} & \underline{0.3207} & \underline{80.38} \\
        $\textbf{\texttt{ToCa}}$  $(\mathcal{N}=7,R=90\%)$ & {4.09$\times$} & 18.299 & 0.6246 & 0.3688 & 79.00 \\
        $\textbf{ClusCa} $ $(\mathcal{N}=6,O=1)$ & 4.05$\times$ & 16.116 & 0.5881 & 0.4230 & 79.74  \\
        $\textbf{SpeCa} $  $({\tau_0}=1.2,{\beta}=0.1)$ & \underline{4.20$\times$} & 16.488 & 0.5888 & 0.4198 & 79.77  \\
        $\textbf{TaylorSeer} $ $(\mathcal{N}=5,O=1)$ & 3.70$\times$ & 17.117 & 0.6316 & 0.3690 & 80.27 \\
        \rowcolor{gray!20}
        $\textbf{ResilPhase} $ $(\mathcal{N}=5,O=1)$ & \textbf{4.28}$\times$ & \textbf{19.481} & \textbf{0.6992} & \textbf{0.2954} & \textbf{80.42} \\
        \bottomrule
    \end{tabular}%
    }
\end{minipage}%
\hfill
\begin{minipage}[t]{0.45\textwidth}
    \vspace{-0.2cm}
    \small 
    \captionsetup{justification=raggedright, singlelinecheck=false, aboveskip=2pt, belowskip=0pt, hypcap=false}
    
    % 【关键终极技巧】同上，固定为相同的高度 1.2cm
    \parbox[t][1.2cm][t]{\linewidth}{
        \captionof{table}{\textbf{Quantitative comparison for DiT-XL/2 class-to-image generation on ImageNet.}}
        \label{table:DiT_Metrics}
    }
    
    \vspace{0.05cm}
    \resizebox{\linewidth}{!}{%
    \begin{tabular}{@{}lccccc@{}}
        \toprule
        \bf Method & \bf Speed $\uparrow$ & \bf FID $\downarrow$ & \bf sFID $\downarrow$ & \bf IS $\uparrow$ \\
        \midrule
        {\textbf{$\text{DDIM-50 steps}$}} & {1.00$\times$}  &  {2.367} &  {4.387} &  {236.06}\\
        \midrule   
        {\textbf{$\text{DDIM-12 steps}$}} & {4.20$\times$}  &  {8.465} &  {8.664} &  \underline{180.41}\\
        \textbf{FORA} ($\mathcal{N}=8$) & {3.64$\times$}  & {16.694} & {23.396} & {128.08}\\
        \textbf{\texttt{ToCa}} ($\mathcal{N}=12, R = 93\%$)  & {3.53$\times$} & {34.110}   & {30.341} & {85.51} \\
        $\textbf{ClusCa} $ $(\mathcal{N}=12,K=4,O=2)$ & {3.84$\times$} & {15.206}  & {9.405} & {124.03}\\
        $\textbf{SpeCa} $  $({\tau_0}=1.5,{\beta}=0.5)$ &
        \underline{4.36$\times$} & \underline{6.866}  & \underline{8.250} & {171.68}\\
        $\textbf{TaylorSeer}$ $(\mathcal{N}=13,O=2)$ & {2.72$\times$} & {15.415} & {16.005} & {120.76}\\
        \rowcolor{gray!20}
        $\textbf{ResilPhase} $ $(\mathcal{N}=5,O=3)$ & \textbf{4.41$\times$} & \textbf{2.832}  & \textbf{5.026} & \textbf{219.19}\\
        \midrule     
        {\textbf{$\text{DDIM-20 steps}$}} & {2.50$\times$}  &  {3.858} &  {5.201} &  {216.76}\\
        \textbf{FORA} ($\mathcal{N}=4$) & {2.60$\times$}  & {4.770} & {7.591} & {210.98}\\
        \textbf{\texttt{ToCa}} ($\mathcal{N}=5, R = 93\%$)  & \underline{2.67$\times$} & {6.321}   & {7.027} & {196.06}\\
        $\textbf{ClusCa} $ $(\mathcal{N}=6,K=8,O=4)$ & {2.61$\times$} & {3.251}  & {5.207} & {217.60}\\
        $\textbf{SpeCa} $  $({\tau_0}=0.1,{\beta}=0.5)$ &
        {2.65$\times$} & \underline{2.633}  & \underline{4.848} & \underline{231.32}\\
        $\textbf{TaylorSeer}$ $(\mathcal{N}=8,O=3)$ & {2.14$\times$} & {4.807} & {7.088} & {197.22}\\
        \rowcolor{gray!20}
        $\textbf{ResilPhase} $ $(\mathcal{N}=3,O=4)$ & \textbf{2.78$\times$} & \textbf{2.347}  & \textbf{4.672} & \textbf{233.57}\\
        \bottomrule
    \end{tabular}%
    }
\end{minipage}

\subsection{Class-Conditional Image Generation}
On DiT-XL/2, ResilPhase uniquely improves quality while accelerating. At a $2.78\times$ speedup, it achieves an FID of 2.347, outperforming both the full 50-step DDIM baseline (2.367) and the SOTA competitor ClusCa (3.251). Its robustness shines at a $4.41\times$ speedup, maintaining a low FID of 2.832 while most caching baselines collapse (FID $> 15$). Even against the strongest competitor in this tier (SpeCa, FID 6.866), ResilPhase delivers a massive 58\% FID reduction. Consistently leading in sFID and IS, ResilPhase proves highly effective at preserving original quality under aggressive acceleration.

\begin{table*}[htbp]
\centering
\vspace{-4mm}
\caption{\textbf{Ablation study of ResilPhase components on the FLUX-1.dev.}}
\label{table:FLUX-ablation}
\vspace{-4mm}
\setlength\tabcolsep{7.0pt}
\small
\resizebox{\textwidth}{!}{
\begin{tabular}{ c | c  c|c  c|c|c|c|c|c|c}
\toprule
\multirow{2}{*}{\bf Configuration} &  \multicolumn{2}{c|}{\bf Predictive Objective} &  \multicolumn{2}{c|}{\bf Phase Mapping} &  \multirow{2}{*}{\bf Speed $\uparrow$} & {\bf ImageReward $\uparrow$} & \bf CLIP$\uparrow$ & \multirow{2}{*}{\bf PSNR$\uparrow$} & \multirow{2}{*}{\bf SSIM$\uparrow$} & \multirow{2}{*}{\bf LPIPS$\downarrow$}\\
\cline{2-3}
\cline{4-5}

 &  {\bf Global Drift} & {\bf Layer-wise Output} & {\bf Chebyshev} & {\bf Balance} &  & \bf DrawBench & \bf Score & & & \\
\midrule
& & \ding{52} &  & &  {3.08$\times$} &   1.0546 & \underline{32.937} & 29.435 & 0.6948 & 0.3323\\
\rowcolor{gray!10}
& \ding{52} & & & &  {3.57$\times$} &   1.0573 & 32.929 & 29.447 & 0.7003 & 0.3284\\
\textbf{Lagrange} & & \ding{52} & \ding{52} & &  {3.08$\times$} & 1.0559 & \bf{32.975} & 29.436 & 0.6947 & 0.3324 \\ 
\rowcolor{gray!10}
$(\mathcal{N}=4, O=1)$ & & \ding{52} & & \ding{52} &  {3.08$\times$} &   1.0580 & 32.874 & \underline{29.929} & \underline{0.7139} & \underline{0.2995}\\
& \ding{52} &  & \ding{52} & &  {3.58$\times$} &   \underline{1.0595} & 32.914 & 29.448 & 0.7004 & 0.3281\\
\rowcolor{gray!20}
& \ding{52} & & & \ding{52} &  \bf{3.58$\times$} &   \bf{1.0647} & 32.918 & \bf{30.020} & \bf{0.7216} & \bf{0.2989}\\
\midrule[0.8pt] % 分隔线
& & \ding{52} &  & &  {3.64$\times$} &   1.0404 & 32.682 & 29.013 & 0.6659 & 0.3853\\
\rowcolor{gray!10}
& \ding{52} & & & &  {4.17$\times$} &   1.0557 & 32.758 & 29.093 & 0.6786 & 0.3808\\
\textbf{Lagrange} & & \ding{52} & \ding{52} & &   {3.64$\times$} & 1.0409 & 32.655 & 29.014 & 0.6659 & 0.3853  \\
\rowcolor{gray!10}
$(\mathcal{N}=5, O=1)$ & & \ding{52} & & \ding{52} &  {3.64$\times$} &   \underline{1.0528} & \bf{32.998} & \underline{29.472} & \underline{0.6955} & \underline{0.3343}\\
& \ding{52} &  & \ding{52} & &  {4.17$\times$} &   1.0517 & 32.797 & 29.034 & 0.6715 & 0.3811\\
\rowcolor{gray!20}
& \ding{52} & & & \ding{52} &  \bf{4.17$\times$} &   \bf{1.0591} & \underline{32.901} & \bf{29.556} & \bf{0.7024} & \bf{0.3342}\\
\midrule[0.8pt] % 分隔线
& & \ding{52} &  & &  {4.06$\times$} &   {1.0098} & \underline{32.803} & 28.706 & 0.6184 & 0.4479\\
\rowcolor{gray!10}
& \ding{52} & & & &  {4.96$\times$} &   1.0150 & 32.658 & 28.734 & 0.6292 & 0.4388\\
\textbf{Lagrange} & & \ding{52} & \ding{52} & &   {4.06$\times$} & 1.0126 & 32.688 & 28.704 & 0.6183 & 0.4478  \\
\rowcolor{gray!10}
$(\mathcal{N}=6, O=1)$ & & \ding{52} & & \ding{52} &  {4.05$\times$} &   \bf{1.0347} & 32.707 & \underline{29.192} & \underline{0.6526} & \bf{0.3831}\\
& \ding{52} &  & \ding{52} & &  {4.97$\times$} &   1.0186 & 32.708 & 28.734 & 0.6291 & 0.4388\\
\rowcolor{gray!20}
& \ding{52} & & & \ding{52} &  \bf{4.97$\times$} &   \underline{1.0258} & \bf{32.847} & \bf{29.536} & \bf{0.6655} & \underline{0.3834}\\
\bottomrule
\end{tabular}
}
\vspace{-0.5mm}
{\scriptsize
\begin{itemize}[leftmargin=10pt,topsep=0pt]
    \item \textbf{Note:} Lagrange refers to the baseline configuration utilizing only Barycentric Lagrange extrapolation, without any phase mapping.
\end{itemize}
}
\vspace{-1mm}
\end{table*}

\subsection{Ablation Study}
\begin{figure}[htbp]
\vspace{-2em}
\centering  % ✅ 关键：让整个图居中
\begin{subfigure}[b]{\linewidth}
  \centering
  \includegraphics[width=0.96\linewidth]{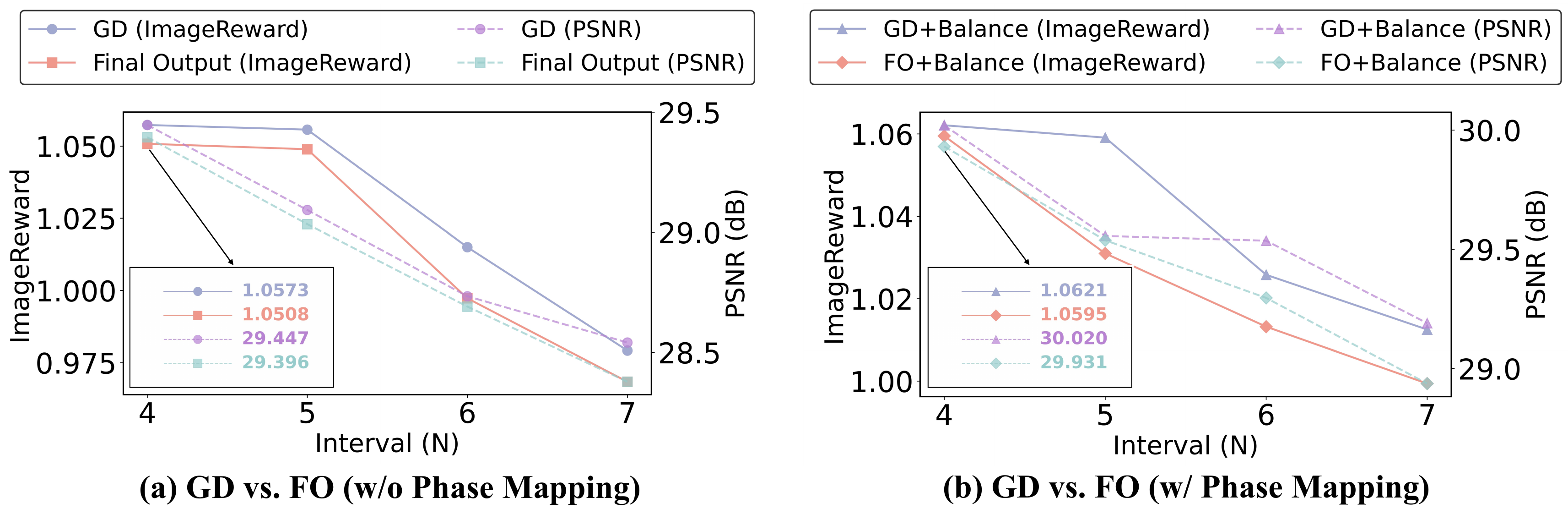}
\end{subfigure}

\vspace{-1em} % 手动调间距（可选）
\caption{Comparing the performance of Global Drift (GD) and Final Output (FO) predictions across extrapolation intervals ($N$) with and without Balance Phase Mapping.}
\label{fig:gd-vs-fo}
\vspace{-6mm}
\end{figure}

\textbf{Ablation Study of ResilPhase Components.} Tab.~\ref{table:FLUX-ablation} shows forecasting Global Drift outperforms layer-wise prediction by eliminating cascading errors. Phase Mapping further boosts quality, with Balanced Mapping proving more suitable for text-to-image tasks than Chebyshev Mapping. We also compared Global Drift against predicting the Final Output, an alternative avoiding intermediate caching. Fig.~\ref{fig:gd-vs-fo} confirms Global Drift consistently maintains superiority over the Final Output across extrapolation intervals.

\begin{figure}[htbp]
\vspace{-2em}
\centering  % ✅ 关键：让整个图居中
\begin{subfigure}[b]{\linewidth}
  \centering
  \includegraphics[width=0.9\linewidth]{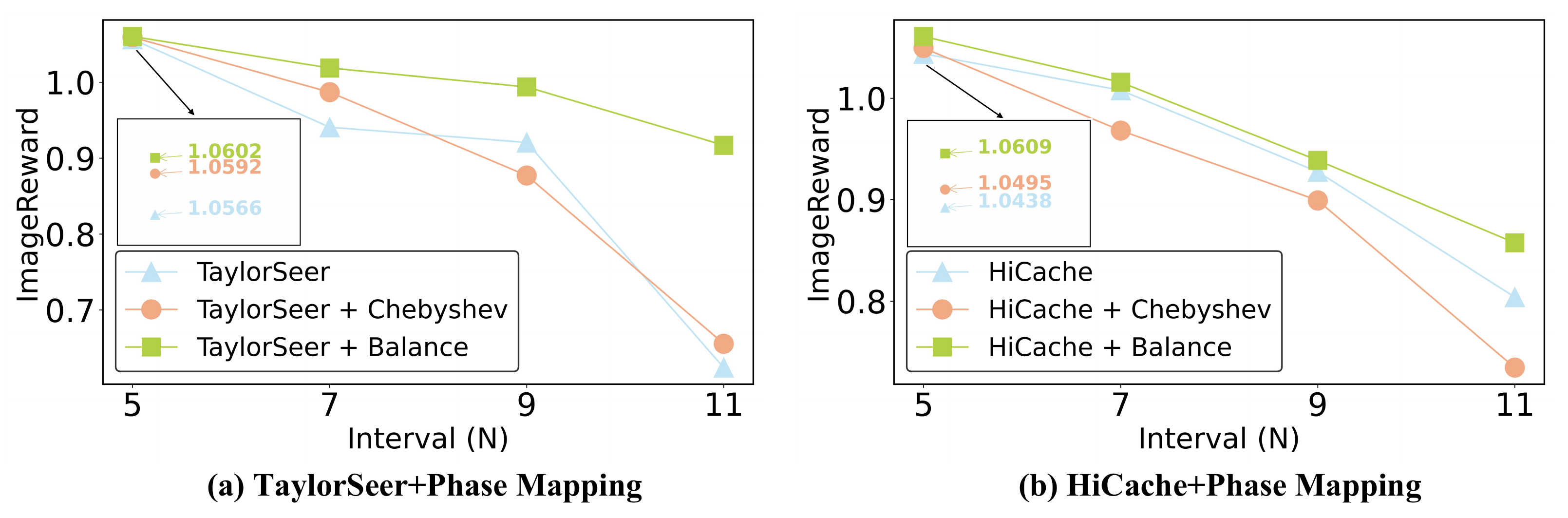}
\end{subfigure}

\vspace{-1em} % 手动调间距（可选）
\caption{Comparing the performance of TaylorSeer and HiCache with different phase mapping strategies on FLUX.1-dev across acceleration intervals.}
\label{fig:taylor-phasemapping}
% \vspace{-10mm}
\end{figure}

\textbf{Generalizability of the Phase Mapping.} Integrating Phase Mapping into TaylorSeer and HiCache consistently boosts ImageReward (Fig.~\ref{fig:taylor-phasemapping}) and other perceptual metrics. Furthermore, our appendix provides theoretical proofs that this mechanism strictly reduces their extrapolation error bounds, which strongly confirms its plug-and-play generalizability for existing polynomial accelerators.
% Integrating Phase Mapping into TaylorSeer and HiCache consistently boosts ImageReward (Fig.~\ref{fig:taylor-phasemapping}) and other perceptual metrics. Furthermore, our appendix provides theoretical proofs that this mechanism strictly reduces their extrapolation error bounds. These empirical and mathematical results strongly confirm its plug-and-play generalizability for existing polynomial accelerators.

\section{Conclusion}
% \vspace{-1mm}
We introduce ResilPhase, a noise-resilient, training-free acceleration framework for Diffusion Transformers. It overcomes spatial, temporal, and numerical bottlenecks in existing paradigms by synergizing an ODE-aligned Global Drift target to eliminate cascading errors, a derivative-free Barycentric Lagrange extrapolator to bypass gradient noise, and a bounded Phase Mapping mechanism to suppress extrapolation instability. Experiments confirm it achieves state-of-the-art speed and quality at aggressive acceleration ratios, establishing a robust real-time inference paradigm and plug-and-play stabilizer for existing accelerators.

% \clearpage\mbox{}Page \thepage\ of the manuscript.
% \clearpage\mbox{}Page \thepage\ of the manuscript.
% \clearpage\mbox{}Page \thepage\ of the manuscript.
% \clearpage\mbox{}Page \thepage\ of the manuscript.

% \clearpage\mbox{}Page \thepage\ of the manuscript. This is the last page.
% \par\vfill\par
% Now we have reached the maximum length of an ECCV \ECCVyear{} submission (excluding references and acknowledgements).
% References should start immediately after the main text, but can continue past p.\ 14 if needed. 
% \clearpage  % TODO FINAL: This \clearpage needs to be removed from both review and camera-ready versions.

\section*{Acknowledgements}
This study is partially supported by the National Natural Science Foundation of China (Grant No.62504204).

% ---- Bibliography ----
%
% BibTeX users should specify bibliography style 'splncs04'.
% References will then be sorted and formatted in the correct style.
%
\bibliographystyle{splncs04}
\bibliography{main}

\clearpage
\appendix

\end{document}

% ---------------------------------------------------------------
% 标题设置：在主标题下方添加 "Supplementary Material" 字样
\title{ResilPhase: Plug-and-Play Phase Mapping and Noise-Resilient Macro-Trajectory Extrapolation for Diffusion Acceleration \\ 
\vspace{0.5em} \Large Supplementary Material} 

\titlerunning{ResilPhase (Supplementary Material)}
\author{}
\authorrunning{}
\institute{}
% ---------------------------------------------------------------
% 作者与机构信息 (盲审阶段保持原样，Camera-ready 时需替换为真实信息)
% \author{First Author\inst{1}\orcidlink{0000-1111-2222-3333} \and
% Second Author\inst{2,3}\orcidlink{1111-2222-3333-4444} \and
% Third Author\inst{3}\orcidlink{2222--3333-4444-5555}}

% \authorrunning{F.~Author et al.}

% \institute{Princeton University, Princeton NJ 08544, USA \and
% Springer Heidelberg, Tiergartenstr.~17, 69121 Heidelberg, Germany
% \email{lncs@springer.com}\\
% \url{http://www.springer.com/gp/computer-science/lncs} \and
% ABC Institute, Rupert-Karls-University Heidelberg, Heidelberg, Germany\\
% \email{\{abc,lncs\}@uni-heidelberg.de}}

\maketitle
\vspace{-1cm}
% ---------------------------------------------------------------
% 正文开始

% 使用 \appendix 将后续的 \section 编号切换为 A, B, C...
\appendix

\section{Experimental Details}

\subsection{Hardware Configuration}
The VBench evaluation for the text-to-video generation task was conducted on four NVIDIA A100 GPUs. All other experiments, including those for text-to-image and class-conditional image generation, were performed on a single NVIDIA A100 GPU.

\subsection{Implementation Details}
For the evaluations of our proposed ResilPhase framework in the main paper, we employed Balanced Mapping for the complex text-to-video generation tasks, while utilizing Chebyshev Mapping for the class-conditional image generation experiments. Regarding the baseline models, in the text-to-image generation experiments, the hyperparameter $K$ (denoting the number of clusters) for the ClusCa baseline was set to 16 for all evaluated configurations reported in Table 1 of the main paper. Furthermore, for the comparative evaluations on FLUX.1-dev, the FreqCa baseline utilized the Discrete Cosine Transform (DCT) for frequency decomposition, strictly following the recommended settings in its original paper.

To ensure optimal computational efficiency, FlashAttention was enabled for all evaluated methods, with the exception of ToCa.

To ensure the reliability and reproducibility of our results, all experiments were conducted using at least three random seeds, with the text-to-video evaluation on VBench specifically utilizing five random seeds. Our framework demonstrates high stability across different runs; the maximum variance across all evaluations was observed in the text-to-image experiments on FLUX.1-dev, which recorded a minimal fluctuation of $32.901 \pm 0.056$ (CLIP Score).

For all quantitative results reported in the tables, we adopt the following formatting convention to facilitate comparison: bold indicates the best performance, \underline{underlined} denotes the second-best result, and \textcolor{blue!70}{blue text} marks the worst performance.

\begin{table*}[htbp]
\centering
\vspace{-2mm}
\caption{\textbf{Additional quantitative comparison against DiCache on FLUX.1-dev.} Methods are compared at similar latency acceleration ratios across three tiers of speedups to evaluate generation quality.}
\vspace{-3mm}
\setlength\tabcolsep{7.0pt} 
%\belowrulesep=0pt
%\aboverulesep=0pt
  \small
  \resizebox{0.98\textwidth}{!}{
  \begin{tabular}{l | c  c | c | c |c |c|c}
    \toprule
    {\bf Method} &\multicolumn{2}{c|}{\bf Acceleration} &{\bf ImageReward $\uparrow$} &\bf CLIP$\uparrow$ & \multirow{2}{*}{\bf PSNR$\uparrow$} & \multirow{2}{*}{\bf SSIM$\uparrow$} & \multirow{2}{*}{\bf LPIPS$\downarrow$}\\
    \cline{2-3}
    {\bf FLUX.1-dev} & {\bf Latency(s) $\downarrow$} & {\bf Speed $\uparrow$} & \bf DrawBench &\bf Score & & & \\
    \midrule
  
  $\textbf{[dev]: 50 steps}$ 
  %\citep{flux2024}         
                           &  {23.69}  & {1.00$\times$} &  {1.0804}  &{32.711}    & - &  -  & -\\

  \midrule
  
  $\textbf{DiCache}$ $(depth=1,th=0.08)$ &  \underline{4.94}   & \underline{4.80$\times$} &\textbf{1.0310}    &\underline{32.065}   & \underline{28.201} & \underline{0.4493} & \underline{0.6742}\\

  \rowcolor{gray!20}
  
  $\textbf{ResilPhase} $ $(\mathcal{N}=6,O=1)$ &  \textbf{4.77}   & \textbf{4.97$\times$} &  \underline{1.0258} & \textbf{32.847} & \textbf{29.536} & \textbf{0.6655} & \textbf{0.3834}\\

    \midrule
  
  $\textbf{DiCache}$ $(depth=1,th=0.06)$ &  \underline{5.74}   & \underline{4.13$\times$} &\underline{1.0288}    &\textbf{33.032}   & \underline{28.247} & \underline{0.4533} & \underline{0.6711}\\

  \rowcolor{gray!20}
  
  $\textbf{ResilPhase} $ $(N=5,O=1)$ &  \textbf{5.68}   & \textbf{4.17$\times$} &  \textbf{1.0591} & \underline{32.901} & \textbf{29.556} & \textbf{0.7024} & \textbf{0.3342}\\

  \midrule
  
  $\textbf{DiCache}$ $(depth=1,th=0.4)$ &  \underline{7.82}   & \underline{3.03$\times$} &\underline{1.0043}    &\textbf{33.239}   & \underline{28.255} & \underline{0.4634} & \underline{0.6682}\\
  
  \rowcolor{gray!20}
  
  $\textbf{ResilPhase} $ $(\mathcal{N}=4,O=1)$ &  \textbf{6.62}   & \textbf{3.58$\times$} &  \textbf{1.0647} & \underline{32.874} & \textbf{30.020} & \textbf{0.7216} & \textbf{0.2989}\\

    \bottomrule
  \end{tabular}}
  
  \label{table:dicache-Metrics}
{\scriptsize
\begin{itemize}[leftmargin=10pt,topsep=0pt]
    % \item \textbf{Note:} ResilPhase used Balanced Mapping with the following hyperparameters: $\alpha=0.65$ for ($N=6, O=1$), $\alpha=0.45$ for ($N=5, O=1$), and $\alpha=0.6$ for ($N=4, O=1$).
    \item \textbf{Note:} For the DiCache baseline, the configurations from top to bottom (across the three acceleration tiers) employ the \texttt{error\_choice} parameter set to \texttt{delta\_minus}, \texttt{delta\_minus}, and \texttt{delta\_y}, respectively.
\end{itemize}
}
\vspace{-6mm}
\end{table*}

\subsection{Extended Evaluation against Recent Baselines on FLUX.1-dev}

\textbf{Quantitative Comparison against DiCache.} Table~\ref{table:dicache-Metrics} presents a dedicated quantitative comparison between our framework and the recently proposed DiCache on the FLUX.1-dev model. Across all three acceleration tiers, ResilPhase consistently achieves higher speedups while delivering vastly superior generation fidelity. Most notably, ResilPhase heavily dominates pixel-level and perceptual metrics. For instance, at the most aggressive tier, ResilPhase reaches a $4.97\times$ speedup with an SSIM of 0.6655 and an LPIPS of 0.3834, decisively outperforming DiCache ($4.80\times$ speedup, SSIM 0.4493, LPIPS 0.6742). Similar staggering margins are observed in the moderate and lower acceleration tiers, where ResilPhase consistently cuts the LPIPS error by half compared to DiCache. While DiCache shows competitive scores on semantic alignment (e.g., CLIP Score), its catastrophic degradation in PSNR, SSIM, and LPIPS strongly indicates severe structural corruption and pixel-level distortion—a finding that is directly corroborated by our visual inspections below.

\begin{figure*}
    \centering
    \includegraphics[trim= 0 0 0 0, clip, width=\linewidth]{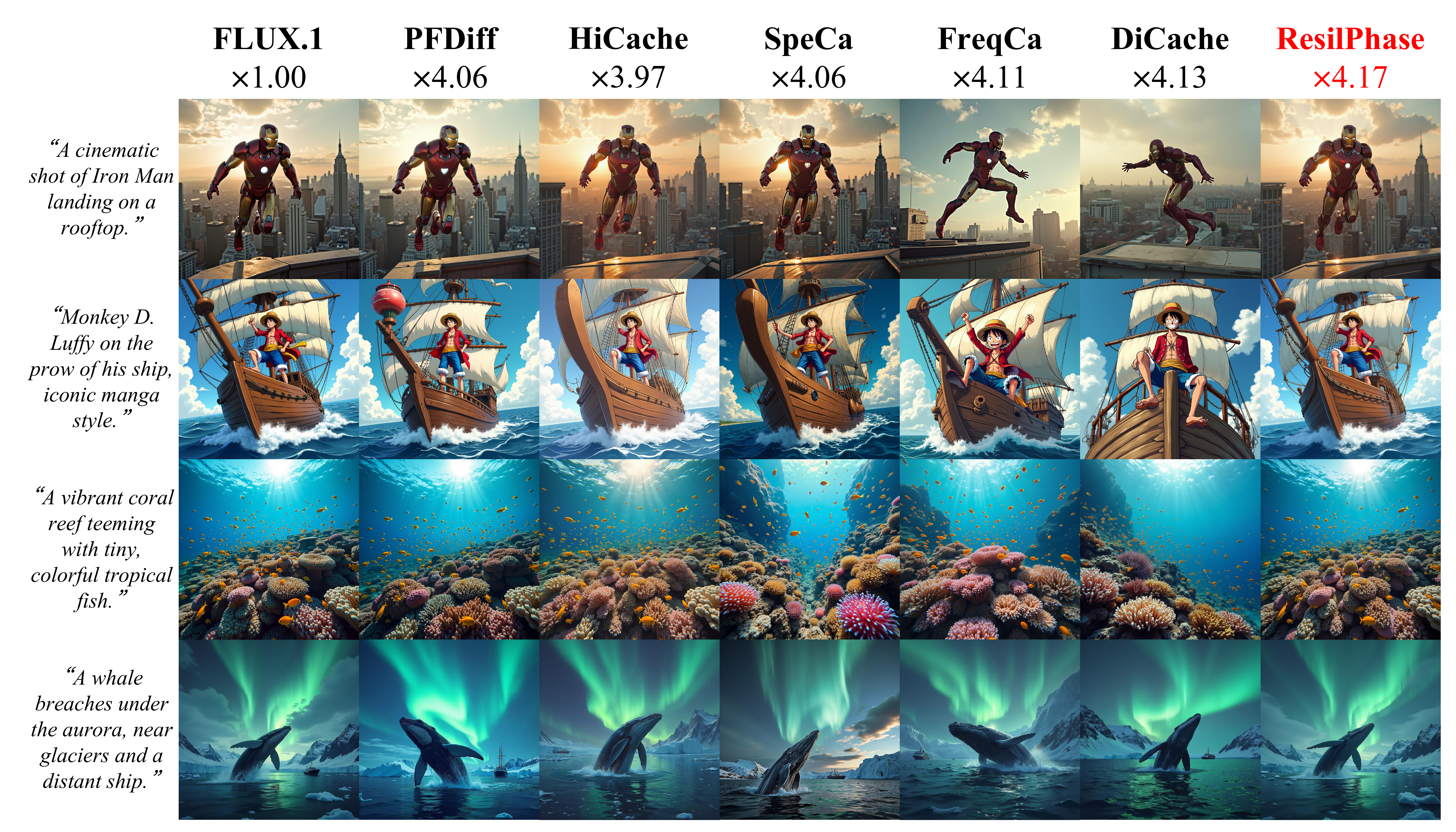}
    \vspace{-6mm}
    \caption{\textbf{Additional qualitative comparison on FLUX.1-dev at the $\sim4.1\times$ acceleration tier.} While recent state-of-the-art baselines (e.g., FreqCa, DiCache, and SpeCa) suffer from severe semantic drift, altered subject poses, and hallucinated artifacts, ResilPhase maintains near-lossless visual fidelity, perfectly preserving the morphology, lighting, and structural details of the unaccelerated original image.}

    \label{fig:FLUX-2}
    \vspace{-3mm}
\end{figure*}

\textbf{Extended Qualitative Analysis.} Figure~\ref{fig:FLUX-2} provides an extended qualitative comparison against recent state-of-the-art acceleration methods (PFDiff, HiCache, SpeCa, FreqCa, and DiCache) on FLUX.1-dev at the $\sim4.1\times$ acceleration tier. The visual results clearly demonstrate that while existing caching and forecasting methods can achieve numerical speedups, they critically fail to maintain semantic consistency and structural integrity. 

Specifically, methods like {FreqCa} and {DiCache} exhibit severe semantic drift and catastrophic structural failure. For instance, in the first row, both methods completely alter Iron Man's landing pose into a horizontal flying or falling posture. Similarly, in the second row, FreqCa drastically distorts the perspective of the pirate ship, while DiCache mutates the sail structure. {SpeCa} and {HiCache} suffer from hallucinated artifacts and lighting inconsistencies, such as adding unnatural red corals (third row) or severely distorting the anatomical shape of the breaching whale into an unrecognizable mass (fourth row). Even {PFDiff}, which performs relatively better, introduces localized errors (e.g., the hallucinated red sphere on Luffy's ship mast and the disappearance of the distant ship in the aurora scene). 

In stark contrast, {ResilPhase} ($\times4.17$) exhibits extraordinary robustness. Despite achieving the highest acceleration ratio in this comparison group, it is the only method that successfully preserves the exact subject poses, complex object morphologies, intricate background details, and accurate global lighting of the unaccelerated $\times1.00$ reference. This visually confirms that predicting the ODE-aligned Global Drift, rather than relying on noisy layer-wise features or unstable frequency decompositions, is essential for high-fidelity diffusion acceleration.

\begin{table*}[htbp]
\centering
\vspace{-3mm}
\caption{\textbf{Quantitative comparison of text-to-image generation for SDXL-base-1.0.} Methods are compared at similar latency acceleration ratios across three tiers of speedups to evaluate generation quality.}
\vspace{-3mm}
\setlength\tabcolsep{7.0pt} 
%\belowrulesep=0pt
%\aboverulesep=0pt
  \small
  \resizebox{0.98\textwidth}{!}{
  \begin{tabular}{l | c  c | c | c |c |c|c}
    \toprule
    {\bf Method} &\multicolumn{2}{c|}{\bf Acceleration} &{\bf ImageReward $\uparrow$} &\bf CLIP$\uparrow$ & \multirow{2}{*}{\bf PSNR$\uparrow$} & \multirow{2}{*}{\bf SSIM$\uparrow$} & \multirow{2}{*}{\bf LPIPS$\downarrow$}\\
    \cline{2-3}
    {\bf stable-diffusion-xl-base-1.0} & {\bf Latency(s) $\downarrow$} & {\bf Speed $\uparrow$} & \bf DrawBench &\bf Score & & & \\
    \midrule
  
  $\textbf{50 steps}$ &  {5.56}   & {1.00$\times$} &{0.5763}    &{33.671}   & - & - & -\\

    \midrule

  \textbf{DeepCache} $(\mathcal{N}=9)$ & 0.94 &  5.91$\times$ &  0.1529  & \underline{32.207} & 28.353 & 0.6478 & 0.4538\\ 

  $\textbf{PFDiff} $ $(\mathcal{K}=10,H=3)$ & 1.01 & 5.50$\times$ & 0.1030 & {31.884} & \underline{28.712} & {0.6447} & {0.4428} \\

  $\textbf{FreqCa\textsuperscript{\textcolor{red}{1}}} $ $(\mathcal{N}=10,O=2)$ &  \underline{0.85}   & \underline{6.54$\times$} &   \underline{0.2915} & {32.079} & 28.628 & \underline{0.6510} & \underline{0.4301}\\

  $\textbf{FreqCa\textsuperscript{\textcolor{red}{2}}} $ $(\mathcal{N}=10,O=2)$ &  {0.87}   & {6.39$\times$} &   \underline{0.2915} & {31.964} & 28.609 & 0.6498 & 0.4367\\

  $\textbf{TaylorSeer} $ $(\mathcal{N}=11,O=2)$ &  {1.80}   & {3.09$\times$} &   0.0538 & 31.199 & 28.163 & 0.5886 & 0.4878\\
  
  \rowcolor{gray!20}
  
  $\textbf{ResilPhase\textsuperscript{\textcolor{red}{1}}} $ $(\mathcal{N}=10,O=1)$ &  \textbf{0.81}   & \textbf{6.86$\times$} &  {0.1305} & {31.549} & {28.181} & {0.5705} & {0.4797}\\
  
  \rowcolor{gray!20}
  
  $\textbf{ResilPhase\textsuperscript{\textcolor{red}{2}}} $ $(\mathcal{N}=10,O=1)$ &  \textbf{0.81}   & \textbf{6.86$\times$} &  \textbf{0.3220} & \textbf{32.391} & \textbf{28.801} & \textbf{0.6516} & \textbf{0.4012}\\

    \midrule
    
  \textbf{DeepCache} $(\mathcal{N}=7)$ & 1.13 &  4.92$\times$ &  0.2890  & {32.523} & 28.514 & 0.6630 & 0.4183\\ 

  $\textbf{PFDiff} $ $(\mathcal{K}=4,H=3)$ & 1.43 & 3.89$\times$ & 0.3890 & \textbf{33.183} & \underline{29.144} & \textbf{0.7135} & \underline{0.3365} \\

  $\textbf{FreqCa\textsuperscript{\textcolor{red}{1}}} $ $(\mathcal{N}=7,O=2)$ &  \underline{1.05}   & \underline{5.30$\times$} &   \underline{0.4357} & {32.713} & 29.007 & 0.6873 & {0.3674}\\

  $\textbf{FreqCa\textsuperscript{\textcolor{red}{2}}} $ $(\mathcal{N}=7,O=2)$ &  {1.08}   & {5.15$\times$} &   {0.3880} & {32.501} & 28.994 & 0.6873 & 0.3731\\

  $\textbf{TaylorSeer} $ $(\mathcal{N}=10,O=2)$ &  {1.82}   & {3.05$\times$} &   0.1668 & 31.629 & 28.207 & 0.5933 & 0.4750\\
  
  \rowcolor{gray!20}
  
  $\textbf{ResilPhase\textsuperscript{\textcolor{red}{1}}} $ $(\mathcal{N}=7,O=1)$ &  \textbf{1.02}   & \textbf{5.45$\times$} &  {0.4235} & {32.642} & {28.563} & {0.6440} & {0.3752}\\
  
  \rowcolor{gray!20}
  
  $\textbf{ResilPhase\textsuperscript{\textcolor{red}{2}}} $ $(\mathcal{N}=7,O=1)$ &  \textbf{1.02}   & \textbf{5.45$\times$} &  \textbf{0.4886} & \underline{32.882} & \textbf{29.260} & \underline{0.6913} & \textbf{0.3253}\\

  \midrule

  \textbf{DeepCache} $(\mathcal{N}=4)$ & {1.61} &  {3.45$\times$} &  0.4189  & {33.230} & 29.223 & 0.7336 & 0.3065\\ 

  $\textbf{PFDiff} $ $(\mathcal{K}=2,H=1)$ & 1.83 & 3.04$\times$ & 0.5535 & \textbf{33.687} & \underline{30.279} & \underline{0.7617} & \underline{0.2440} \\

  $\textbf{FreqCa\textsuperscript{\textcolor{red}{1}}} $ $(\mathcal{N}=4,O=2)$ &  \underline{1.58}   & \underline{3.52$\times$} &   {0.5454} & {33.369} & 29.954 & 0.7526 & {0.2628}\\

  $\textbf{FreqCa\textsuperscript{\textcolor{red}{2}}} $ $(\mathcal{N}=4,O=2)$ &  {1.62}   & {3.43$\times$} &   {0.5130} & {33.425} & 29.954 & 0.7541 & 0.2658\\

  $\textbf{TaylorSeer} $ $(\mathcal{N}=8,O=2)$ &  {1.94}   & {2.87$\times$} &   0.3576 & 32.529 & 28.405 & 0.6267 & 0.4211\\
  
  \rowcolor{gray!20}
  
  $\textbf{ResilPhase\textsuperscript{\textcolor{red}{1}}} $ $(\mathcal{N}=4,O=1)$ &  \textbf{1.55}   & \textbf{3.59$\times$} &  \textbf{0.5789} & {33.488} & {29.858} & {0.7509} & {0.2444}\\
  
  \rowcolor{gray!20}
  
  $\textbf{ResilPhase\textsuperscript{\textcolor{red}{2}}} $ $(\mathcal{N}=4,O=1)$ &  \textbf{1.55}   & \textbf{3.59$\times$} &  \underline{0.5600} & \underline{33.539} & \textbf{30.214} & \textbf{0.7627} & \textbf{0.2290}\\

    \bottomrule
  \end{tabular}}
  
  \label{table:sdxl-Metrics}
{\scriptsize
\begin{itemize}[leftmargin=10pt,topsep=0pt]
    % \item \textbf{Note:} ResilPhase used Balanced Mapping with the following hyperparameters: $\alpha=0.65$ for ($N=6, O=1$), $\alpha=0.45$ for ($N=5, O=1$), and $\alpha=0.6$ for ($N=4, O=1$).
    \item \textbf{Note:} The superscripts denote different method configurations. For FreqCa, FreqCa\textsuperscript{\textcolor{red}{1}} utilizes Fast Fourier Transform (FFT) for frequency decomposition, while FreqCa\textsuperscript{\textcolor{red}{2}} employs Discrete Cosine Transform (DCT). For our proposed method, ResilPhase\textsuperscript{\textcolor{red}{1}} applies Chebyshev Mapping, whereas ResilPhase\textsuperscript{\textcolor{red}{2}} utilizes Balanced Mapping.
\end{itemize}
}
\vspace{-6mm}
\end{table*}

\subsection{Additional Evaluation on SDXL-base-1.0}

\textbf{Quantitative Study.} As presented in Table~\ref{table:sdxl-Metrics}, ResilPhase consistently achieves the optimal trade-off between inference speed and generation quality across all three acceleration tiers. In the most aggressive acceleration regime (the top block), ResilPhase achieves a remarkable $6.86\times$ speedup—the highest among all evaluated methods. Notably, despite this extreme speedup, {ResilPhase$^2$} (employing Balanced Mapping) dominates across almost all quality metrics, securing the best ImageReward (0.3220), PSNR (28.801), SSIM (0.6516), and LPIPS (0.4012). It significantly outperforms the most competitive baseline, FreqCa, which exhibits inferior quality even at lower speedups ($6.39\times \sim 6.54\times$). At moderate acceleration ratios (the $\sim5.45\times$ and $\sim3.59\times$ tiers), ResilPhase\textsuperscript{\textcolor{red}{2}} continues to deliver state-of-the-art performance, particularly excelling in perceptual fidelity (the lowest LPIPS) and human preference alignment (ImageReward). Furthermore, consistent with our findings on FLUX.1-dev, ResilPhase\textsuperscript{\textcolor{red}{2}} generally surpasses ResilPhase$^1$ (Chebyshev Mapping) across most metrics. This corroborates our theoretical analysis that the data-driven Balanced Mapping is more adaptable and better suited for handling the complex, highly non-linear temporal dynamics inherent in text-to-image generation tasks.

\begin{figure*}
    \centering
    \vspace{-3mm}
    \includegraphics[trim= 0 0 0 0, clip, width=\linewidth]{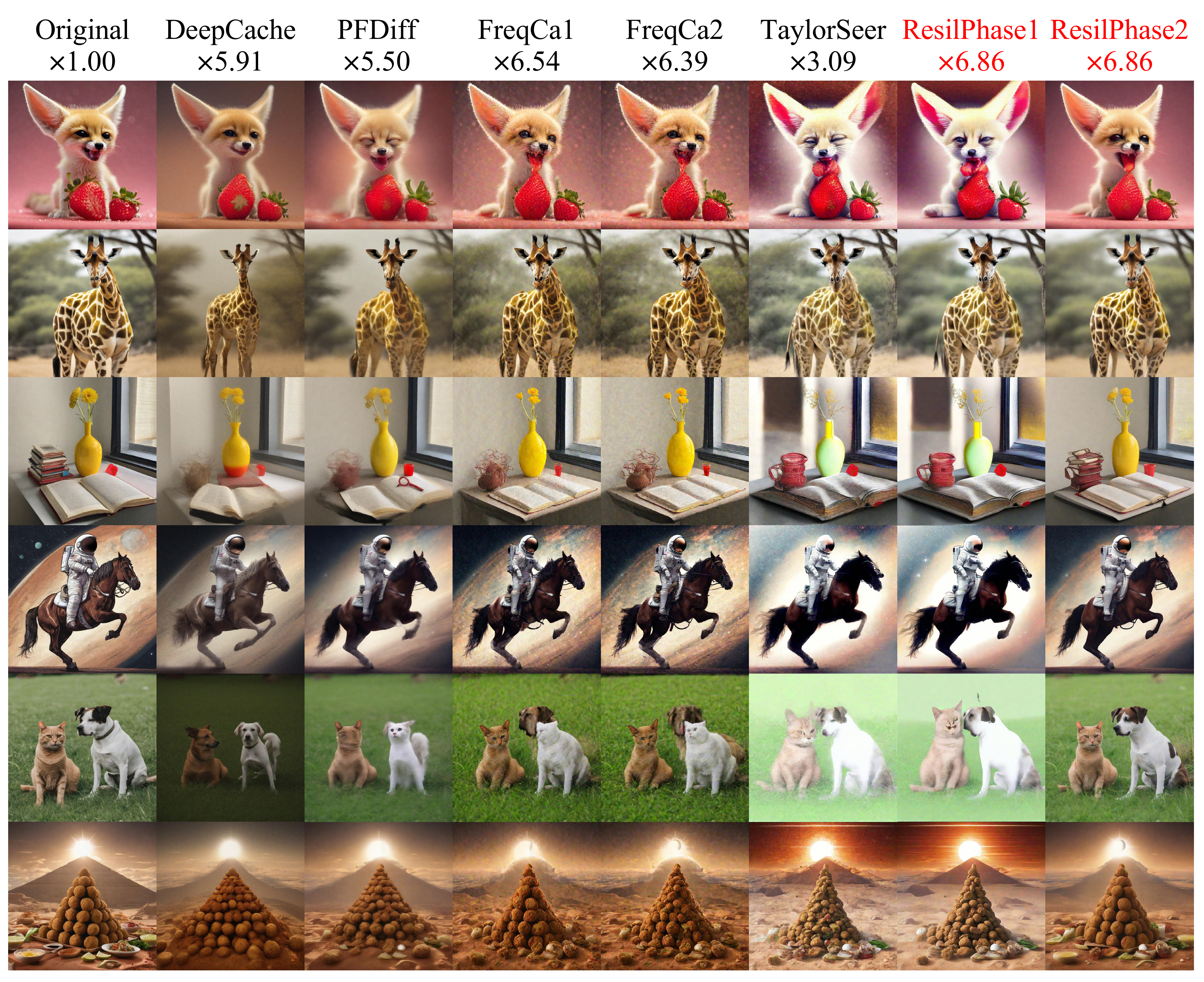}
    \vspace{-6mm}

\caption{\textbf{Qualitative comparison on SDXL-base-1.0 at the highest acceleration tier.} While baselines like DeepCache suffer from darkening and FreqCa/TaylorSeer exhibit severe semantic drift or extreme color shifts at $\sim6.86\times$ speedups, ResilPhase\textsuperscript{\textcolor{red}{2}} successfully maintains superior visual fidelity, accurate lighting, and semantic alignment.}

    \label{fig:sd-fast}
    \vspace{-3mm}
\end{figure*}

\begin{figure*}
    \centering
    \vspace{-2mm}
    \includegraphics[trim= 0 0 0 0, clip, width=\linewidth]{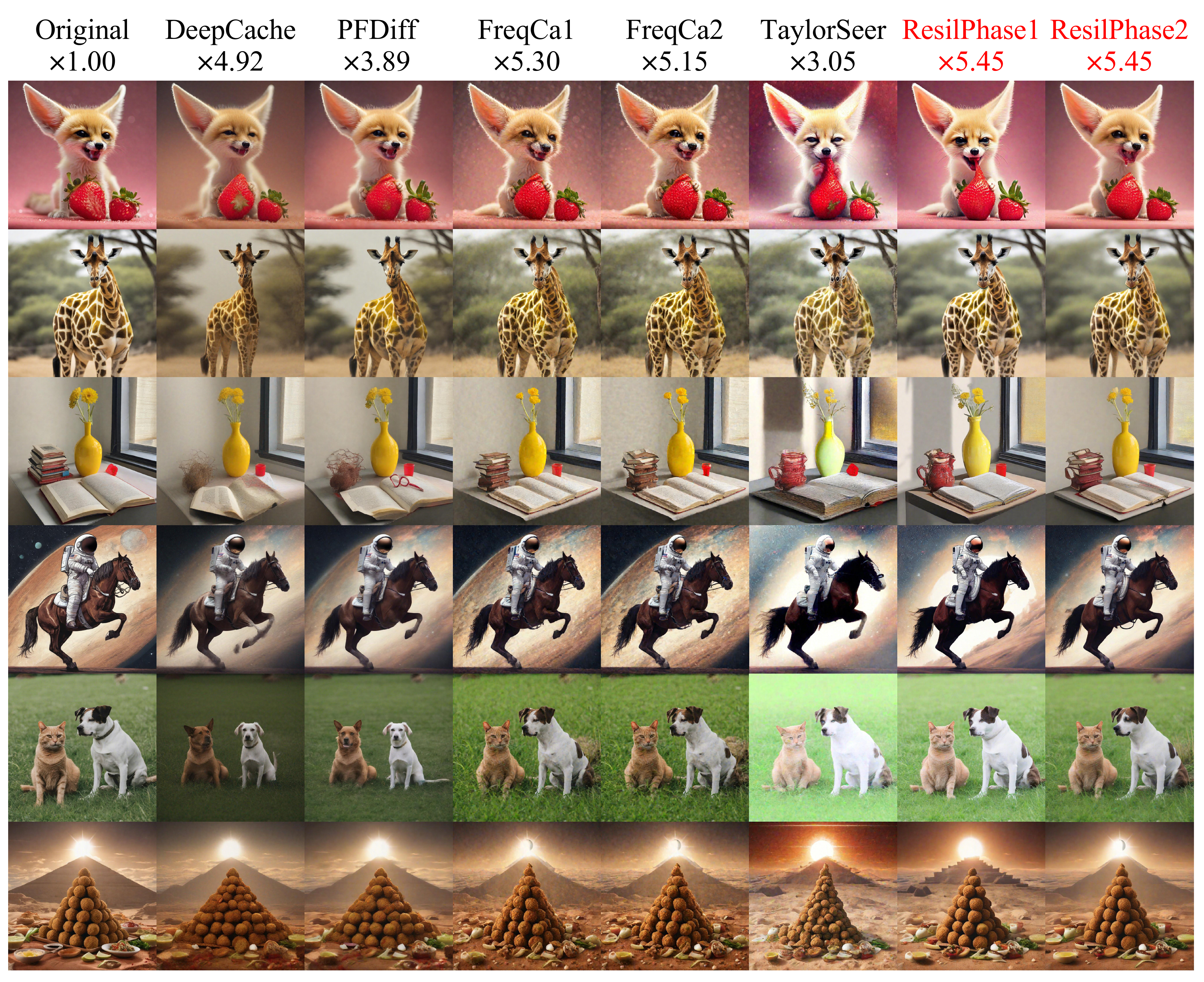}
    \vspace{-6mm}
\caption{\textbf{Qualitative comparison on SDXL-base-1.0 at the second acceleration tier.} Even at a $\sim5.45\times$ speedup, competing methods like TaylorSeer and PFDiff introduce noticeable color over-saturation and artifacts. ResilPhase\textsuperscript{\textcolor{red}{2}} demonstrates robust zero-shot generalizability, closely preserving the original morphology and color palette of the unaccelerated model.}

    \label{fig:sd-medium}
    \vspace{-6mm}
\end{figure*}

\textbf{Qualitative Study.} Visual comparisons corresponding to the most extreme acceleration tier ($\sim6.86\times$) and the second tier ($\sim5.45\times$) are presented in Figure~\ref{fig:sd-fast} and Figure~\ref{fig:sd-medium}, respectively. The visual results strongly corroborate our quantitative findings. As acceleration ratios increase, baseline methods suffer from severe degradation. Specifically, DeepCache exhibits significant darkening and loss of texture (e.g., the cat and dog scene). PFDiff introduces unnatural visual artifacts, while FreqCa struggles with structural integrity and semantic drift (e.g., hallucinating extra animals or blurring facial features). Notably, derivative-based methods like TaylorSeer suffer from catastrophic color shifts, over-saturation, and contrast blowout, turning the grass unnaturally neon and blowing out the background lighting (e.g., the astronaut and pyramid scenes). 

While ResilPhase\textsuperscript{\textcolor{red}{1}} mitigates some structural errors, it still falls victim to similar color distortion and over-exposure issues as TaylorSeer, confirming that fixed node mappings are less suited for the complex latent distribution of text-to-image tasks. In stark contrast, ResilPhase\textsuperscript{\textcolor{red}{2}} demonstrates extraordinary robustness. Across both the $5.45\times$ and the extreme $6.86\times$ speedups, ResilPhase\textsuperscript{\textcolor{red}{2}} consistently preserves the original color palette, accurate lighting, semantic composition, and high-frequency details. It is the only accelerated method that visually rivals the unaccelerated baseline, effectively eliminating the artifacts and style drifts that plague competing approaches.

\subsection{Additional Evaluation for Text-to-Video Generation}
For our Text-to-Video experiments, we also adopted an evaluation methodology inspired by the approach presented in TeaCache. To this end, we sampled a total of 70 prompts from the T2V-CompBench\cite{sun2024t2v} benchmark for video generation. These prompts were specifically curated to assess the generated videos across seven desired attributes, with 10 prompts being designated for each individual attribute.

\begin{table}[t]
\centering
\caption{\textbf{Quantitative evaluation for HunyuanVideo text-to-video generation on the T2V-CompBench benchmark, grouped by two tiers of similar latency acceleration ratios.}
}
\vspace{-3mm}
\setlength\tabcolsep{1pt} % Adjust column spacing
\small
% \resizebox{1.0\linewidth}{!}{
\begin{tabular}{l |c| c | c | c}
\toprule
\textbf{Method} & \textbf{Speed $\uparrow$} & \textbf{PSNR $\uparrow$} & \textbf{SSIM $\uparrow$} & \textbf{LPIPS $\downarrow$}  \\
\midrule
\textbf{50-steps} & 1.00$\times$ & - & - & - \\

\midrule

$\textbf{FORA}$ $(\mathcal{N}=6)$ & 4.92$\times$ & 15.398 & 0.5752 & 0.4809 \\
\textbf{TeaCache} $({l}=0.4)$ & 4.68$\times$ & {17.263} & {0.6321} & {0.3767} \\
$\textbf{\textbf{ToCa}}$  $(\mathcal{N}=10,R=90\%)$ & 4.86$\times$ & 17.006 & 0.5609 & 0.4405 \\
$\textbf{DiCache}$ $(depth=1,th=1)$ & 4.13$\times$ & \underline{17.952} & \textbf{0.6632} & \underline{0.3657} \\
$\textbf{ClusCa} $ $(\mathcal{N}=9,O=1,K=32)$ & \underline{4.98$\times$} & 14.200 & 0.5330 & 0.5130 \\
$\textbf{SpeCa} $  $({\tau_0}=1.5,{\beta}=0.2)$ & {4.53$\times$} & 15.713 & 0.5560 & 0.4268  \\
$\textbf{TaylorSeer} $ $(\mathcal{N}=7,O=1)$ & 4.71$\times$ & 15.068 & 0.5517 & 0.4533 \\
\rowcolor{gray!20}
$\textbf{ResilPhase} $ $(\mathcal{N}=6,O=1)$ & \textbf{5.04}$\times$ & \textbf{17.981} & {0.6333} & \textbf{0.3507} \\

\midrule

$\textbf{FORA}$ $(\mathcal{N}=4)$ & 3.55$\times$ & 16.069 & 0.5926 & 0.4187 \\
\textbf{TeaCache} $({l}=0.3)$ & 4.04$\times$ & {18.272} & \underline{0.6641} & \underline{0.3181} \\
$\textbf{\textbf{ToCa}}$  $(\mathcal{N}=7,R=90\%)$ & \underline{4.18$\times$} & 17.432 & 0.5893 & 0.3841 \\
$\textbf{DiCache}$ $(depth=1,th=0.5)$ & 4.13$\times$ & \textbf{18.656} & {0.6573} & {0.3207} \\
$\textbf{ClusCa} $ $(\mathcal{N}=6,O=1,K=32)$ & 4.17$\times$ & 15.663 & 0.5701 & 0.4170 \\
$\textbf{SpeCa} $  $({\tau_0}=1.2,{\beta}=0.1)$ & \textbf{4.27$\times$} & 15.760 & 0.5532 & 0.4278  \\
$\textbf{TaylorSeer} $ $(\mathcal{N}=5,O=1)$ & 3.71$\times$ & 16.746 & 0.6132 & 0.3640 \\
\rowcolor{gray!20}
$\textbf{ResilPhase} $ $(\mathcal{N}=5,O=1)$ & \textbf{4.27}$\times$ & \underline{18.487} & \textbf{0.6654} & \textbf{0.3129} \\

\bottomrule
\end{tabular}
\label{table:HunyuanVideo-t2vbench}
\vspace{-1mm}

\vspace{-3mm}
\end{table}

\textbf{Quantitative Study.} The quantitative results from this evaluation, presented in Table~\ref{table:HunyuanVideo-t2vbench}, affirm the consistent superiority of our method. In the higher acceleration tier of approximately $5\times$, ResilPhase ($\mathcal{N}=6, O=1$) not only achieves the top speed ($5.04\times$) but also leads in critical quality metrics, scoring the highest PSNR (17.981) and the best (lowest) LPIPS (0.3507). While the newly evaluated DiCache achieves a higher SSIM (0.6632) in this block, it does so at a drastically slower speed ($4.13\times$), making ResilPhase the clear winner in the extreme acceleration regime. This superiority extends to the second acceleration tier (around $4\times$), where the ResilPhase ($\mathcal{N}=5, O=1$) configuration reaches a top speed of $4.27\times$ and outperforms all competitors in both SSIM (0.6654) and perceptual fidelity (LPIPS of 0.3129). Although DiCache marginally leads in PSNR (18.656) in this tier, ResilPhase remains significantly faster and maintains a better LPIPS. Across both tiers, while TeaCache and DiCache emerge as strong competitors, ResilPhase consistently delivers the most optimal balance of inference speed and perceptual visual quality. Notably, ResilPhase consistently secures the best LPIPS scores across all settings, directly reflecting the crisp, blur-free outputs observed in our visual analysis.

\begin{figure*}
    \centering
    \includegraphics[trim= 0 0 0 0, clip, width=\linewidth]{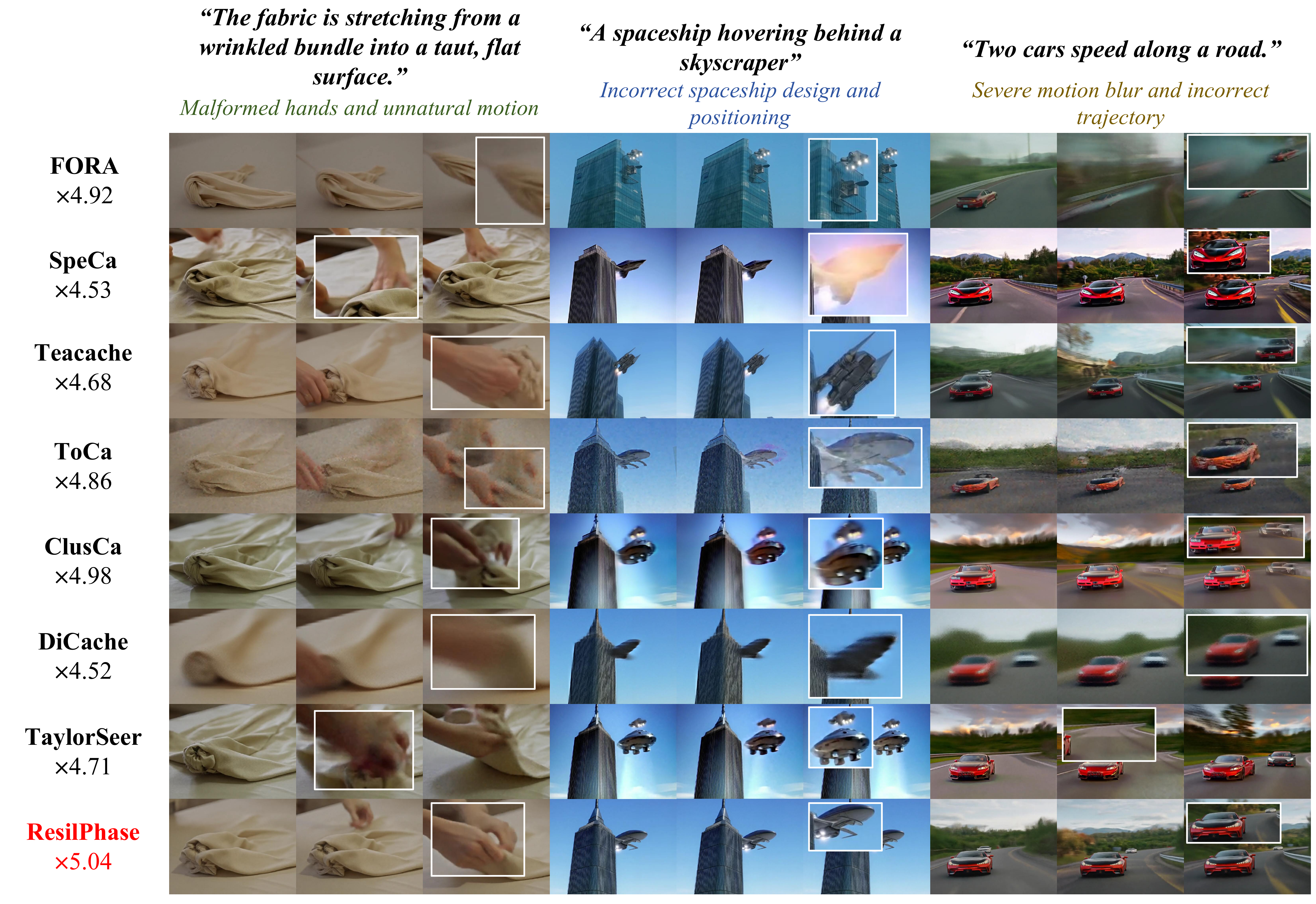}
    \vspace{-6mm}
\caption{Qualitative comparison on T2V-CompBench with HunyuanVideo. While baselines like ToCa and ClusCa exhibit blurring and ghosting at high speedups, ResilPhase maintains superior visual fidelity and semantic accuracy at $5.04\times$ acceleration, avoiding the distortions seen in competing methods.}

    \label{fig:hyvideo-t2v}
    \vspace{-3mm}
\end{figure*}

\textbf{Qualitative Study.} As shown in Figure~\ref{fig:hyvideo-t2v}, a qualitative comparison on challenging prompts reveals the superior temporal coherence and semantic accuracy of ResilPhase. While competing methods consistently produce distorted hands and misinterpret spatial relationships, such as the spaceship's position, they also suffer from severe motion artifacts. Notably, ToCa's outputs suffer from total structural collapse, ClusCa introduces noticeable ghosting, and TaylorSeer exhibits temporal pop-in artifacts. 

It is worth noting that while the recently proposed DiCache avoids the severe anatomical distortions (e.g., malformed hands) and missing entities that plague other baselines, it critically fails to preserve image sharpness. The frames generated by DiCache are pervasively blurry across all prompts, lacking fine-grained textures and high-frequency details---a degradation that is even more pronounced when viewing the accompanying video files in our supplementary materials. In stark contrast, ResilPhase avoids all these pitfalls, generating videos with high fidelity, crisp semantic details, and smooth, coherent motion, even at a $5.04\times$ speedup.

\begin{figure*}
    \centering
    \includegraphics[trim= 0 0 0 0, clip, width=\linewidth]{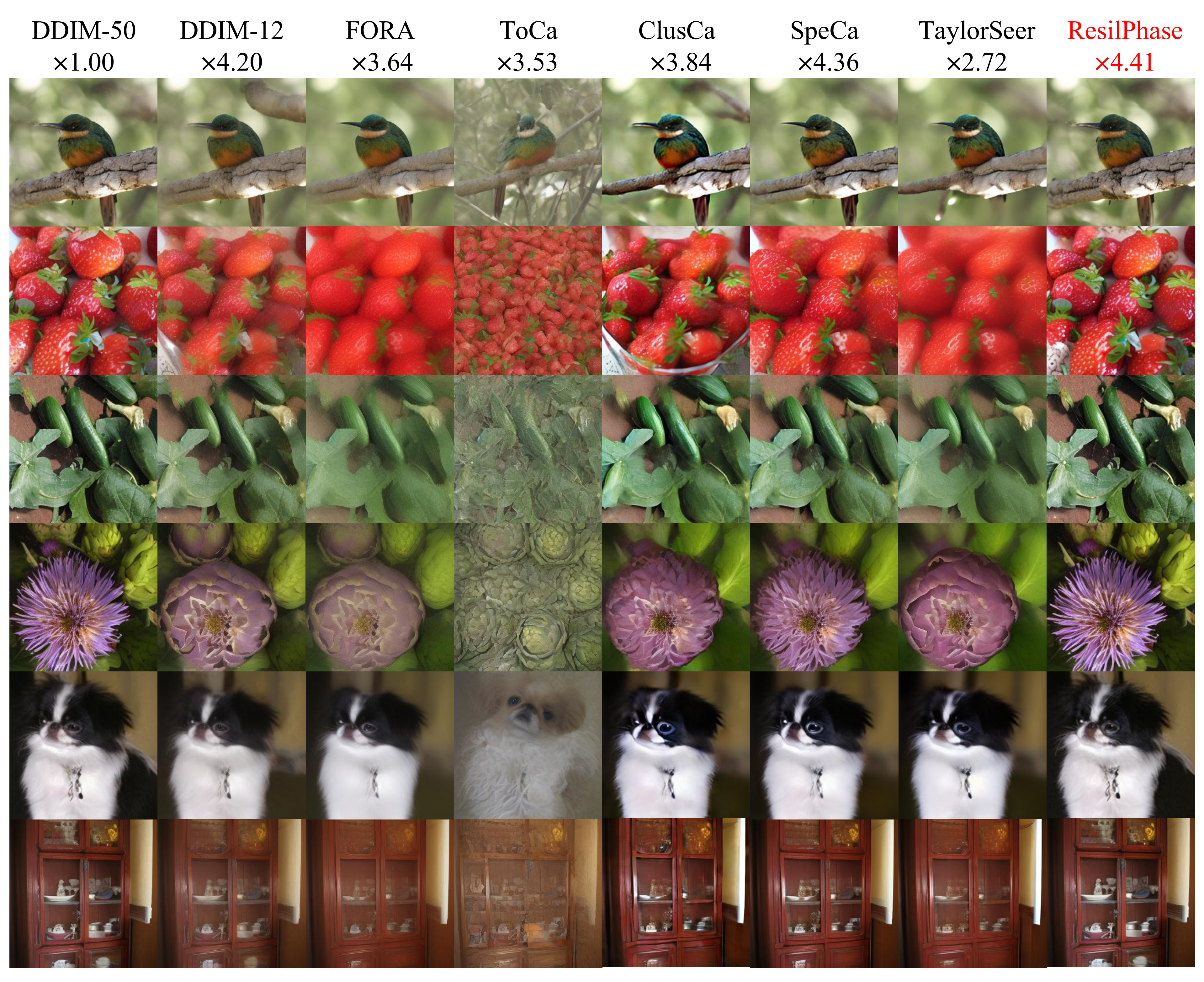}
    \vspace{-6mm}
    \caption{
        Qualitative comparison for class-conditional generation on DiT-XL/2. 
        At a 4.41$\times$ speedup, ResilPhase is the only accelerated method that maintains exceptional fidelity while competitors like TaylorSeer degrade and ToCa collapses. 
        Crucially, its output is the only one to rival the quality of the original 50-step DDIM-50 reference.
    }
    \label{fig:DiT}
    \vspace{-5mm}
\end{figure*}

\subsection{Qualitative Comparison for Class-Conditional Image Generation}

Figure~\ref{fig:DiT} provides an extended qualitative comparison on the class-conditional DiT-XL/2 model. Despite operating at the highest acceleration ratio ($4.41\times$), ResilPhase uniquely maintains exceptional fidelity, closely rivaling the unaccelerated 50-step DDIM reference. 

In contrast, competing methods suffer significant degradation. {ToCa} exhibits a catastrophic loss of structural integrity, producing unrecognizable mosaic noise (e.g., strawberries and flowers). Other baselines maintain basic semantics but lose critical high-frequency details. For instance, {ClusCa} unnaturally darkens shadows and renders fruits with a plastic-like smoothness. {SpeCa} and TaylorSeer (even at a lower $2.72\times$ speedup) severely blur fine textures, smudging strawberry seeds, flower petals, and the dog's fur into soft, undefined patches. ResilPhase robustly avoids these over-smoothing and contrast-shifting artifacts, successfully preserving crisp specular highlights, distinct leaf veins, and accurate lighting, directly corroborating our quantitative findings.

% \ \\ \ \\ \ \\ \ \\ \ \\ \ \\ \ \\ \ \\ \ \\ \ \\ \ \\ 

\section{Analysis of Phase Mapping Schemes and Hyperparameter $\alpha$}
\vspace{-2mm}
To determine the optimal phase mapping strategy for each generative task, we conducted a comprehensive ablation study comparing our two proposed methods: the fixed Chebyshev Mapping and the adaptive Balanced Mapping, whose performance is governed by the hyperparameter $\alpha$. The following analysis across a range of $\alpha$ values validates the specific configurations used in our main experiments and demonstrates how the ideal choice between Chebyshev and Balanced Mapping is contingent on the underlying model architecture and generative task.

The comprehensive evaluation reported in Tables~\ref{table:FLUX-Metrics-alpha-1}--\ref{table:DiT-alpha2} reveals a distinct, task-dependent preference for phase mapping strategies. For Class-Conditional Image Generation, the parameter-free Chebyshev Mapping proves to be the most effective choice, consistently yielding superior or highly competitive FID scores compared to the adaptive approach. In contrast, for Text-to-Image and Text-to-Video generation, Balanced Mapping demonstrates a clear superiority, significantly outperforming the Chebyshev baseline in both alignment and fidelity metrics. Empirically, we identify that the optimal configuration for these complex text-conditioned tasks lies within the robust $\alpha$ range of $[0.35, 0.85]$. Within this wide stable region, ResilPhase achieves peak performance across various acceleration ratios. This extensive analysis directly justifies our default setting of $\alpha = 0.55$ adopted in the main experiments, as it consistently acts as a highly stable and reliable anchor point within this optimal range.

% \clearpage 
\begin{table*}
\centering
\caption{\textbf{Ablation study of Phase Mapping and $\alpha$ scaling on FLUX.1-dev. We compare Chebyshev Mapping with Balanced Mapping, specifically analyzing the influence of the parameter $\alpha$ on the final generation results.
}}
\vspace{-3mm}
\setlength\tabcolsep{7.0pt} 
%\belowrulesep=0pt
%\aboverulesep=0pt
  \small
  \resizebox{1.00\textwidth}{!}{
  \begin{tabular}{c | c | c | c | c |c |c|c}
    \toprule
    \multirow{2}{*}{\bf Configuration} & Phase & \multirow{2}{*}{\bf $\alpha$}  & {\bf ImageReward $\uparrow$} &\bf CLIP$\uparrow$ & \multirow{2}{*}{\bf PSNR$\uparrow$} & \multirow{2}{*}{\bf SSIM$\uparrow$} & \multirow{2}{*}{\bf LPIPS$\downarrow$}\\
    % \cline{2-3}
    & Mapping & & \bf DrawBench &\bf Score & & & \\
    \midrule

\multirow{30}{*}{$\textbf{ResilPhase} $ $(\mathcal{N}=6,O=1)$} & Chebyshev Mapping & -  &  1.0086 & 32.688 & \textcolor{blue!70}{28.734} & \textcolor{blue!70}{0.6291} & \textcolor{blue!70}{0.4388} \\

\cline{2-8}
  
  & Balanced Mapping & 0.05 &  \textcolor{blue!70}{1.0023} & 32.711 & 28.758 & 0.6319 & 0.4346 \\
  & Balanced Mapping & 0.10 &  1.0244 & 32.749 & 28.823 & 0.6381 & 0.4253 \\
  & Balanced Mapping & 0.15 &  1.0177 & 32.715 & 28.913 & 0.6453 & 0.4148 \\
  & Balanced Mapping & 0.20 &  1.0299 & 32.751 & 29.007 & 0.6524 & 0.4034 \\
  & Balanced Mapping & 0.25 &  1.0131 & 32.659 & 29.007 & 0.6572 & 0.3948 \\
  & Balanced Mapping & 0.30 &  1.0140 & 32.701 & 29.139 & 0.6604 & 0.3893 \\
  & Balanced Mapping & 0.35 &  1.0303 & 32.668 & 29.193 & 0.6627 & 0.3854 \\
  & Balanced Mapping & 0.40 &  \textbf{1.0339} & 32.640 & 29.233 & 0.6640 & 0.3839 \\
  & Balanced Mapping & 0.45 &  \underline{1.0319} & 32.640 & 29.268 & 0.6648 & 0.3828 \\
  & Balanced Mapping & 0.50 &  1.0302 & 32.658 & 29.298 & 0.6653 & \underline{0.3823} \\
  & Balanced Mapping & 0.55 &  1.0243 & 32.753 & 29.322 & \textbf{0.6656} & \textbf{0.3819} \\
  & Balanced Mapping & 0.60 &  1.0281 & 32.687 & 29.340 & \textbf{0.6656} & 0.3825 \\
  & Balanced Mapping & 0.65 &  1.0258 & \textbf{32.847} & 29.356 & \underline{0.6655} & 0.3834 \\
  & Balanced Mapping & 0.70 &  1.0243 & \underline{32.838} & 29.367 & 0.6654 & 0.3842 \\
  & Balanced Mapping & 0.75 &  1.0242 & 32.789 & 29.376 & 0.6652 & 0.3851 \\
  & Balanced Mapping & 0.80 &  1.0178 & 32.722 & 29.383 & 0.6649 & 0.3863 \\
  & Balanced Mapping & 0.85 &  1.0263 & 32.716 & \underline{29.387} & 0.6645 & 0.3872 \\
  & Balanced Mapping & 0.90 &  1.0088 & 32.605 & \textbf{29.388} & 0.6638 & 0.3883 \\
  & Balanced Mapping & 0.95 &  1.0079 & 32.631 & \textbf{29.388} & 0.6634 & 0.3894 \\
  & Balanced Mapping & 1.00 &  1.0121 & \textcolor{blue!70}{32.580} & \textbf{29.388} & 0.6628 & 0.3907 \\
  & Balanced Mapping & 1.30 &  1.0063 & 32.557 & 29.376 & 0.6609 & 0.3957 \\
  & Balanced Mapping & 1.50 &  1.0027 & 32.601 & 29.370 & 0.6600 & 0.3978 \\
  & Balanced Mapping & 1.80 &  1.0042 & 32.588 & 29.363 & 0.6590 & 0.3999 \\
  & Balanced Mapping & 2.00 &  1.0027 & 32.536 & 29.360 & 0.6586 & 0.4009 \\
  & Balanced Mapping & 2.50 &  1.0047 & 32.564 & 29.355 & 0.6579 & 0.4023 \\
  & Balanced Mapping & 5.00 &  1.0085 & 32.610 & 29.353 & 0.6576 & 0.4028 \\
  & Balanced Mapping & 10.00 &  1.0049 & 32.516 & 29.353 & 0.6575 & 0.4030 \\
  & Balanced Mapping & 20.00 &  1.0127 & 32.588 & 29.353 & 0.6578 & 0.4028 \\
  & Balanced Mapping & 50.00 &  1.0127 & 32.588 & 29.353 & 0.6578 & 0.4028 \\

    \bottomrule
  \end{tabular}}
  
  \label{table:FLUX-Metrics-alpha-1}
\vspace{-4mm}
\end{table*}
% \clearpage

\ \\ \ \\ \ \\ \ \\ \ \\ 
% \clearpage

\begin{table*}[htbp]
\centering
\caption{\textbf{Ablation study of Phase Mapping and $\alpha$ scaling on FLUX.1-dev. We compare Chebyshev Mapping with Balanced Mapping, specifically analyzing the influence of the parameter $\alpha$ on the final generation results.
}}
\vspace{-3mm}
\setlength\tabcolsep{7.0pt} 
%\belowrulesep=0pt
%\aboverulesep=0pt
  \small
  \resizebox{1.00\textwidth}{!}{
  \begin{tabular}{c | c | c | c | c |c |c|c}
    \toprule
    \multirow{2}{*}{\bf Configuration} & Phase & \multirow{2}{*}{\bf $\alpha$}  & {\bf ImageReward $\uparrow$} &\bf CLIP$\uparrow$ & \multirow{2}{*}{\bf PSNR$\uparrow$} & \multirow{2}{*}{\bf SSIM$\uparrow$} & \multirow{2}{*}{\bf LPIPS$\downarrow$}\\
    % \cline{2-3}
    & Mapping & & \bf DrawBench &\bf Score & & & \\
    \midrule

\multirow{30}{*}{$\textbf{ResilPhase} $ $(\mathcal{N}=5,O=1)$} & Chebyshev Mapping & -  &  1.0517 & \textcolor{blue!70}{32.655} & \textcolor{blue!70}{29.034} & \textcolor{blue!70}{0.6715} & \textcolor{blue!70}{0.3811} \\

\cline{2-8}
  
  & Balanced Mapping & 0.05 &  1.0619 & 32.731 & 29.052 & 0.6736 & 0.3781 \\
  & Balanced Mapping & 0.10 &  \textbf{1.0664} & 32.768 & 29.113 & 0.6787 & 0.3701 \\
  & Balanced Mapping & 0.15 &  1.0704 & 32.904 & 29.193 & 0.6842 & 0.3615 \\
  & Balanced Mapping & 0.20 &  1.0611 & 32.912 & 29.271 & 0.6888 & 0.3539 \\
  & Balanced Mapping & 0.25 &  1.0546 & 32.905 & 29.349 & 0.6930 & 0.3470 \\
  & Balanced Mapping & 0.30 &  \underline{1.0660} & 32.897 & 29.411 & 0.6958 & 0.3428 \\
  & Balanced Mapping & 0.35 &  1.0637 & \textbf{32.936} & 29.472 & 0.6981 & 0.3394 \\
  & Balanced Mapping & 0.40 &  1.0600 & 32.906 & 29.519 & 0.7002 & 0.3368 \\
  & Balanced Mapping & 0.45 &  1.0591 & 32.901 & 29.556 & 0.7024 & 0.3342 \\
  & Balanced Mapping & 0.50 &  1.0467 & 32.871 & 29.586 & \underline{0.7032} & \underline{0.3329} \\
  & Balanced Mapping & 0.55 &  1.0542 & 32.894 & 29.608 & \textbf{0.7034} & \textbf{0.3328} \\
  & Balanced Mapping & 0.60 &  1.0474 & 32.889 & 29.622 & 0.7031 & 0.3342 \\
  & Balanced Mapping & 0.65 &  1.0386 & \underline{32.920} & 29.634 & \textbf{0.7034} & 0.3345 \\
  & Balanced Mapping & 0.70 &  1.0341 & 32.902 & 29.643 & \textbf{0.7034} & 0.3350 \\
  & Balanced Mapping & 0.75 &  1.0311 & 32.862 & 29.648 & 0.7029 & 0.3360 \\
  & Balanced Mapping & 0.80 &  1.0242 & 32.862 & 29.652 & 0.7028 & 0.3363 \\
  & Balanced Mapping & 0.85 &  1.0271 & 32.809 & 29.653 & 0.7026 & 0.3371 \\
  & Balanced Mapping & 0.90 &  1.0310 & 32.762 & \underline{29.656} & 0.7022 & 0.3377 \\
  & Balanced Mapping & 0.95 &  1.0274 & 32.832 & \textbf{29.657} & 0.7020 & 0.3383 \\
  & Balanced Mapping & 1.00 &  1.0280 & 32.800 & 29.654 & 0.7015 & 0.3392 \\
  & Balanced Mapping & 1.30 &  1.0205 & 32.806 & 29.638 & 0.7001 & 0.3420 \\
  & Balanced Mapping & 1.50 &  1.0211 & 32.733 & 29.629 & 0.6994 & 0.3437 \\
  & Balanced Mapping & 1.80 &  1.0133 & 32.745 & 29.619 & 0.6982 & 0.3462 \\
  & Balanced Mapping & 2.00 &  1.0162 & 32.756 & 29.615 & 0.6977 & 0.3471 \\
  & Balanced Mapping & 2.50 &  1.0148 & 32.752 & 29.610 & 0.6972 & 0.3482 \\
  & Balanced Mapping & 5.00 &  1.0182 & 32.770 & 29.606 & 0.6968 & 0.3489 \\
  & Balanced Mapping & 10.00 &  1.0161 & 32.752 & 29.607 & 0.6968 & 0.3490 \\
  & Balanced Mapping & 20.00 &  \textcolor{blue!70}{1.0129} & 32.760 & 29.606 & 0.6967 & 0.3491 \\
  & Balanced Mapping & 50.00 &  \textcolor{blue!70}{1.0129} & 32.760 & 29.606 & 0.6967 & 0.3491 \\

\midrule

\multirow{30}{*}{$\textbf{ResilPhase} $ $(\mathcal{N}=4,O=1)$} & Chebyshev Mapping & -  &  1.0595 & 32.914 & \textcolor{blue!70}{29.448} & \textcolor{blue!70}{0.7004} & \textcolor{blue!70}{0.3281} \\

\cline{2-8}
  
  & Balanced Mapping & 0.05 &  \underline{1.0653} & \textbf{32.939} & 29.467 & 0.7010 & 0.3275 \\
  & Balanced Mapping & 0.10 &  1.0614 & \underline{32.931} & 29.514 & 0.7032 & 0.3242 \\
  & Balanced Mapping & 0.15 &  \textbf{1.0680} & 32.876 & 29.590 & 0.7069 & 0.3192 \\
  & Balanced Mapping & 0.20 &  1.0545 & 32.769 & 29.673 & 0.7110 & 0.3129 \\
  & Balanced Mapping & 0.25 &  1.0602 & \textcolor{blue!70}{32.746} & 29.744 & 0.7142 & 0.3075 \\
  & Balanced Mapping & 0.30 &  1.0613 & 32.788 & 29.807 & 0.7163 & 0.3051 \\
  & Balanced Mapping & 0.35 &  1.0565 & 32.770 & 29.865 & 0.7179 & 0.3022 \\
  & Balanced Mapping & 0.40 &  1.0501 & 32.801 & 29.911 & 0.7189 & 0.3006 \\
  & Balanced Mapping & 0.45 &  1.0546 & 32.826 & 29.945 & 0.7197 & 0.2999 \\
  & Balanced Mapping & 0.50 &  1.0536 & 32.824 & 29.975 & 0.7206 & 0.2995 \\
  & Balanced Mapping & 0.55 &  1.0621 & 32.842 & 30.002 & \textbf{0.7216} & \textbf{0.2989} \\
  & Balanced Mapping & 0.60 &  1.0581 & 32.874 & 30.020 & 0.7212 & \underline{0.2992} \\
  & Balanced Mapping & 0.65 &  1.0505 & 32.846 & 30.031 & \underline{0.7215} & 0.2996 \\
  & Balanced Mapping & 0.70 &  1.0462 & 32.884 & \textbf{30.033} & 0.7212 & 0.3002 \\
  & Balanced Mapping & 0.75 &  1.0450 & 32.865 & \underline{30.032} & 0.7211 & 0.3009 \\
  & Balanced Mapping & 0.80 &  1.0463 & 32.866 & 30.029 & 0.7206 & 0.3017 \\
  & Balanced Mapping & 0.85 &  1.0448 & 32.909 & 30.024 & 0.7199 & 0.3032 \\
  & Balanced Mapping & 0.90 &  1.0445 & 32.858 & 30.019 & 0.7195 & 0.3038 \\
  & Balanced Mapping & 0.95 &  1.0476 & 32.899 & 30.012 & 0.7189 & 0.3047 \\
  & Balanced Mapping & 1.00 &  1.0457 & 32.857 & 30.006 & 0.7185 & 0.3057 \\
  & Balanced Mapping & 1.30 &  1.0264 & 32.876 & 29.972 & 0.7152 & 0.3112 \\
  & Balanced Mapping & 1.50 &  1.0279 & 32.883 & 29.954 & 0.7141 & 0.3133 \\
  & Balanced Mapping & 1.80 &  1.0323 & 32.858 & 29.934 & 0.7128 & 0.3153 \\
  & Balanced Mapping & 2.00 &  1.0314 & 32.857 & 29.929 & 0.7127 & 0.3159 \\
  & Balanced Mapping & 2.50 &  1.0286 & 32.881 & 29.917 & 0.7119 & 0.3174 \\
  & Balanced Mapping & 5.00 &  1.0243 & 32.834 & 29.913 & 0.7115 & 0.3176 \\
  & Balanced Mapping & 10.00 &  \textcolor{blue!70}{1.0212} & 32.854 & 29.914 & 0.7117 & 0.3176 \\
  & Balanced Mapping & 20.00 &  1.0223 & 32.893 & 29.912 & 0.7116 & 0.3178 \\
  & Balanced Mapping & 50.00 &  1.0223 & 32.893 & 29.912 & 0.7116 & 0.3178 \\

    \bottomrule
  \end{tabular}}
  
  \label{table:FLUX-Metrics-alpha-2}
\vspace{-4mm}
\end{table*}
\clearpage

\begin{table*}[htbp]
\centering
\caption{\textbf{Ablation study of Phase Mapping and $\alpha$ scaling on HunyuanVideo. We compare Chebyshev Mapping with Balanced Mapping, specifically analyzing the influence of the parameter $\alpha$ on the final video generation results.
}}
\vspace{-3mm}
\setlength\tabcolsep{7.0pt} 
%\belowrulesep=0pt
%\aboverulesep=0pt
  \small
  \resizebox{0.85\textwidth}{!}{
  \begin{tabular}{c | c | c | c | c |c }
    \toprule
    \multirow{2}{*}{\bf Configuration} & Phase & \multirow{2}{*}{\bf $\alpha$}  &  \multirow{2}{*}{\bf PSNR$\uparrow$} & \multirow{2}{*}{\bf SSIM$\uparrow$} & \multirow{2}{*}{\bf LPIPS$\downarrow$}\\
    % \cline{2-3}
    & Mapping & & & \\
    \midrule

\multirow{30}{*}{$\textbf{ResilPhase} $  $(\mathcal{N}=6,O=1)$ } & Chebyshev Mapping & -  &  \textcolor{blue!70}{17.400} & \textcolor{blue!70}{0.6381} & \textcolor{blue!70}{0.3719} \\

\cline{2-6}
  
  & Balanced Mapping & 0.05 & 17.483 & 0.6407 & 0.3685 \\
  & Balanced Mapping & 0.10 & 17.644 & 0.6460 & 0.3643 \\
  & Balanced Mapping & 0.15 & 17.877 & 0.6516 & 0.3578 \\
  & Balanced Mapping & 0.20 & 18.132 & 0.6581 & 0.3501 \\
  & Balanced Mapping & 0.25 & 18.352 & 0.6629 & 0.3424 \\
  & Balanced Mapping & 0.30 & 18.518 & 0.6660 & 0.3387 \\
  & Balanced Mapping & 0.35 & 18.640 & 0.6678 & 0.3380 \\
  & Balanced Mapping & 0.40 & 18.732 & 0.6693 & 0.3365 \\
  & Balanced Mapping & 0.45 & 18.821 & 0.6700 & \textbf{0.3338} \\
  & Balanced Mapping & 0.50 & 18.878 & \textbf{0.6709} & 0.3342 \\
  & Balanced Mapping & 0.55 & 18.920 & \textbf{0.6709} & \underline{0.3341} \\
  & Balanced Mapping & 0.60 & 18.937 & \underline{0.6706} & 0.3350 \\
  & Balanced Mapping & 0.65 & 18.943 & 0.6694 & 0.3357 \\
  & Balanced Mapping & 0.70 & 18.972 & 0.6690 & 0.3364 \\
  & Balanced Mapping & 0.75 & 18.986 & 0.6685 & 0.3368 \\
  & Balanced Mapping & 0.80 & \textbf{18.996} & 0.6681 & 0.3373 \\
  & Balanced Mapping & 0.85 & 18.990 & 0.6669 & 0.3392 \\
  & Balanced Mapping & 0.90 & \underline{18.992} & 0.6668 & 0.3405 \\
  & Balanced Mapping & 0.95 & 18.989 & 0.6661 & 0.3414 \\
  & Balanced Mapping & 1.00 & 18.985 & 0.6655 & 0.3418 \\
  & Balanced Mapping & 1.30 & 18.972 & 0.6638 & 0.3453 \\
  & Balanced Mapping & 1.50 & 18.966 & 0.6625 & 0.3476 \\
  & Balanced Mapping & 1.80 & 18.965 & 0.6619 & 0.3482 \\
  & Balanced Mapping & 2.00 & 18.955 & 0.6612 & 0.3494 \\
  & Balanced Mapping & 2.50 & 18.952 & 0.6608 & 0.3499 \\
  & Balanced Mapping & 5.00 & 18.946 & 0.6605 & 0.3506 \\
  & Balanced Mapping & 10.00 & 18.947 & 0.6605 & 0.3512 \\
  & Balanced Mapping & 20.00 & 18.945 & 0.6604 & 0.3504 \\
  & Balanced Mapping & 50.00 & 18.945 & 0.6604 & 0.3504 \\

\midrule

\multirow{30}{*}{$\textbf{ResilPhase} $  $(\mathcal{N}=5,O=1)$ } & Chebyshev Mapping & -  &  \textcolor{blue!70}{18.091} & \textcolor{blue!70}{0.6669} & \textcolor{blue!70}{0.3336} \\

\cline{2-6}
  
  & Balanced Mapping & 0.05 & 18.169 & 0.6689 & 0.3307 \\
  & Balanced Mapping & 0.10 & 18.308 & 0.6719 & 0.3289 \\
  & Balanced Mapping & 0.15 & 18.519 & 0.6778 & 0.3234 \\
  & Balanced Mapping & 0.20 & 18.711 & 0.6825 & 0.3173 \\
  & Balanced Mapping & 0.25 & 18.936 & 0.6883 & 0.3103 \\
  & Balanced Mapping & 0.30 & 19.084 & 0.6917 & 0.3069 \\
  & Balanced Mapping & 0.35 & 19.192 & 0.6938 & 0.3039 \\
  & Balanced Mapping & 0.40 & 19.299 & 0.6957 & 0.3019 \\
  & Balanced Mapping & 0.45 & 19.379 & 0.6976 & 0.2986 \\
  & Balanced Mapping & 0.50 & 19.427 & 0.6977 & 0.2977 \\
  & Balanced Mapping & 0.55 & \textbf{19.572} & \underline{0.6992} & \underline{0.2954} \\
  & Balanced Mapping & 0.60 & 19.498 & 0.6991 & 0.2956 \\
  & Balanced Mapping & 0.65 & 19.523 & \textbf{0.6993} & \textbf{0.2952} \\
  & Balanced Mapping & 0.70 & 19.542 & 0.6992 & 0.2958 \\
  & Balanced Mapping & 0.75 & 19.549 & 0.6986 & 0.2963 \\
  & Balanced Mapping & 0.80 & \underline{19.566} & 0.6987 & 0.2964 \\
  & Balanced Mapping & 0.85 & 19.563 & 0.6981 & 0.2968 \\
  & Balanced Mapping & 0.90 & 19.563 & 0.6975 & 0.2974 \\
  & Balanced Mapping & 0.95 & 19.563 & 0.6970 & 0.2978 \\
  & Balanced Mapping & 1.00 & 19.481 & 0.6965 & 0.3008 \\
  & Balanced Mapping & 1.30 & 19.542 & 0.6940 & 0.3039 \\
  & Balanced Mapping & 1.50 & 19.534 & 0.6929 & 0.3058 \\
  & Balanced Mapping & 1.80 & 19.512 & 0.6916 & 0.3072 \\
  & Balanced Mapping & 2.00 & 19.508 & 0.6913 & 0.3077 \\
  & Balanced Mapping & 2.50 & 19.507 & 0.6908 & 0.3095 \\
  & Balanced Mapping & 5.00 & 19.500 & 0.6900 & 0.3097 \\
  & Balanced Mapping & 10.00 & 19.492 & 0.6898 & 0.3098 \\
  & Balanced Mapping & 20.00 & 19.499 & 0.6902 & 0.3099 \\
  & Balanced Mapping & 50.00 & 19.499 & 0.6902 & 0.3099 \\

    \bottomrule
  \end{tabular}
}  
  \label{table:hyvideo-alpha}
\vspace{-4mm}
\end{table*}
\clearpage

\begin{table*}[htbp]
\centering
\caption{\textbf{Ablation study of Phase Mapping and $\alpha$ scaling on DiT-XL/2. We compare Chebyshev Mapping with Balanced Mapping, specifically analyzing the influence of the parameter $\alpha$ on the final class-conditional image generation results.
}}
\vspace{-3mm}
\setlength\tabcolsep{7.0pt} 
%\belowrulesep=0pt
%\aboverulesep=0pt
  \small
  \resizebox{0.9\textwidth}{!}{
  \begin{tabular}{c | c | c | c | c |c }
    \toprule
    \multirow{2}{*}{\bf Configuration} & Phase & \multirow{2}{*}{\bf $\alpha$}  &  \multirow{2}{*}{\bf FID$\downarrow$} & \multirow{2}{*}{\bf sFID$\downarrow$} & \multirow{2}{*}{\bf IS$\uparrow$}\\
    % \cline{2-3}
    & Mapping & & & \\
    \midrule

\multirow{26}{*}{$\textbf{ResilPhase} $  $(\mathcal{N}=5,O=3)$ } & Chebyshev Mapping & -  &  \textbf{2.832} & {5.026} & \textbf{219.19} \\

\cline{2-6}
  
  & Balanced Mapping & 0.05 & 2.919 & 4.878 & 217.12 \\
  & Balanced Mapping & 0.10 & 2.912 & 4.862 & 217.38 \\
  & Balanced Mapping & 0.15 & 2.903 & 4.832 & 217.67 \\
  & Balanced Mapping & 0.20 & 2.897 & 4.793 & 218.16 \\
  & Balanced Mapping & 0.25 & \underline{2.890} & 4.742 & 218.43 \\
  & Balanced Mapping & 0.30 & 2.892 & 4.691 & 218.87 \\
  & Balanced Mapping & 0.35 & 2.911 & 4.647 & 219.07 \\
  & Balanced Mapping & 0.40 & 2.936 & 4.607 & 217.90 \\
  & Balanced Mapping & 0.45 & 2.972 & 4.573 & 218.84 \\
  & Balanced Mapping & 0.50 & 3.009 & 4.547 & 218.40 \\
  & Balanced Mapping & 0.55 & 3.066 & \textbf{4.504} & \underline{219.10} \\
  & Balanced Mapping & 0.60 & 3.133 & 4.510 & 216.99 \\
  & Balanced Mapping & 0.65 & 3.209 & 4.524 & 216.35 \\
  & Balanced Mapping & 0.70 & 3.285 & \underline{4.505} & 215.59 \\
  & Balanced Mapping & 0.75 & 3.372 & 4.509 & 214.37 \\
  & Balanced Mapping & 0.80 & 3.473 & 4.520 & 213.20 \\
  & Balanced Mapping & 0.85 & 3.583 & 4.540 & 212.40 \\
  & Balanced Mapping & 0.90 & 3.694 & 4.566 & 211.46 \\
  & Balanced Mapping & 0.95 & 3.815 & 4.599 & 210.40 \\
  & Balanced Mapping & 1.00 & 3.935 & 4.635 & 209.18 \\
  & Balanced Mapping & 1.50 & 5.170 & 5.103 & 197.65 \\
  & Balanced Mapping & 3.00 & 6.972 & 5.941 & 183.87 \\
  & Balanced Mapping & 5.00 & 7.252 & 6.081 & 182.08 \\
  & Balanced Mapping & 10.00 & 7.271 & 6.091 & 181.97 \\
  & Balanced Mapping & 20.00 & 7.272 & 6.091 & 181.98 \\
  & Balanced Mapping & 50.00 & \textcolor{blue!70}{55.009} & \textcolor{blue!70}{33.620} & \textcolor{blue!70}{59.760} \\

\midrule

\multirow{26}{*}{$\textbf{ResilPhase} $  $(\mathcal{N}=3,O=4)$ } & Chebyshev Mapping & -  &  \textbf{2.347} & {4.672} & \textbf{233.57} \\

\cline{2-6}
  
  & Balanced Mapping & 0.05 & 2.516 & 4.564 & 228.30 \\
  & Balanced Mapping & 0.10 & 2.512 & 4.555 & 228.44 \\
  & Balanced Mapping & 0.15 & 2.505 & 4.540 & 228.78 \\
  & Balanced Mapping & 0.20 & 2.497 & 4.523 & 229.26 \\
  & Balanced Mapping & 0.25 & 2.492 & 4.504 & 229.65 \\
  & Balanced Mapping & 0.30 & 2.486 & 4.488 & 230.13 \\
  & Balanced Mapping & 0.35 & \underline{2.481} & 4.473 & 230.51 \\
  & Balanced Mapping & 0.40 & 2.482 & 4.460 & 230.86 \\
  & Balanced Mapping & 0.45 & 2.486 & 4.450 & 231.05 \\
  & Balanced Mapping & 0.50 & 2.494 & 4.443 & 231.01 \\
  & Balanced Mapping & 0.55 & 2.504 & 4.438 & \underline{231.06} \\
  & Balanced Mapping & 0.60 & 2.516 & 4.433 & 230.94 \\
  & Balanced Mapping & 0.65 & 2.534 & 4.429 & 230.85 \\
  & Balanced Mapping & 0.70 & 2.554 & 4.424 & 230.69 \\
  & Balanced Mapping & 0.75 & 2.576 & 4.417 & 230.52 \\
  & Balanced Mapping & 0.80 & 2.600 & 4.415 & 230.27 \\
  & Balanced Mapping & 0.85 & 2.625 & 4.412 & 229.76 \\
  & Balanced Mapping & 0.90 & 2.652 & \underline{4.408} & 229.25 \\
  & Balanced Mapping & 0.95 & 2.680 & \textbf{4.407} & 229.02 \\
  & Balanced Mapping & 1.00 & 2.713 & \textbf{4.407} & 228.58 \\
  & Balanced Mapping & 1.50 & 3.053 & 4.453 & 222.97 \\
  & Balanced Mapping & 3.00 & 3.679 & 4.636 & 215.44 \\
  & Balanced Mapping & 5.00 & 3.826 & 4.700 & 213.71 \\
  & Balanced Mapping & 10.00 & 3.849 & 4.710 & 213.45 \\
  & Balanced Mapping & 20.00 & 3.849 & 4.711 & 213.41 \\
  & Balanced Mapping & 50.00 & \textcolor{blue!70}{13.860} & \textcolor{blue!70}{9.417} & \textcolor{blue!70}{147.43} \\

    \bottomrule
  \end{tabular}
}  
  \label{table:DiT-alpha1}
\vspace{-4mm}
\end{table*}
\clearpage

\begin{table*}[htbp]
\centering
\caption{\textbf{Ablation study of Phase Mapping and $\alpha$ scaling on DiT-XL/2. We compare Chebyshev Mapping with Balanced Mapping, specifically analyzing the influence of the parameter $\alpha$ on the final class-conditional image generation results.
}}
\vspace{-3mm}
\setlength\tabcolsep{7.0pt} 
%\belowrulesep=0pt
%\aboverulesep=0pt
  \small
  \resizebox{0.9\textwidth}{!}{
  \begin{tabular}{c | c | c | c | c |c }
    \toprule
    \multirow{2}{*}{\bf Configuration} & Phase & \multirow{2}{*}{\bf $\alpha$}  &  \multirow{2}{*}{\bf FID$\downarrow$} & \multirow{2}{*}{\bf sFID$\downarrow$} & \multirow{2}{*}{\bf IS$\uparrow$}\\
    % \cline{2-3}
    & Mapping & & & \\
    \midrule

\multirow{26}{*}{$\textbf{ResilPhase} $  $(\mathcal{N}=2,O=3)$ } & Chebyshev Mapping & -  &  \textbf{2.306} & {4.427} & \textbf{236.75} \\

\cline{2-6}
  
  & Balanced Mapping & 0.05 & \underline{2.383} & 4.383 & 235.47 \\
  & Balanced Mapping & 0.10 & 2.384 & 4.382 & 235.45 \\
  & Balanced Mapping & 0.15 & 2.385 & 4.382 & \underline{235.49} \\
  & Balanced Mapping & 0.20 & 2.388 & 4.380 & 235.42 \\
  & Balanced Mapping & 0.25 & 2.392 & 4.378 & 235.34 \\
  & Balanced Mapping & 0.30 & 2.397 & 4.376 & 235.32 \\
  & Balanced Mapping & 0.35 & 2.401 & 4.375 & 235.24 \\
  & Balanced Mapping & 0.40 & 2.406 & 4.374 & 235.18 \\
  & Balanced Mapping & 0.45 & 2.413 & 4.372 & 235.07 \\
  & Balanced Mapping & 0.50 & 2.422 & 4.371 & 234.77 \\
  & Balanced Mapping & 0.55 & 2.431 & 4.369 & 234.67 \\
  & Balanced Mapping & 0.60 & 2.441 & \textbf{4.368} & 234.50 \\
  & Balanced Mapping & 0.65 & 2.451 & \textbf{4.368} & 234.17 \\
  & Balanced Mapping & 0.70 & 2.463 & \underline{4.369} & 233.87 \\
  & Balanced Mapping & 0.75 & 2.476 & 4.370 & 233.71 \\
  & Balanced Mapping & 0.80 & 2.490 & 4.370 & 233.40 \\
  & Balanced Mapping & 0.85 & 2.504 & 4.370 & 233.09 \\
  & Balanced Mapping & 0.90 & 2.520 & 4.370 & 232.73 \\
  & Balanced Mapping & 0.95 & 2.533 & 4.370 & 232.61 \\
  & Balanced Mapping & 1.00 & 2.549 & 4.370 & 232.44 \\
  & Balanced Mapping & 1.50 & 2.685 & 4.385 & 230.17 \\
  & Balanced Mapping & 3.00 & 2.872 & 4.421 & 227.12 \\
  & Balanced Mapping & 5.00 & 2.896 & 4.429 & 226.76 \\
  & Balanced Mapping & 10.00 & 2.899 & 4.429 & 226.72 \\
  & Balanced Mapping & 20.00 & 2.899 & 4.430 & 226.75 \\
  & Balanced Mapping & 50.00 & \textcolor{blue!70}{4.999} & \textcolor{blue!70}{5.241} & \textcolor{blue!70}{201.75} \\

    \bottomrule
  \end{tabular}
}  
  \label{table:DiT-alpha2}
\vspace{-4mm}
\end{table*}
\clearpage

\section{Theoretical Error Comparison: ResilPhase vs. Derivative-Based Baselines}

In this section, we provide a comprehensive theoretical comparison between our proposed derivative-free ResilPhase framework and existing derivative-based forecasting methods, specifically TaylorSeer (Taylor expansion) and HiCache (Scaled-Hermite expansion).

\subsection{Error Bounds of Taylor-Style and Hermite-Style Forecasting}

The prediction errors of derivative-based forecasting methods generally arise 
from finite difference approximations and the inherent truncation of the polynomial expansion. The error growth in both paradigms is fundamentally dominated by the prediction step size, denoted as $k$.

To maintain consistent mathematical notation, we explicitly define the continuous prediction step size $k$ as the relative temporal distance between the target extrapolation step $t_{\text{target}}$ and the most recent fully computed anchor step $t_m$, i.e., $k = t_{\text{target}} - t_m$. Under a standard uniform discrete sampling schedule, assuming adjacent timesteps have a normalized unit spacing ($\Delta t = 1$), applying an acceleration ratio of $N$ implies the model skips $N - 1$ intermediate steps. Consequently, the maximum continuous prediction distance required during the extrapolation phase is exactly $|k|_{\max} = N - 1$.

\textbf{Error Bound of TaylorSeer.} Based on Taylor's Theorem, for a function 
approximated by an $m$-th order polynomial, the truncation error is bounded by the $(m+1)$-th derivative. Additionally, TaylorSeer approximates these derivatives using finite differences, introducing an approximation error. Combining these two terms, the total error bound $E_m(k)$ grows with the power of the step size $k$:

\begin{equation}
E_m(k) \leq \frac{M_{m+1}}{(m + 1)!}|k|^{m+1} + \sum_{i=1}^{m} \frac{C_i}{i! \cdot |N|^{i-1}}|k|^i.
\end{equation}

where $M_{m+1}$ represents the maximum magnitude of the high-order derivative, and $C_i$ relates to finite difference coefficients. Substituting the aforementioned maximum prediction distance $|k|_{\max} = N - 1$ into this error formula, it becomes evident that the upper bound is fundamentally governed by the explosive term $(N - 1)^{m+1}$.

\textbf{Error Bound of HiCache (Scaled-Hermite).} HiCache replaces the Taylor basis with scaled Hermite polynomials to suppress the error bound using a scaling factor $\sigma \in (0,1)$. According to its theoretical formulation, the total error $E_{\text{total}}^{\text{Hermite}}$ is decomposed into truncation, approximation, and numerical errors:
\begin{equation}
E_{\text{total}}^{\text{Hermite}} \approx E_{\text{truncation}} + E_{\text{approximation}} + E_{\text{numerical}}.
\end{equation}
Crucially, both the truncation and approximation errors are heavily dependent on the unbounded prediction step size $k$. The truncation error bound is given by:
\begin{equation}
E_{\text{truncation}}^{\text{Hermite}} \leq C_2 \frac{(\sigma \sqrt{2|k|})^{m+1}}{\sqrt{2\pi}(m+1)!} \exp\left(\frac{(\sigma k)^2}{2}\right),
\end{equation}
And the finite-difference approximation error is bounded by $E_{\text{approximation}}^{\text{Hermite}} \leq C_4 \frac{k}{\sqrt{m}}$. Both error terms will grow uncontrollably as the maximum linear step size $|k|_{\max} = N-1$ increases at high acceleration ratios.

\subsection{Error Bound of ResilPhase and Comparative Analysis}

Unlike TaylorSeer and HiCache, which rely on noisy finite-difference derivatives, ResilPhase adopts a derivative-free Barycentric Lagrange extrapolation framework operating within a bounded phase space $s \in [-1, 1]$.

\textbf{Error Bound of ResilPhase.} By formulating the prediction target as the ODE-aligned Global Drift and projecting discrete timesteps into the continuous phase space via Phase Mapping, the extrapolation error polynomial for ResilPhase is strictly defined by the classical Lagrange error bound:
\begin{equation}
E_m^{\text{ResilPhase}}(s) \leq \frac{M_{m+1}^s}{(m+1)!}\prod_{j=0}^{m}|s - s_j|,
\end{equation}
where $s_j$ are the mapped historical nodes, and $M_{m+1}^s$ is the maximum $(m+1)$-th derivative of the macro-trajectory with respect to the phase variable $s$. 

\textbf{Comparative Analysis on Task Complexity.} The critical advantage of ResilPhase lies in its ability to explicitly control the node term $\prod_{j=0}^{m}|s - s_j|$ through different Phase Mapping strategies, adapting to varying task complexities:

\begin{itemize}
    \item \textbf{For Class-Conditional Tasks (Simpler Dynamics):} Class-conditional generation (e.g., ImageNet on DiT-XL/2) exhibits relatively stable and predictable ODE trajectories. In such low-variance environments, ResilPhase employs Chebyshev Mapping. By algebraically distributing the nodes $\{s_j\}$ according to Chebyshev roots, classical approximation theory mathematically guarantees that the maximum value of the node term is minimized over the interval $[-1, 1]$:
    \begin{equation}
    \max_s \prod_{j=0}^{m}|s - s_j| \le \frac{1}{2^m}.
    \end{equation}
    Consequently, the maximum extrapolation error of ResilPhase with Chebyshev Mapping is strictly bounded by:
    \begin{equation}
    E_{\text{max}}^{\text{ResilPhase}} \propto \frac{M_{m+1}^s}{2^m (m+1)!}.
    \end{equation}
    Compared to the $O((N-1)^{m+1})$ growth in TaylorSeer and the explosive $O(\exp((\sigma k)^2/2))$ term in HiCache, ResilPhase with Chebyshev Mapping offers a significantly lower, mathematically tight theoretical error bound, making it optimal for class-conditional tasks.
    
    \item \textbf{For Text-to-Image/Video Tasks (Complex Dynamics):} Conversely, prompt-driven generation tasks (e.g., FLUX.1 and HunyuanVideo) involve highly complex, high-variance cross-attention interactions that create localized non-linearities in the continuous trajectory. In these scenarios, the rigid mathematical distribution of Chebyshev nodes struggles to capture sudden dynamic shifts. Therefore, ResilPhase utilizes the data-driven Balanced Mapping. Rather than minimizing a theoretical uniform bound, Balanced Mapping dynamically reallocates mapped nodes based on the real-time temporal distribution (mean and absolute deviation) of recent historical steps. This adaptive reallocation confines numerical outliers and focuses the extrapolation polynomial on the most critical, high-variance phases of the text-conditioned ODE trajectory.
\end{itemize}

In summary, derivative-based methods like TaylorSeer and HiCache suffer from unbounded linear step sizes ($|k|$) and noisy finite-difference approximations, inevitably leading to error explosion at high speedups. ResilPhase fundamentally avoids these pitfalls by combining a derivative-free macro-trajectory objective with bounded Phase Mappings. By matching Chebyshev Mapping with simpler class-conditional tasks for strict mathematical minimization, and Balanced Mapping with complex prompt-driven tasks for dynamic node reallocation, ResilPhase consistently ensures a lower and more stable error bound across diverse generative tasks.

\section{Generalizability Analysis of Phase Mapping on Derivative-Based Accelerators}

In the previous section, we established that existing derivative-based methods (such as TaylorSeer and HiCache) suffer from error explosion due to their reliance on an unbounded linear prediction step size ($|k|_{\max} = N-1$). In this section, we provide a theoretical proof and empirical validation to demonstrate that our proposed Phase Mapping mechanism can be applied as a ``plug-and-play'' regularizer to universally suppress these extrapolation error bounds.

\subsection{Universal Error Reduction with Phase Mapping}

When Phase Mapping (e.g., Chebyshev or Balanced Mapping) is integrated into TaylorSeer or HiCache, the prediction is no longer performed in the unbounded linear time domain. Instead, it relies on the mapped phase coordinates $s$. The derivative estimation and the effective prediction step size now strictly depend on the distance between the mapped nodes:
\begin{equation}
\Delta s = |s(t_m) - s(t_m - N + 1)|.
\end{equation}

\textbf{Proof of Error Reduction:} Both Chebyshev Mapping and Balanced Mapping non-linearly transform the discrete timesteps into a strictly bounded interval, typically within $[-1, 1]$. Consequently, the mapped distance $\Delta s$ between any two nodes is mathematically constrained by the finite length of this interval. In contrast, without phase mapping, the linear distance grows linearly with the acceleration ratio ($N - 1$). Because the mapped phase domain is strictly bounded (typically within $[-1, 1]$, making the absolute maximum distance $\leq 2$), the mapped step size $\Delta s$ is fundamentally compressed. For practical acceleration ratios (e.g., $N \geq 4$), this mathematically guarantees:
\begin{equation}
\Delta s \leq 2 < N - 1.
\end{equation}
Even for smaller $N$, the non-linear compression ensures $\Delta s$ remains 
significantly bounded compared to unbounded linear time.

By substituting the unbounded linear step size $|k|$ with the constrained phase distance $\Delta s$, the error bounds of both derivative-based frameworks are strictly reduced:

\begin{itemize}
    \item \textbf{For TaylorSeer:} The substitution directly compresses the dominant polynomial term $|k|^{m+1}$ into $|\Delta s|^{m+1}$. Because the truncation error bound is a monotonically increasing function of the step size, the significant reduction in distance ($\Delta s \ll N - 1$) guarantees a strictly lower overall error upper bound.
    
    \item \textbf{For HiCache:} Applying Phase Mapping yields an even more profound stabilization. By replacing the linear step size with the mapped distance $\Delta s$, the explosive exponential term in its truncation error bound is heavily suppressed:
    \begin{equation}
    E_{\text{truncation}}^{\text{mapped}} \propto \exp\left(\frac{(\sigma \Delta s)^2}{2}\right) \ll \exp\left(\frac{(\sigma (N-1))^2}{2}\right).
    \end{equation}
    Simultaneously, its finite-difference approximation error is reduced from $O(N-1)$ to $O(\Delta s)$. 
\end{itemize}

\textbf{Theoretical Remark on Non-Uniformity and Derivative Magnitude:} It is worth noting that mapping discrete uniform timesteps $t$ into the non-linear phase space $s$ inherently results in non-uniformly spaced historical nodes $s_j$. While traditional finite difference formulations strictly assume uniform grids, applying divided differences allows derivative-based solvers to operate on this non-uniform $s$-domain. Furthermore, according to the chain rule, this non-linear temporal compression mapping could theoretically increase the magnitude of the derivative terms in the mapped $s$-domain (since $dt/ds > 1$). However, mathematically, the exponential reduction achieved by the strictly bounded power term $(\Delta s)^{m+1}$ (or the $\exp\left(\frac{(\sigma \Delta s)^2}{2}\right)$ term) vastly dominates any linear or polynomial growth in the derivative coefficients. This fundamentally guarantees that the overall theoretical error bound is massively suppressed, as unequivocally confirmed by our empirical results.

Therefore, regardless of whether the underlying basis functions are Taylor 
series or Hermite polynomials, Phase Mapping acts as a universal, mathemat
ically rigorous regularizer that strictly bounds the extrapolation domain and 
minimizes the maximum prediction error.

\subsection{Experimental Validation on Derivative-Based Frameworks}

To empirically validate the theoretical error reductions derived above, we conduct an ablation study applying our Phase Mapping mechanisms as a ``plug-and-play'' enhancement to the TaylorSeer and HiCache frameworks. We evaluate their performance equipped with either Chebyshev Mapping or Balanced Mapping across three generative tasks: Text-to-Image, Text-to-Video, and Class-Conditional Image Generation. 

The results across Tables~\ref{table:taylor-FLUX-ablation}--\ref{table:taylor-dit-ablation} demonstrate that Phase Mapping consistently improves TaylorSeer's performance across all tasks and acceleration ratios. More importantly, Tables~\ref{table:hicache-FLUX-ablation} validates the profound impact of Phase Mapping on the recently proposed HiCache framework. According to HiCache's original theory, its scaled Hermite polynomials are supposed to sufficiently suppress prediction errors. However, our empirical results reveal that applying our Balanced Mapping to HiCache yields massive improvements on FLUX.1-dev. For instance, at the highly aggressive $\mathcal{N}=11$ acceleration tier, plugging Balanced Mapping into HiCache boosts the ImageReward from 0.8040 to an impressive 0.9024, while simultaneously raising the SSIM and lowering the LPIPS.

Consistent with our previous findings across the main paper, Balanced Mapping consistently works best for complex, non-linear text-to-image and text-to-video tasks, while Chebyshev Mapping remains superior for class-conditional generation. This comprehensive evaluation rigorously confirms that our Phase Mapping mechanism is a highly robust, transferable, and essential enhancement for any acceleration method relying on polynomial prediction, regardless of the underlying mathematical basis.

\begin{table*}[htbp]
\centering
\caption{\textbf{ Ablation study of Phase Mapping components on TaylorSeer for FLUX.1-dev.}}
\label{table:taylor-FLUX-ablation}
\vspace{-3mm}
\setlength\tabcolsep{7.0pt}
\small
\resizebox{\textwidth}{!}{
\begin{tabular}{ c | c  c|c|c|c|c|c}
\toprule
\multirow{2}{*}{\bf Configuration}  &  \multicolumn{2}{c|}{\bf Phase Mapping} &   {\bf ImageReward $\uparrow$} & \bf CLIP$\uparrow$ & \multirow{2}{*}{\bf PSNR$\uparrow$} & \multirow{2}{*}{\bf SSIM$\uparrow$} & \multirow{2}{*}{\bf LPIPS$\downarrow$}\\
\cline{2-3}

 &  {\bf Chebyshev} & {\bf Balance} & \bf DrawBench & \bf Score & & & \\
\midrule

\multirow{3}{*}{$\textbf{TaylorSeer} $ $(\mathcal{N}=11,O=2)$} & & & \textcolor{blue!70}{0.6241} & \underline{31.895} & \textcolor{blue!70}{27.940} & \textcolor{blue!70}{0.3014} & \textcolor{blue!70}{0.8012}\\
& \ding{52} & & \underline{0.6556} & \textcolor{blue!70}{30.843} & \underline{28.160} & \underline{0.4563} & \underline{0.6733}\\
& & \ding{52} & \textbf{0.9169} & \textbf{32.211} & \textbf{28.520} & \textbf{0.5982} & \textbf{0.4905} \\ 

\midrule

\multirow{3}{*}{$\textbf{TaylorSeer} $ $(\mathcal{N}=7,O=2)$} & & & \textcolor{blue!70}{0.9406} & \textbf{32.657} & \textcolor{blue!70}{28.049} & \textcolor{blue!70}{0.3931} & \textcolor{blue!70}{0.7119}\\
& \ding{52} & & \underline{0.9869} & \underline{32.608} & \underline{28.502} & \underline{0.5849} & \underline{0.5029}\\
& & \ding{52} & \textbf{1.0186} & \textcolor{blue!70}{32.581} & \textbf{28.916} & \textbf{0.6359} & \textbf{0.4133} \\ 

\midrule

\multirow{3}{*}{$\textbf{TaylorSeer} $ $(\mathcal{N}=5,O=2)$} & & & \textcolor{blue!70}{1.0566} & \textcolor{blue!70}{32.811} & \underline{29.132} & \underline{0.6701} & \underline{0.3757}\\
& \ding{52} & & \underline{1.0592} & \underline{32.822} & \textcolor{blue!70}{29.028} & \textcolor{blue!70}{0.6656} & \textcolor{blue!70}{0.3857}\\
& & \ding{52} & \textbf{1.0602} & \textbf{32.952} & \textbf{29.423} & \textbf{0.6929} & \textbf{0.3386} \\ 

\bottomrule
\end{tabular}
}
\vspace{-0.5mm}
{\scriptsize
\begin{itemize}[leftmargin=10pt,topsep=0pt]
    \item \textbf{Note:} For all Balanced Mapping configurations, the hyperparameter $\alpha$ is set to $0.55$
\end{itemize}
}
\vspace{-4mm}
\end{table*}

\begin{table*}[htbp]
\centering
\caption{\textbf{ Ablation study of Phase Mapping components on TaylorSeer for HunyuanVideo on the VBench benchmark.}}
\label{table:taylor-hyvideo-ablation-1}
\vspace{-3mm}
\setlength\tabcolsep{7.0pt}
\small
\resizebox{\textwidth}{!}{
\begin{tabular}{ c | c  c|c|c|c}
\toprule
\multirow{2}{*}{\bf Configuration}  &  \multicolumn{2}{c|}{\bf Phase Mapping} & \multirow{2}{*}{\bf PSNR$\uparrow$} & \multirow{2}{*}{\bf SSIM$\uparrow$} & \multirow{2}{*}{\bf LPIPS$\downarrow$}\\
\cline{2-3}

 &  {\bf Chebyshev} & {\bf Balance} & & & \\
\midrule

\multirow{3}{*}{$\textbf{TaylorSeer} $ $(\mathcal{N}=7,O=1)$} & & & \textcolor{blue!70}{15.520} & \textcolor{blue!70}{0.5641} & \underline{0.4581}\\
& \ding{52} & & \textcolor{blue!70}{15.520} & \underline{0.5643} & \textcolor{blue!70}{0.4582}\\
& & \ding{52} & \textbf{18.111} & \textbf{0.6344} & \textbf{0.3552} \\ 

\midrule

\multirow{3}{*}{$\textbf{TaylorSeer} $ $(\mathcal{N}=5,O=1)$} & & & \textcolor{blue!70}{17.117} & \textcolor{blue!70}{0.6316} & \textcolor{blue!70}{0.3690}\\
& \ding{52} & & \underline{17.125} & \underline{0.6320} & \underline{0.3689}\\
& & \ding{52} & \textbf{19.134} & \textbf{0.6839} & \textbf{0.3013} \\ 

\bottomrule
\end{tabular}
}
\vspace{-0.01mm}
{\scriptsize
\begin{itemize}[leftmargin=10pt,topsep=0pt]
    \item \textbf{Note:} For all Balanced Mapping configurations, the hyperparameter $\alpha$ is set to $0.55$
\end{itemize}
}
\vspace{-4mm}
\end{table*}

\begin{table*}[htbp]
\centering
\caption{\textbf{ Ablation study of Phase Mapping components on TaylorSeer for HunyuanVideo on the T2V-CompBench benchmark.}}
\label{table:taylor-hyvideo-ablation-2}
\vspace{-3mm}
\setlength\tabcolsep{7.0pt}
\small
\resizebox{\textwidth}{!}{
\begin{tabular}{ c | c  c|c|c|c}
\toprule
\multirow{2}{*}{\bf Configuration}  &  \multicolumn{2}{c|}{\bf Phase Mapping} & \multirow{2}{*}{\bf PSNR$\uparrow$} & \multirow{2}{*}{\bf SSIM$\uparrow$} & \multirow{2}{*}{\bf LPIPS$\downarrow$}\\
\cline{2-3}

 &  {\bf Chebyshev} & {\bf Balance} & & & \\
\midrule

\multirow{3}{*}{$\textbf{TaylorSeer} $ $(\mathcal{N}=7,O=1)$} & & & \underline{15.068} & \underline{0.5517} & \underline{0.4533}\\
& \ding{52} & & \textcolor{blue!70}{15.064} & \textcolor{blue!70}{0.5516} & \textcolor{blue!70}{0.4535}\\
& & \ding{52} & \textbf{17.278} & \textbf{0.5956} & \textbf{0.3682} \\ 

\midrule

\multirow{3}{*}{$\textbf{TaylorSeer} $ $(\mathcal{N}=5,O=1)$} & & & \underline{16.746} & \textcolor{blue!70}{0.6132} & \textcolor{blue!70}{0.3640}\\
& \ding{52} & & \textcolor{blue!70}{16.743} & \underline{0.6134} & \underline{0.3638}\\
& & \ding{52} & \textbf{18.239} & \textbf{0.6496} & \textbf{0.3132} \\ 

\bottomrule
\end{tabular}
}
\vspace{-0.01mm}
{\scriptsize
\begin{itemize}[leftmargin=10pt,topsep=0pt]
    \item \textbf{Note:} For all Balanced Mapping configurations, the hyperparameter $\alpha$ is set to $0.55$
\end{itemize}
}
\vspace{-4mm}
\end{table*}

\begin{table*}[htbp]
\centering
\caption{\textbf{Ablation study of Phase Mapping components on TaylorSeer for DiT-XL/2.}}
\label{table:taylor-dit-ablation}
\vspace{-3mm}
\setlength\tabcolsep{7.0pt}
\small
\resizebox{\textwidth}{!}{
\begin{tabular}{ c | c  c|c|c|c}
\toprule
\multirow{2}{*}{\bf Configuration}  &  \multicolumn{2}{c|}{\bf Phase Mapping} & \multirow{2}{*}{\bf FID$\downarrow$} & \multirow{2}{*}{\bf sFID$\downarrow$} & \multirow{2}{*}{\bf IS$\uparrow$}\\
\cline{2-3}

 &  {\bf Chebyshev} & {\bf Balance} & & & \\
\midrule

\multirow{3}{*}{$\textbf{TaylorSeer} $ $(\mathcal{N}=13,O=2)$} & & & \underline{15.415} & \textbf{16.005} & \underline{120.76}\\
& \ding{52} & & \textbf{10.429} & \underline{18.908} & \textbf{150.41}\\
& & \ding{52} & \textcolor{blue!70}{22.711} & \textcolor{blue!70}{26.638} & \textcolor{blue!70}{97.14} \\ 

\midrule

\multirow{3}{*}{$\textbf{TaylorSeer} $ $(\mathcal{N}=8,O=3)$} & & & \underline{4.807} & \underline{7.088} & \underline{197.22}\\
& \ding{52} & & \textbf{3.822} & \textbf{6.873} & \textbf{197.81}\\
& & \ding{52} & \textcolor{blue!70}{6.345} & \textcolor{blue!70}{9.873} & \textcolor{blue!70}{186.81} \\ 

\midrule

\multirow{3}{*}{$\textbf{TaylorSeer} $ $(\mathcal{N}=4,O=3)$} & & & \underline{2.676} & \underline{5.038} & \underline{230.41}\\
& \ding{52} & & \textbf{2.580} & \textbf{5.006} & \textbf{231.36}\\
& & \ding{52} & \textcolor{blue!70}{3.020} & \textcolor{blue!70}{5.483} & \textcolor{blue!70}{228.36} \\ 

\bottomrule
\end{tabular}
}
\vspace{-0.01mm}
{\scriptsize
\begin{itemize}[leftmargin=10pt,topsep=0pt]
    \item \textbf{Note:} For all Balanced Mapping configurations, the hyperparameter $\alpha$ is set to $0.55$
\end{itemize}
}
\vspace{-4mm}
\end{table*}

\begin{table*}[htbp]
\centering
\caption{\textbf{ Ablation study of Phase Mapping components on HiCache for FLUX.1-dev.}}
\label{table:hicache-FLUX-ablation}
\vspace{-3mm}
\setlength\tabcolsep{7.0pt}
\small
\resizebox{\textwidth}{!}{
\begin{tabular}{ c | c  c|c|c|c|c|c}
\toprule
\multirow{2}{*}{\bf Configuration}  &  \multicolumn{2}{c|}{\bf Phase Mapping} &   {\bf ImageReward $\uparrow$} & \bf CLIP$\uparrow$ & \multirow{2}{*}{\bf PSNR$\uparrow$} & \multirow{2}{*}{\bf SSIM$\uparrow$} & \multirow{2}{*}{\bf LPIPS$\downarrow$}\\
\cline{2-3}

 &  {\bf Chebyshev} & {\bf Balance} & \bf DrawBench & \bf Score & & & \\
\midrule

\multirow{3}{*}{$\textbf{HiCache} $ $(\mathcal{N}=11,O=2)$} & & & \underline{0.8040} & \underline{31.604} & \underline{28.268} & \underline{0.5261} & \underline{0.5955}\\
& \ding{52} & & \textcolor{blue!70}{0.7345} & \textcolor{blue!70}{31.138} & \textcolor{blue!70}{28.205} & \textcolor{blue!70}{0.5241} & \textcolor{blue!70}{0.6151}\\
& & \ding{52} & \textbf{0.9024} & \textbf{32.019} & \textbf{28.429} & \textbf{0.5750} & \textbf{0.5407} \\ 

\midrule

\multirow{3}{*}{$\textbf{HiCache} $ $(\mathcal{N}=7,O=2)$} & & & \textcolor{blue!70}{1.0079} & \underline{32.622} & \textcolor{blue!70}{28.705} & \textcolor{blue!70}{0.6147} & \textcolor{blue!70}{0.4542}\\
& \ding{52} & & \underline{1.0081} & \textcolor{blue!70}{32.388} & \underline{28.561} & \underline{0.6180} & \underline{0.4422}\\
& & \ding{52} & \textbf{1.0198} & \textbf{32.672} & \textbf{28.889} & \textbf{0.6269} & \textbf{0.4346} \\ 

\midrule

\multirow{3}{*}{$\textbf{HiCache} $ $(\mathcal{N}=5,O=2)$} & & & \textcolor{blue!70}{1.0438} & \underline{32.898} & \textcolor{blue!70}{29.240} & \textcolor{blue!70}{0.6830} & \underline{0.3568}\\
& \ding{52} & & \underline{1.0495} & \textbf{32.975} & \underline{29.342} & \underline{0.6872} & \textcolor{blue!70}{0.3615}\\
& & \ding{52} & \textbf{1.0609} & \textcolor{blue!70}{32.862} & \textbf{29.408} & \textbf{0.6925} & \textbf{0.3547} \\ 

\bottomrule
\end{tabular}
}
\vspace{-0.5mm}
{\scriptsize
\begin{itemize}[leftmargin=10pt,topsep=0pt]
    \item \textbf{Note:} For all Balanced Mapping configurations, the hyperparameter $\alpha$ is set to $0.55$
\end{itemize}
}
\vspace{-4mm}
\end{table*}

\section{In-Depth Analysis of ODE-Aligned Macro-Trajectory Targeting}

In this section, we provide an in-depth theoretical and empirical analysis to justify our proposed ODE-Aligned Macro-Trajectory Targeting. We mathematically demonstrate why forecasting the Global Drift (GD) strictly yields lower approximation errors than forecasting the Final Output (FO), and we validate how this macroscopic formulation fundamentally resolves the severe memory bottlenecks of existing layer-wise predictors.

\subsection{Theoretical Error Comparison: Global Drift vs. Final Output}

From a dynamic systems perspective, the complete forward pass of a Diffusion Transformer at timestep $t$ can be formulated as mapping an input latent state $x_t$ to a final absolute output $G(x_t)$. This mapping fundamentally consists of the highly correlated input prior $x_t$ and the residual displacement, which we define as the Global Drift $D(x_t)$:
\begin{equation}
G(x_t) = x_t + D(x_t).
\end{equation}

When adopting a Final Output (FO) forecasting strategy, a polynomial predictor $P_{\text{FO}}(t)$ of degree $m$ is directly fitted to the historical absolute outputs $G(x_{t_j})$. According to classical polynomial interpolation theory, the theoretical error bound for forecasting the target timestep $t$ is governed by the $(m+1)$-th derivative of the target function $G(x_t)$:
\begin{equation}
E_{\text{FO}}(t) = \|G(x_t) - P_{\text{FO}}(t)\| \leq \frac{\|G^{(m+1)}(\xi)\|}{(m+1)!} \prod_{j=0}^{m} |t - t_j|,
\end{equation}
where $\xi$ lies within the extrapolation interval. Since $G(x_t) = x_t + D(x_t)$, the high-order derivative of the absolute output inherently absorbs the derivative of the input prior:
\begin{equation}
\|G^{(m+1)}(\xi)\| \leq \left\| \frac{d^{m+1} x_t}{dt^{m+1}} \right\| + \|D^{(m+1)}(\xi)\|.
\end{equation}
In a diffusion probability flow ODE, the latent state trajectory $x_t$ is highly dynamic and oscillatory across discrete sampling steps. Consequently, its higher-order temporal derivatives $\frac{d^{m+1} x_t}{dt^{m+1}}$ possess exceptionally large magnitudes. By forcing the polynomial to fit $G(x_t)$, the predictor implicitly attempts to model this chaotic analytical prior, suffering from a massively inflated truncation error bound driven by $\|x^{(m+1)}_t\|$.

Conversely, our proposed Global Drift (GD) forecasting explicitly isolates the smooth residual dynamics. A polynomial $P_{\text{GD}}(t)$ is fitted solely to the historical drifts $D(x_{t_j})$. The final output is then reconstructed by analytically adding the exact, fully preserved input prior $x_t$: $\hat{G}(x_t) = x_t + P_{\text{GD}}(t)$. The prediction error is thus strictly isolated to the drift term:
\begin{equation}
E_{\text{GD}}(t) = \|G(x_t) - (x_t + P_{\text{GD}}(t))\| = \|D(x_t) - P_{\text{GD}}(t)\| \leq \frac{\|D^{(m+1)}(\xi)\|}{(m+1)!} \prod_{j=0}^{m} |t - t_j|.
\end{equation}

By perfectly retaining the true mathematical prior $x_t$ through residual addition rather than numerical approximation, our GD formulation completely eliminates the highly erratic $\left\| \frac{d^{m+1} x_t}{dt^{m+1}} \right\|$ term from the interpolation error bound. Therefore, it mathematically guarantees a strictly tighter and highly stable error bound:
\begin{equation}
E_{\text{GD}}(t) \ll E_{\text{FO}}(t).
\end{equation}

\subsection{VRAM Consumption Analysis}

\begin{figure*}
    \centering
    \includegraphics[trim= 0 0 0 0, clip, width=\linewidth]{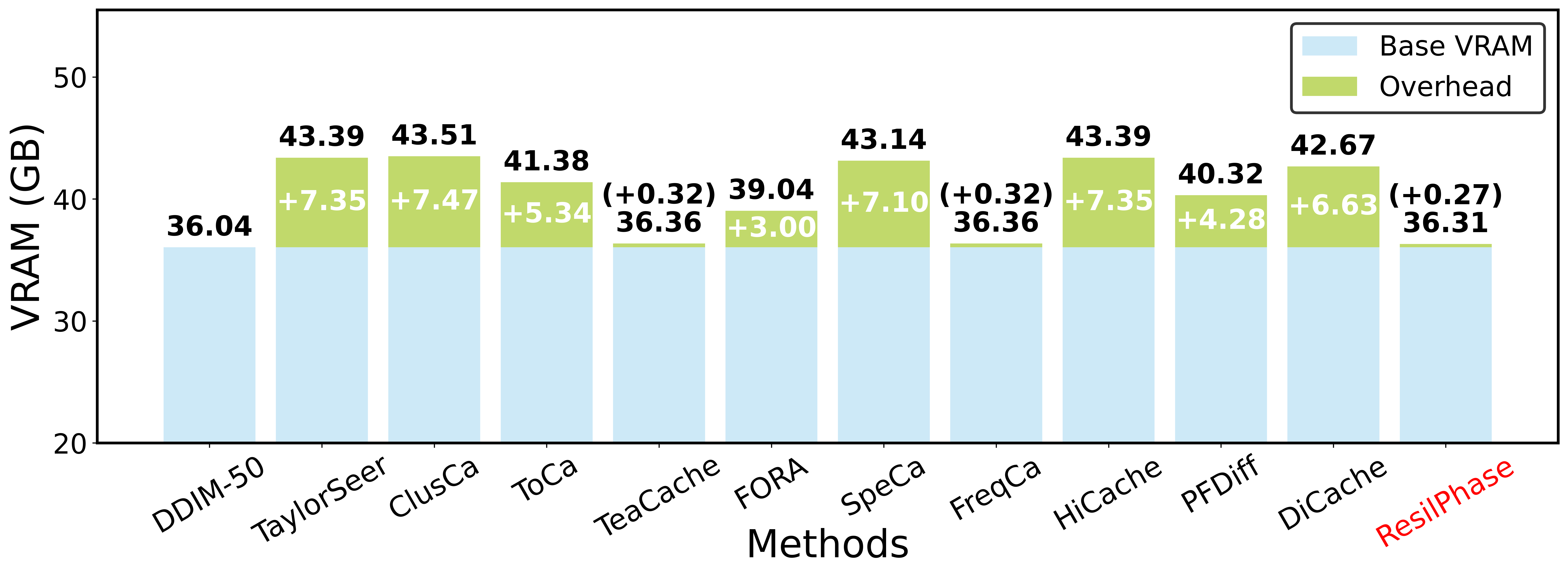}
    \vspace{-6mm}
    \caption{\textbf{VRAM consumption comparison on FLUX.1-dev.} All methods are evaluated at their highest acceleration tier ($\sim 5\times$ speedup) corresponding to Table 1 in the main paper. While layer-wise caching methods incur massive memory overheads (e.g., +7.35 GB for TaylorSeer and HiCache), ResilPhase achieves a near-zero memory footprint (+0.27 GB) overhead, rivaling the original unaccelerated DDIM-50 baseline.}
    \label{fig:vram_comparison}
    \vspace{-3mm}
\end{figure*}

Figure \ref{fig:vram_comparison} illustrates the peak VRAM consumption of various acceleration methods on FLUX.1-dev. The results reveal a severe bottleneck in prevailing layer-wise forecasting paradigms: methods such as TaylorSeer, ClusCa, and HiCache demand up to 7.47 GB of additional memory to cache highly multidimensional micro-features across all intermediate Transformer blocks. In stark contrast, ResilPhase incurs a negligible +0.27 GB overhead, maintaining a footprint virtually identical to the unaccelerated $\times 1.00$ baseline. This extraordinary memory efficiency directly validates our core theoretical design: by shifting the prediction objective from localized layer-wise features to the ODE-aligned Global Drift, ResilPhase completely eliminates the need to cache intermediate block outputs, establishing itself as the most deployment-friendly acceleration framework.

% \ \\ \ \\ \ \\ \ \\ \ \\ \ \\ \ \\ \ \\ \ \\ \ \\ \ \\ \ \\ 
\clearpage
\section{Analysis of Feature Derivative Instability}
\begin{figure}[ht]
\centering  % ✅ 关键：让整个图居中
\begin{subfigure}[b]{\linewidth}
  \centering
  \includegraphics[width=\linewidth]{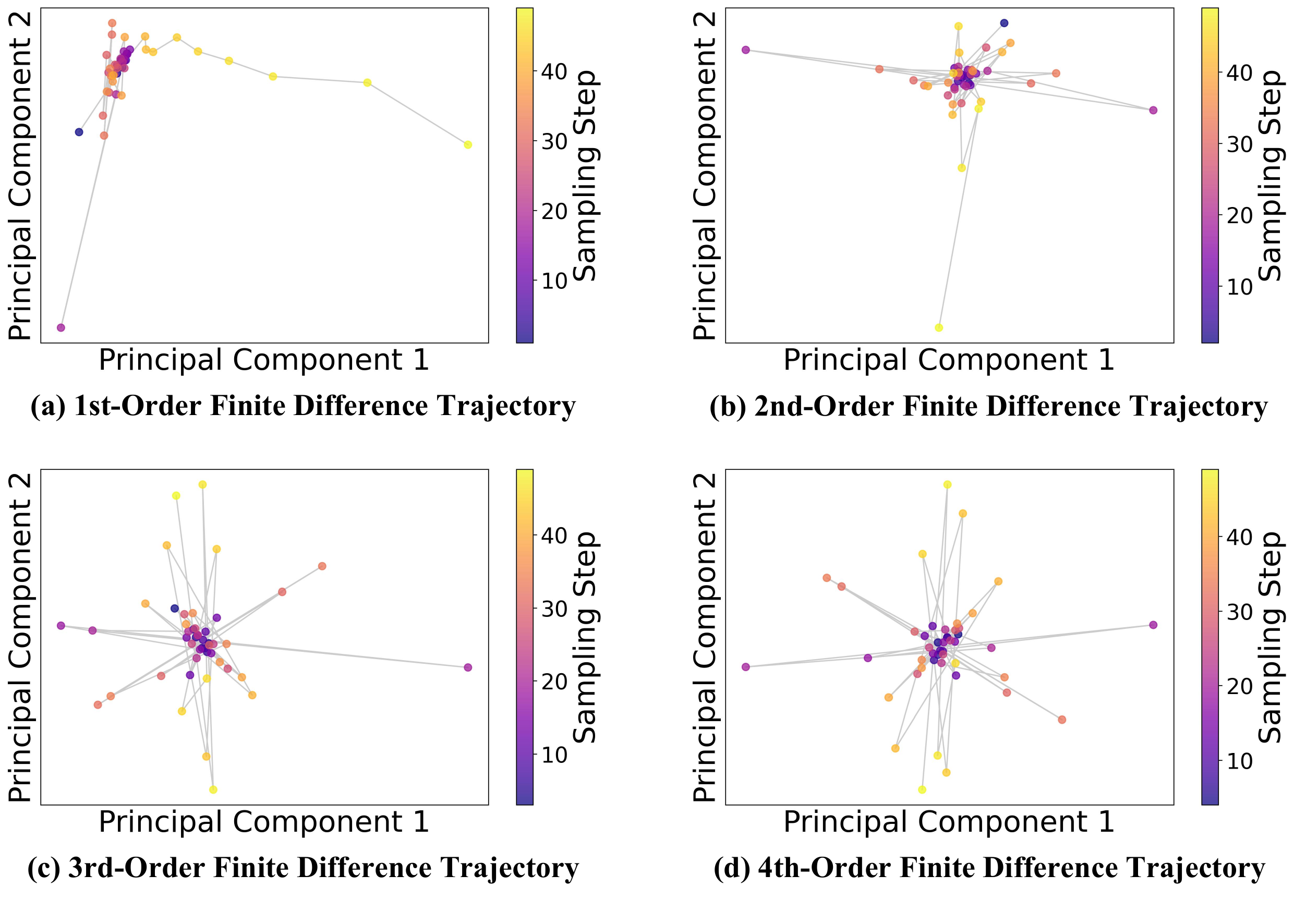}
  
\end{subfigure}

\vspace{-1em} % 手动调间距（可选）

\caption{\textbf{Trajectory of Attention Feature Derivatives in the First Layer.}PCA visualization of 1st-4th order finite difference derivatives. The highly erratic and non-smooth trajectories (especially in b-d) demonstrate that derivative estimates are inherently unstable and unsuitable for polynomial forecasting.
}

\label{fig:attention-1}
\vspace{-0.05em}
\end{figure}

\begin{figure}[ht]
\centering  % ✅ 关键：让整个图居中
\begin{subfigure}[b]{\linewidth}
  \centering
  \includegraphics[width=\linewidth]{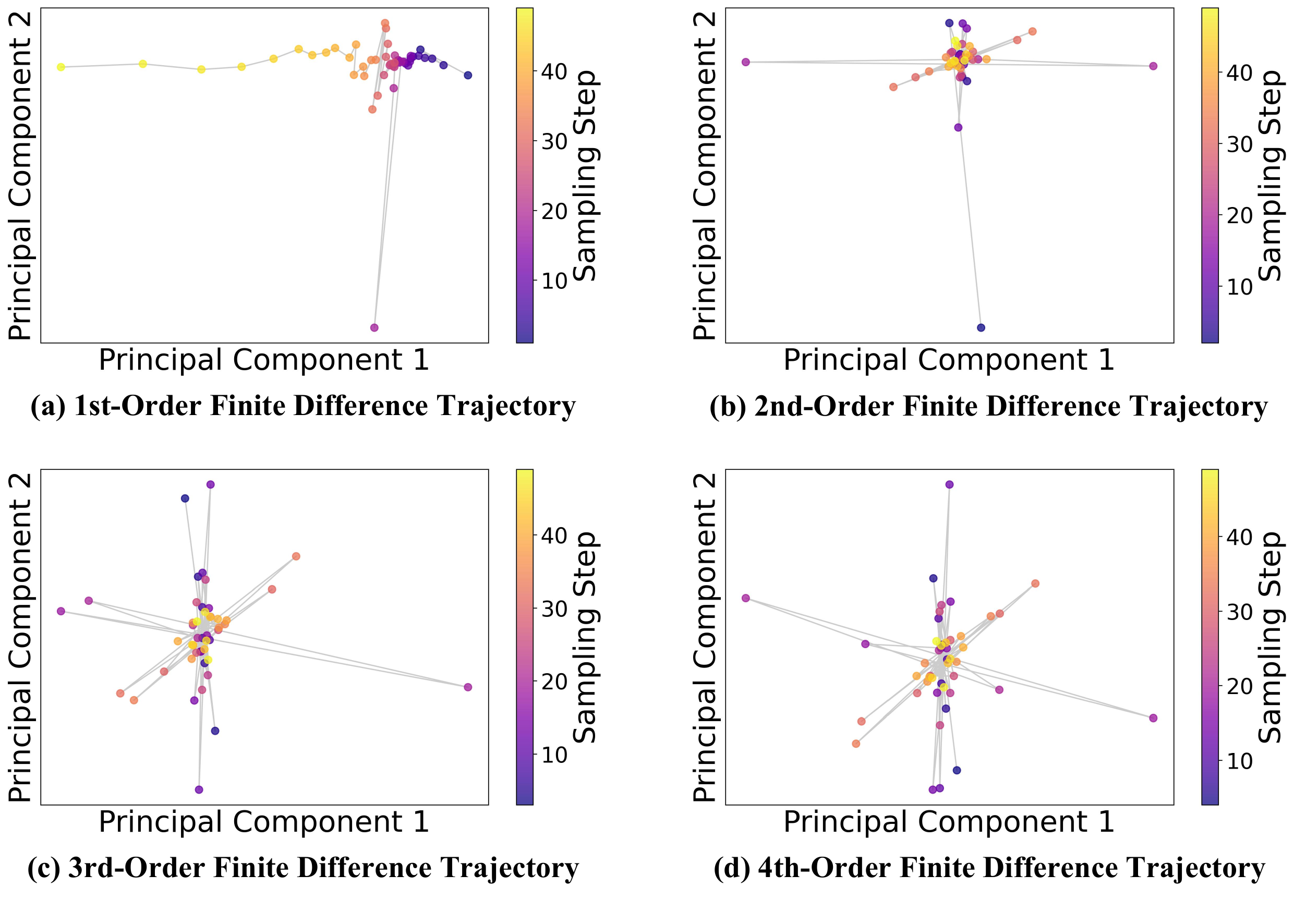}
  
\end{subfigure}

\vspace{-1em} % 手动调间距（可选）

\caption{\textbf{Trajectory of MLP Feature Derivatives in the First Layer.} Similar to attention features, MLP derivatives exhibit noisy and discontinuous trajectories across all orders. This reinforces that layer-wise, derivative-based prediction relies on unstable signals, leading to significant errors.
}

\label{fig:mlp-1}
\vspace{-0.05em}
\end{figure}

\begin{figure}[ht]
\centering  % ✅ 关键：让整个图居中
\begin{subfigure}[b]{\linewidth}
  \centering
  \includegraphics[width=\linewidth]{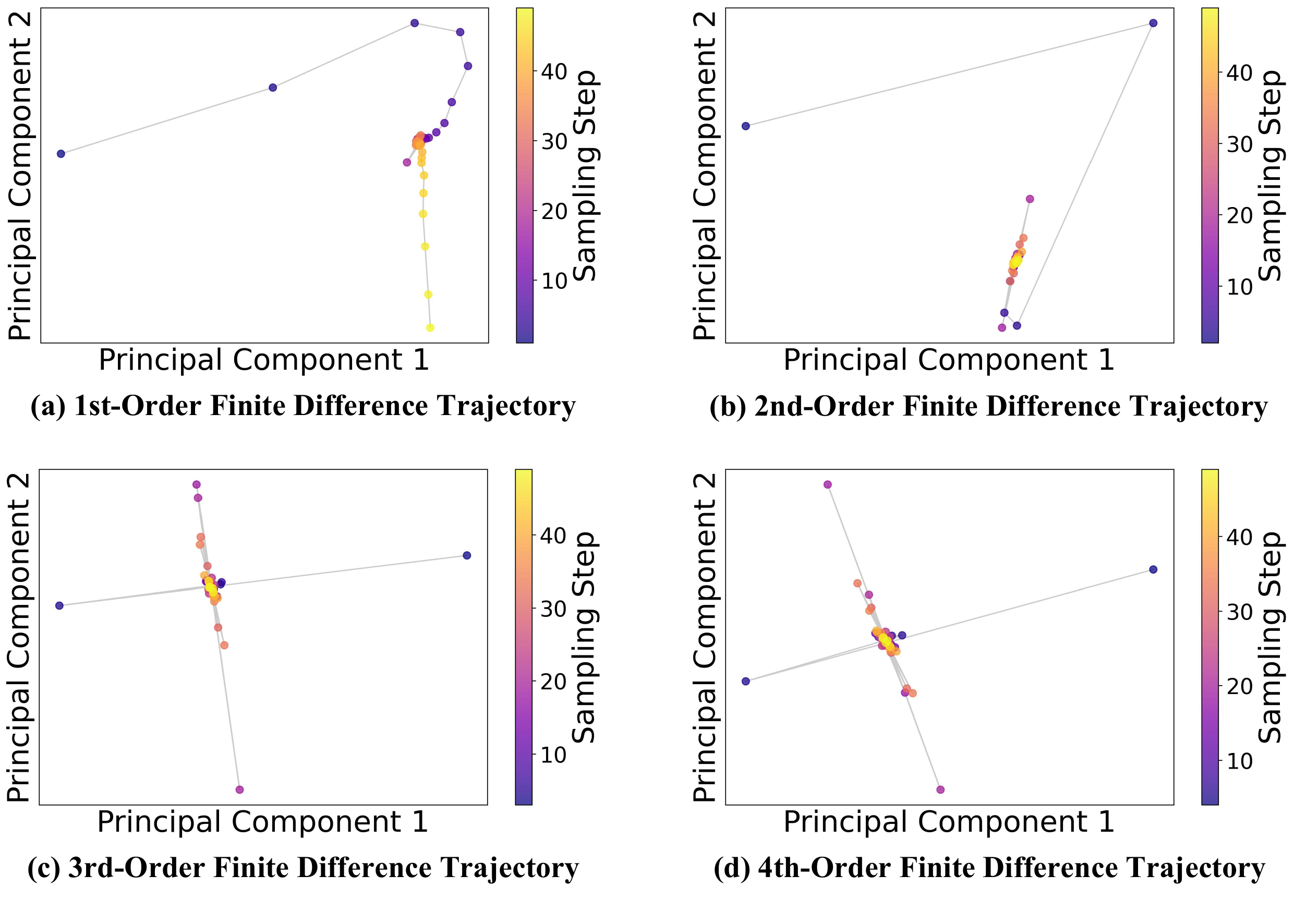}
  
\end{subfigure}

\vspace{-1em} % 手动调间距（可选）

\caption{\textbf{Trajectory of Attention Feature Derivatives in the Final Layer.}Instability persists even in the final layer. The chaotic trajectories of finite difference derivatives further validate that the assumption of smooth higher-order derivatives does not hold at any stage of the network.
}

\label{fig:attention-last}
\vspace{-0.05em}
\end{figure}

\begin{figure}[ht]
\centering  % ✅ 关键：让整个图居中
\begin{subfigure}[b]{\linewidth}
  \centering
  \includegraphics[width=\linewidth]{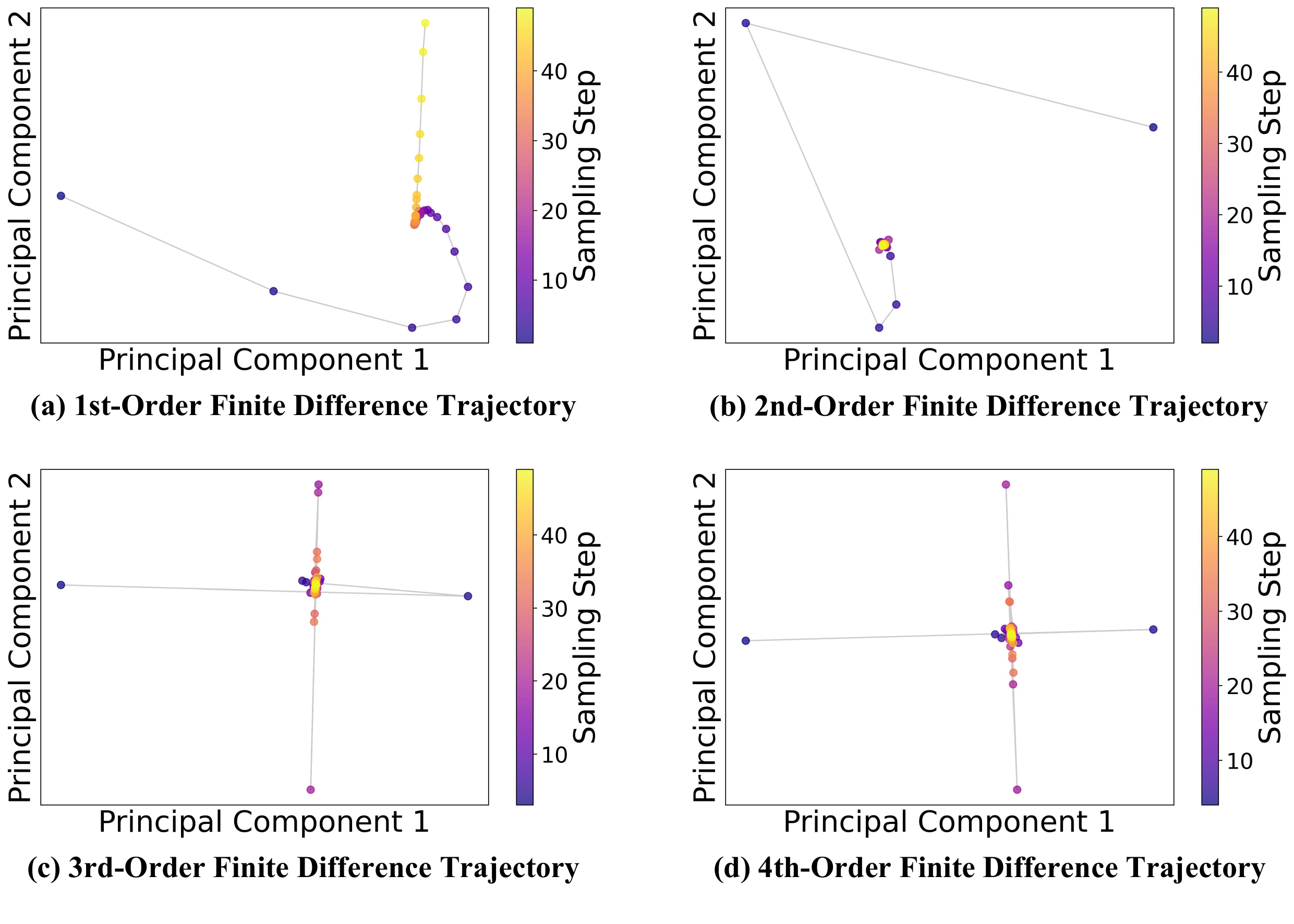}
  
\end{subfigure}

\vspace{-1em} % 手动调间距（可选）

\caption{\textbf{Trajectory of MLP Feature Derivatives in the Final Layer.}Consistent with early layers, the final layer's MLP derivatives show unpredictable behavior. This confirms that severe instability is a systemic issue across network depths, undermining derivative-based forecasting methods.
}

\label{fig:mlp-last}
\vspace{-0.05em}
\end{figure}

\begin{figure}[ht]
\centering  % ✅ 关键：让整个图居中
\begin{subfigure}[b]{\linewidth}
  \centering
  \includegraphics[width=\linewidth]{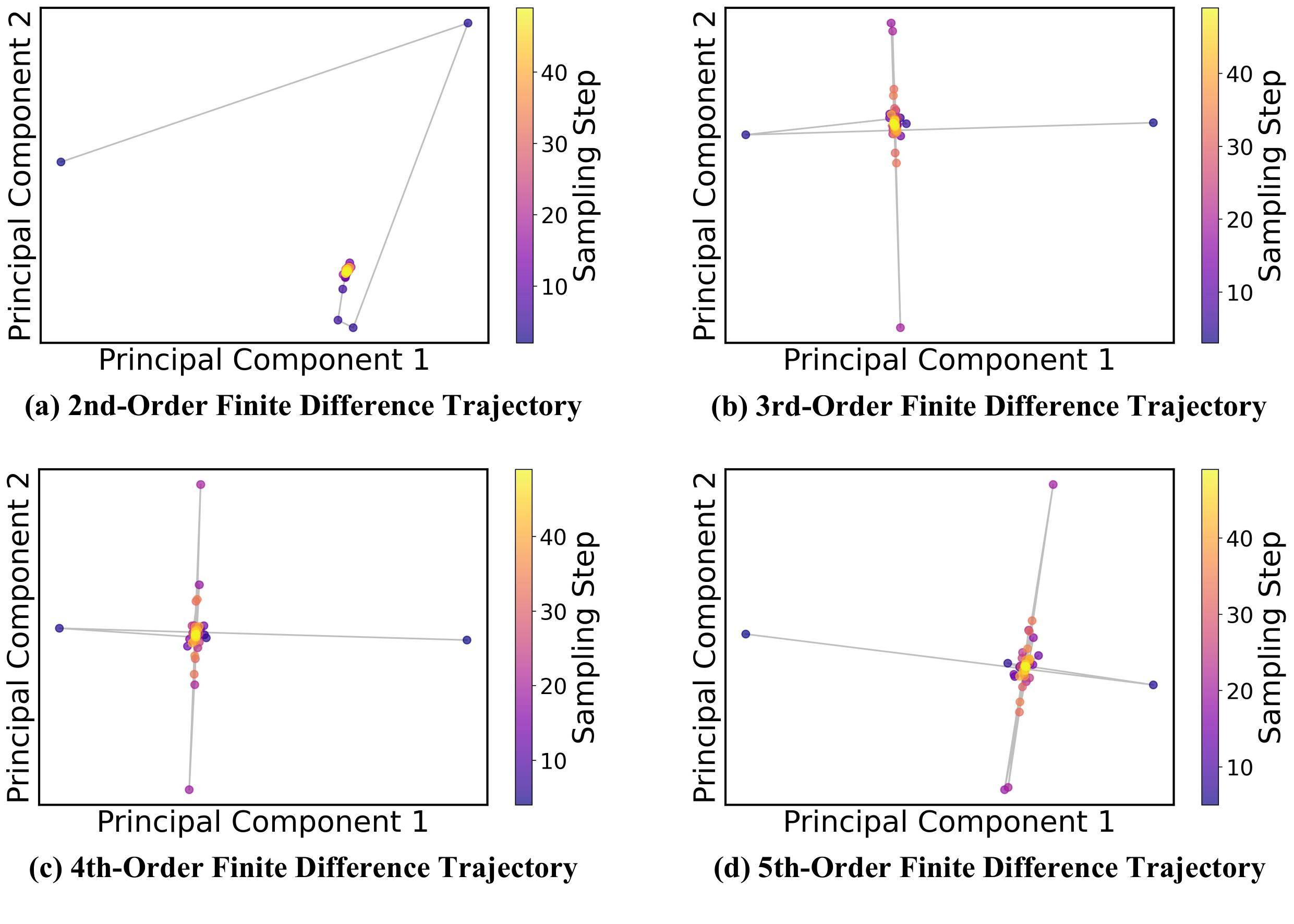}
  
\end{subfigure}

\vspace{-1em} % 手动调间距（可选）

\caption{\textbf{Trajectory of Higher-Order Global Residual Derivatives.} PCA projections of 2nd- to 5th-order finite difference approximations of the Global Residual. As visualized across all higher orders, the trajectories are highly erratic, characterized by extreme, unpredictable jumps and lack of continuity. This conclusively demonstrates that the intrinsic chaotic noise persists even at higher derivative orders, rendering any derivative-based polynomial forecasting highly unreliable.}

\label{fig:higher-order}
\vspace{-0.05em}
\end{figure}

To investigate the premise of derivative-based forecasting, we analyzed the trajectories of feature derivatives across the network. As visualized in Figures~\ref{fig:attention-1} through~\ref{fig:mlp-last}, the 1st- to 4th-order derivatives of layer-wise features—spanning both attention and MLP components in the first and final layers—are consistently erratic and non-smooth. This instability extends to the Global Residual itself, whose higher-order derivatives (2nd- to 5th-order) are shown to be equally chaotic in Figure~\ref{fig:higher-order}. This pervasive instability across both layer-wise and global features leads to a critical conclusion: the core assumption of using derivative-based polynomials to model feature trajectories is fundamentally flawed and will inevitably introduce significant approximation errors.

